\documentclass[10pt,journal,cspaper,compsoc]{IEEEtran}

\ifCLASSOPTIONcompsoc
\else
\fi

\ifCLASSINFOpdf
  \usepackage[pdftex]{graphicx}
\else
\fi

\usepackage[cmex10]{amsmath}

\ifCLASSOPTIONcompsoc
\usepackage[tight,normalsize,sf,SF]{subfigure}
\else
\usepackage[tight,footnotesize]{subfigure}
\fi

\usepackage{colortbl}
\usepackage{color}
\usepackage{url}
\usepackage{multirow}
\usepackage{rotating}

\begin{document}
\title{Shape, Illumination, and Reflectance \\ from Shading}

\author{Jonathan~T.~Barron,~\IEEEmembership{Member,~IEEE,}
        and~Jitendra~Malik,~\IEEEmembership{Fellow,~IEEE}%
\IEEEcompsocitemizethanks{\IEEEcompsocthanksitem  J.T. Barron and J. Malik are with the Department of Electrical Engineering
and Computer Science, University of California at Berkeley, Berkeley, CA
94720.
\protect\\
E-mail: {barron,malik}@eecs.berkeley.edu
}%
\thanks{}}

\definecolor{color1}{rgb}{ 0, 0.3569, 0.8353 }
\definecolor{color2}{rgb}{ 0.8235, 0.1725, 0.1725}

\newcommand{\pd}[2]{\frac{\partial#1}{\partial#2}}

\newcommand{\logrho}{A}
\newcommand{\light}{L}
\newcommand{\lossfun}{f}

\newcommand{\K}{\kappa}
\newcommand{\sixwidth}{0.95in}
\newcommand{\fivewidth}{1.23in}
\newcommand{\fourwidth}{1.45in}
\newcommand{\fourwidthP}{1.62in}
\newcommand{\fourwidthA}{.9in}
\newcommand{\eightwidth}{.7in}
\newcommand{\eightwidthB}{.60in}
\newcommand{\eightwidthp}{.72in}
\newcommand{\tenwidth}{.56in}
\newcommand{\tableAwidth}{.72in}
\newcommand{\tableBwidth}{.15in}
\newcommand{\fourwidthpad}{1.15in}
\newcommand{\threewidth}{1.5in}
\newcommand{\twowidth}{3.25in}
\newcommand{\onewidth}{6.75in}
\newcommand{\comparewidth}{0.49in}
\newcommand{\mitwidth}{0.52in}
\newcommand{\halfmitwidth}{0.26in}

\newcommand{\realwidth}{0.375in}
\newcommand{\halfrealwidth}{0.19in}
\newcommand{\ablatewidth}{0.72in}
\newcommand{\ablatewidthB}{0.72in}
\newcommand{\fakesub}{\hspace{.07in}}
\newcommand{\fakesubvert}{\vspace{.07in}}
\newcommand{\argmin}[1]{\underset{#1}{\operatorname{arg\,min}\,}}
\newcommand{\argmax}[1]{\underset{#1}{\operatorname{arg\,max}\,}}
\renewcommand{\min}[1]{\underset{#1}{\operatorname{min}\,}}
\newcommand{\SMSE}{S\text{-}\mathrm{MSE}}
\newcommand{\AMSE}{R\text{-}\mathrm{MSE}}
\newcommand{\IMSE}{\mathit{e}}
\newcommand{\ZMAE}{Z\text{-}\mathrm{MAE}}
\newcommand{\NMAE}{N\text{-}\mathrm{MAE}}
\newcommand{\mitMSE}{RS\text{-}\mathrm{MSE}}
\newcommand{\lightMSE}{L\,\text{-}\mathrm{MSE}}
\newcommand{\MSE}{\mathrm{MSE}}
\newcommand{\LightMSE}{\mathit{light}\text{-}\mathrm{MSE}}
\newcommand{\sign}{\mathrm{sign}}
\newcommand{\half}{\nicefrac{1}{2}}
\newcommand{\GR}{R_\mathcal{G}}
\newcommand{\GZ}{Z_\mathcal{G}}
\newcommand{\figone}{1.48in}
\newcommand{\figonea}{2.96in}
\newcommand{\Var}{\mathrm{Var}}
\newcommand{\Zprob}{U}
\newcommand{\Lprob}{V}
\newcommand{\Zweight}{\psi}
\newcommand{\Lweight}{\omega}
\newcommand{\Zerr}{Z^{\mathit{err}}}
\newcommand{\etal}{\textit{et~al.\,}}

\newcommand{\Laplacian}[1]{\mathcal{L}\!\left( #1 \right)}
\newcommand{\invLaplacian}[1]{\mathcal{L}^{-1}\!\left( #1 \right)}
\newcommand{\Gaussian}{\mathcal{G}}
\providecommand{\norm}[1]{\left\lVert#1\right\rVert}
\newcommand{\fuck}[1]{{\color{red}\bf #1}}
\newcommand{\Lz}{Z_\mathcal{L}}
\newcommand{\set}[1]{\boldsymbol{#1}}

\definecolor{Yellow}{rgb}{1,1, 0.6}

\markboth{IEEE TRANSACTIONS ON PATTERN ANALYSIS AND MACHINE INTELLIGENCE}%
{Barron & Malik: Shape, Illumination, and Reflectance from Shading}

\IEEEcompsoctitleabstractindextext{%
\begin{abstract}
A fundamental problem in computer vision is that of inferring the intrinsic, 3D structure of the world from flat, 2D images of that world. Traditional methods for recovering scene properties such as shape, reflectance, or illumination rely on multiple observations of the same scene to overconstrain the problem. Recovering these same properties from a single image seems almost impossible in comparison --- there are an infinite number of shapes, paint, and lights that exactly reproduce a single image. However, certain explanations are more likely than others: surfaces tend to be smooth, paint tends to be uniform, and illumination tends to be natural. We therefore pose this problem as one of statistical inference, and define an optimization problem that searches for the \emph{most likely} explanation of a single image. Our technique can be viewed as a superset of several classic computer vision problems (shape-from-shading, intrinsic images, color constancy, illumination estimation, etc) and outperforms all previous solutions to those constituent problems.
\end{abstract}

}

\maketitle

\IEEEdisplaynotcompsoctitleabstractindextext

\IEEEpeerreviewmaketitle

\section{Introduction}

\IEEEPARstart{A}{t} the core of computer vision is the problem of taking a single image, and estimating the physical world which produced that image. The physics of image formation makes this ``inverse optics'' problem terribly challenging and underconstrained: the space of shapes, paint, and light that exactly reproduce an image is vast.

\begin{figure}[b!]
	\centering
		\begin{tabular}{@{\,}c@{\,}c@{\,}c@{\,}}
		\resizebox{0.98in}{!}{\includegraphics{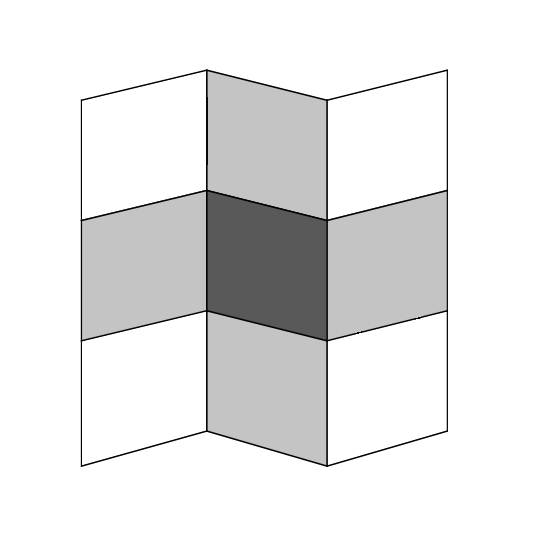}} &
		\resizebox{0.98in}{!}{\includegraphics{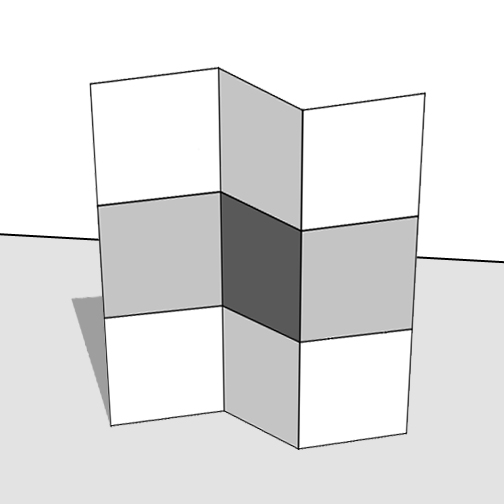}} \\
		\mbox{\scriptsize (a) an image} & \mbox{\scriptsize (b) a likely explanation}
		\end{tabular}
		\\ \vspace{0.15in}
		\begin{tabular}{@{\,}ccc@{\,}}
		\resizebox{0.98in}{!}{\includegraphics{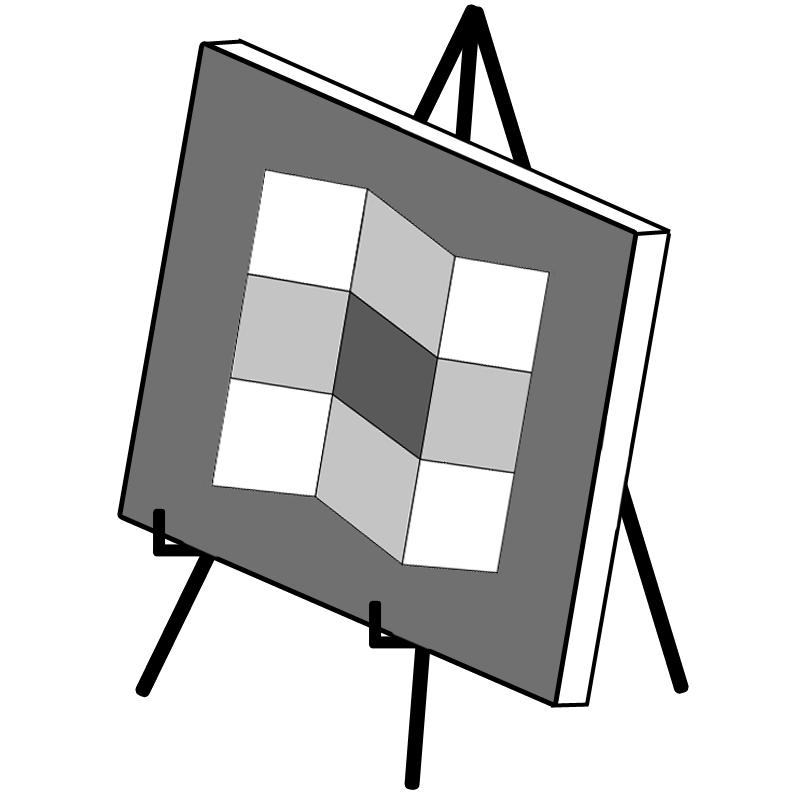}} &
		\resizebox{0.98in}{!}{\includegraphics{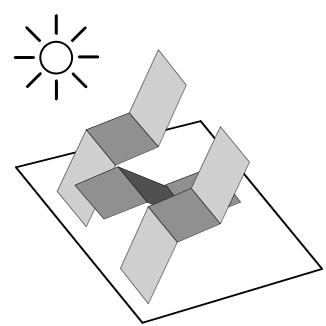}} &
		\resizebox{0.98in}{!}{\includegraphics{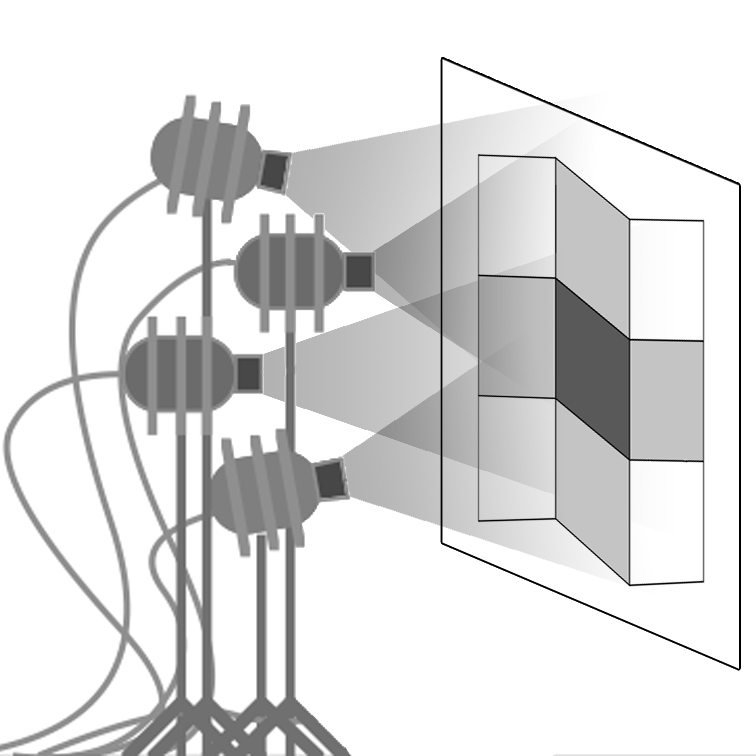}} \\
		\resizebox{0.98in}{!}{\mbox{\scriptsize (c) painter's explanation}} &
		\resizebox{0.98in}{!}{\mbox{\scriptsize (d) sculptor's explanation}} &
		\resizebox{0.98in}{!}{\mbox{\scriptsize (e) gaffer's explanation}}
		\end{tabular}
	\caption{
A visualization of Adelson and Pentland's ``workshop'' metaphor \cite{adelson1996perception}. The image in \ref{fig:Workshop}(a) clearly corresponds to the interpretation in \ref{fig:Workshop}(b), but it could be a painting, a sculpture, or an arrangement of lights.
		\label{fig:Workshop}}
\end{figure}

This problem is perhaps best motivated using Adelson and Pentland's ``workshop'' metaphor \cite{adelson1996perception}: consider the image in Figure~\ref{fig:Workshop}(a), which has a clear percept as a twice-bent surface with a stroke of dark paint (Figure~\ref{fig:Workshop}(b)). But this scene could have been created using any number of physical worlds --- it could be realistic painting on a canvas (Figure~\ref{fig:Workshop}(c)), a complicated arrangement of bent shapes (Figure~\ref{fig:Workshop}(d)), a sophisticated projection produced by a collection of lights (Figure~\ref{fig:Workshop}(e)), or anything in between. The job of a perceptual system is analogous to that of a prudent manager in this ``workshop'', where we would like to reproduce the scene using as little effort from our three artists as possible, giving us Figure~\ref{fig:Workshop}(b).

This metaphor motivates the formulation of this problem as one of statistical inference. Though there are infinitely many possible explanations for a single image, some are more likely than others. Our goal is therefore to recover \emph{the most likely explanation} that explains an input image. We will demonstrate that in natural depth maps, reflectance maps, and illumination models, very strong statistical regularities arise that are similar to those found in natural images \cite{field1987relations,ruderman1994statistics}. We will construct priors similar to those used in natural image statistics, but applied separately to shape, reflectance, and illumination. Our algorithm is simply an optimization problem in which we recover the most likely shape, reflectance, and illumination under these priors that exactly reproduces a single image. Our priors are powerful enough that these intrinsic scene properties can be recovered from a single image, but are general enough that they work across a variety of objects.

\begin{figure}[t!]
	\centering
    \resizebox{3.25in}{!}{\includegraphics{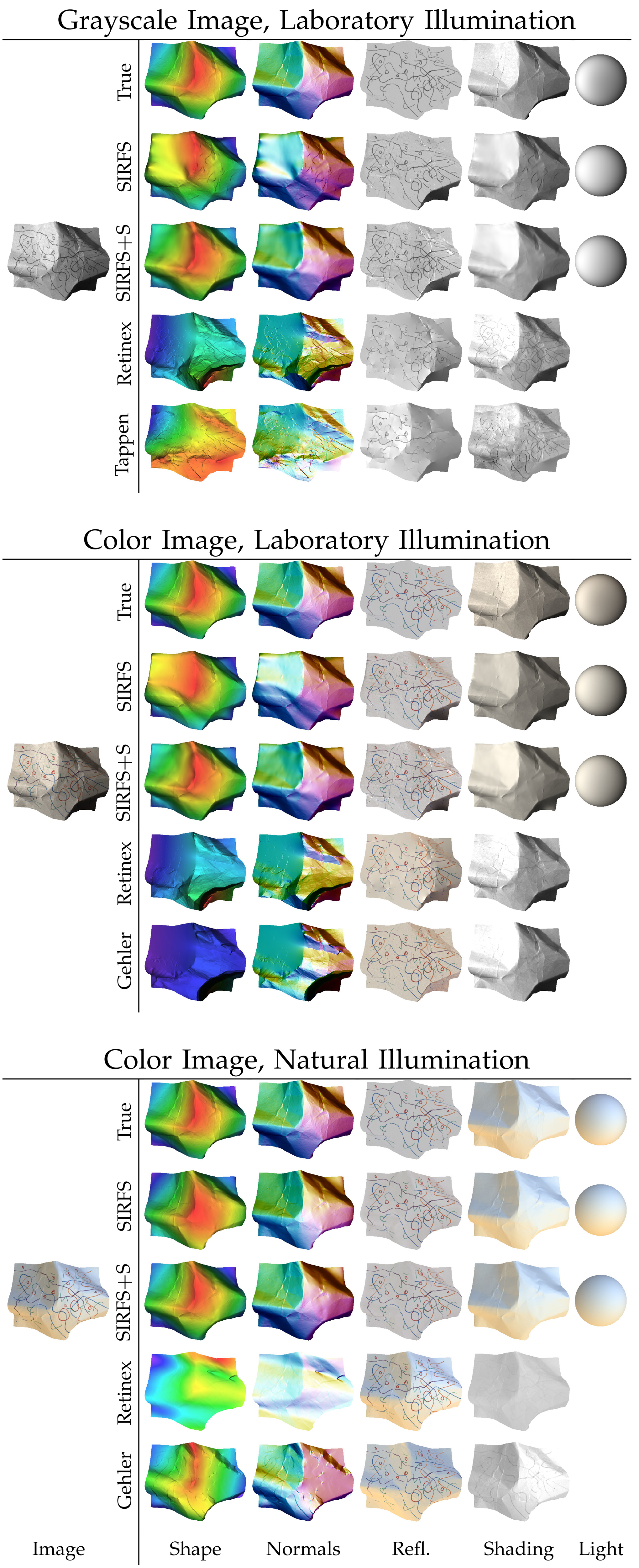}}
	\caption{A single image from our dataset, under three color/illumination conditions. For each condition, we present the ground-truth, the output of SIRFS, the output of SIRFS$+$S (which uses external shape information), and the two best-performing intrinsic image techniques (for which we do SFS on the recovered shading to recover shape).
	 \label{fig:MITImages}}
\end{figure}

The output of our model relative to ground-truth can be seen in Figure~\ref{fig:MITImages}.  Our model is capable of producing qualitatively correct reconstructions of shape, surface normals, shading, reflectance, and illumination, from a single image. We quantitatively evaluate our model on variants of the MIT intrinsic images dataset~\cite{grosse09intrinsic}, on which we quantitatively outperform all previously published intrinsic image or shape-from-shading algorithms. We additionally present qualitative results for many more real-world images, for which we do not have ground-truth explanations.

Earlier versions of this work have been presented in a piecemeal fashion, over the course of many papers \cite{Barron2011,Barron2012A,Barron2012B}. This paper is meant to simplify and unify those previous methods.

This paper will proceed as follows: In Section~\ref{sec:past}, we will review past work as it relates to our own. In Section~\ref{sec:SIRFS} we will formulate our problem as one of statistical inference and optimization, with respect to a set of priors over shape, reflectance, and illumination. In Sections~\ref{sec:reflectance}, \ref{sec:shape}, and \ref{sec:illumination} we present and motivate our priors on reflectance, shape, and illumination, respectively. In Section~\ref{sec:optimization} we explain how we solve our proposed optimization problem.
In Section~\ref{sec:experiments} we present a series of experiments with our model on variants of the MIT Intrinsic Images dataset \cite{grosse09intrinsic} and on real-world images, and in Section~\ref{sec:conclusion} we conclude.

\section{Prior work}
\label{sec:past}

The question of how humans solve the underconstrained problem of perceiving shape, reflectance, and illumination from a single image appears to be at least one thousand years old, dating back to the scientist Alhazen, who noted that "Nothing of what is visible, apart from light and color, can be perceived by pure sensation, but only by discernment, inference, and recognition, in addition to sensation." In the 19th century the problem was studied by such prominent vision scientists as von Helmholtz, Hering and Mach \cite{gilchrist}, who framed the problem as one of ``lightness constancy'' --- how humans, when viewing a flat surface with patches of varying reflectances subject to spatially varying illumination, are able to form a reasonably veridical percept of the reflectance (``lightness'') in spite of the fact that a darker patch under brighter illumination may well have more light traveling from it to the eye compared to a lighter patch which is less well illuminated.

Land's Retinex theory of lightness constancy \cite{Retinex} has been particularly influential in computer vision since its introduction in 1971. It provided a computational approach to the problem in the ``Mondrian World'', a 2D world of flat patches of piecewise constant reflectance. Retinex theory was later made practical by Horn \cite{HornLightness}, who was able to obtain a decomposition of an image into its shading and reflectance components using the prior belief that sharp edges tend to be reflectance, and smooth variation tends to be shading. 

In 1978, Barrow and Tenebaum defined what they called the problem of ``intrinsic images'': recovering properties such as shape, reflectance, and illumination from a single image \cite{Barrow}. In doing so, they described a challenge in computer vision which is still largely unsolved, and which our work directly addresses. Because this problem is so fundamentally underconstrained and challenging, the computer vision community has largely focused its attention on more constrained and tractable sub-problems. Over time, ``intrinsic images'' has become synonymous with the problem that Retinex addressed, that of separating an image into shading and reflectance components \cite{grosse09intrinsic,HornLightness,Retinex}. This area has seen seen some recent progress  \cite{Bell01learninglocal,ShenYJL11,Tappen,Gehler2011}, though the performance of Retinex, despite its age, has proven hard to improve upon \cite{grosse09intrinsic}.  The limiting factor in many of these ``intrinsic image'' algorithms appears to be that they treat ``shading'' as a kind of image, ignoring the fact that shading is, by construction, the product of some shape and some model of illumination.  By addressing a superset of this ``intrinsic image'' problem and recovering shape and illumination instead of shading, our model produces better results than any intrinsic image technique.

Related to the problem of lightness constancy or ``intrinsic images'' is the problem of color constancy, which can be thought of as a generalization of lightness constancy from grayscale to color, in which the problem is simplified by assuming that there is just one single model of illumination for an entire image, rather than a spatially-varying ``shading'' effect. Early techniques for color constancy used gamut mapping techniques~\cite{Forsyth1990}, finite dimensional models of reflectance and illumination~\cite{Maloney1986}, and physically based techniques for exploiting specularities~\cite{Klinker1990ijcv}. More recent work uses contemporary probabilistic tools, such as modeling the correlation between colors in a  scene~\cite{Finlayson2001}, or performing inference over priors on reflectance and illumination~\cite{Brainard1997}. All of this work shares the assumptions of ``intrinsic image'' algorithms that shape (and to a lesser extent, shading) can be ignored or abstracted away.

The second subset of the Barrow and Tenenbaum's original ``intrinsic image'' formulation that the computer vision research community has focused on is the ``shape-from-shading'' (SFS) problem. SFS is traditionally defined as: recovering the shape of an object given a single image of it, assuming illumination and reflectance are known (or assuming reflectance is uniform across the entire image). This problem formulation is very complimentary to the shape-vs-reflectance version of the ``intrinsic images'' problem, as it focuses on the parts of the problem which ``intrinsic images'' ignores, and vice-versa.

The shape-from-shading problem was first formulated in the computer vision community by Horn in 1975 \cite{horn75}, though the problem existed in other fields as that of ``photoclinometry'' \cite{Rindfleisch}. The history of SFS is well surveyed in \cite{horntext,zhang2002shape}. Despite being a severe simplification of the complete intrinsic images problem, SFS is still a very ill-posed and underconstrained, and challenging problem. One~notable difficulty in SFS is the Bas-relief ambiguity \cite{belhumeur1999bas}, which  states (roughly) that the absolute orientation and scaling of a surface is ambiguous given only shading information.  This ambiguity holds true not only for SFS algorithms, but for human vision as well \cite{koenderinkConstancy}.  We address this ambiguity by imposing priors on shape, building on notions of ``smoothness'' priors in SFS \cite{ikeuchi1981numerical}, and by optionally allowing for external observations of shape (such as those produced by a stereo system or depth sensor) to be introduced.

Our model can be viewed as a generalization of an ``intrinsic image'' algorithm or color constancy algorithm in which shading is explicitly parametrized as a function of shape and illumination. Similarly, our model can be viewed as a shape-from-shading algorithm in which reflectance and illumination are unknown, and are recovered. Our model therefore addresses the ``complete'' intrinsic images problem, as it was first formulated. By addressing the complete problem, rather than two sub-problems in isolation, we outperform all previous algorithms for either subproblem. This is consistent with our understanding of human perception, as humans use spatial cues when estimating reflectance and shading~\cite{gilchrist,Boyaci2006}.

Because the intrinsic images problem is so challenging given only a single image, a much more popular area of research in computer vision has been to introduce additional data to better constrain the problem. Instances of this approach are photometric stereo \cite{woodham1980photometric}, which use additional images with different illumination conditions to estimate shape, and in later work reflectance and illumination \cite{basri2007photometric}. Our algorithm produces the same kinds of output as the most advanced photometric stereo algorithm, while requiring only a single image. ``Structure from motion'' or binocular stereo \cite{hartley2003multiple,Bundle} uses multiple images to recover shape, but ignores shading, reflectance, and illumination. Inverse global illumination \cite{yu1999inverse} recovers reflectance and illumination given shape and multiple images, while we recover shape and require only a single image.

Recent work has explored using learning to directly infer the spatial layout of a scene from a single image \cite{Hoiem_2007_5818,saxena2008make3d}. These techniques ignore illumination and reflectance, and produce only a coarse estimate of shape.

A similar approach to our technique is that of category-specific morphable models \cite{blanz1999morphable} which, given a single image of a very specific kind of object (a face, usually), estimates shape, reflectance, and illumination. These techniques use extremely specific models (priors) of the objects being estimated, and therefore do not work for general objects, while our priors are general enough to be applicable on a wide variety  of objects: a single model learned on teabags and squirrels can be applied to images of coffee cups and turtles.

The driving force behind our model are our priors on shape, reflectance, and illumination. To construct these priors we build upon past work on natural image statistics, which has demonstrated that simple statistics govern local patches of natural images \cite{field1987relations,ruderman1994statistics,HuangImages}, and that these statistics can be used for denoising \cite{Portilla03imagedenoising}, inpainting \cite{FOE}, deblurring \cite{Fergus06}, etc. But these statistical regularities arise in natural images only because of \emph{the statistical regularities in the underlying worlds that produced those images}. The primary contribution of this work is extended these ideas from natural images to the world that produced that natural image, which is assumed to be composed of natural depth maps and natural reflectance images. There has been some study of the statistics of natural depth maps \cite{HuangRange}, reflectance images \cite{Romeiro:2010} and models of illumination \cite{Dror28092004}, but ours is the first to use these statistical observations for recovering all such intrinsic scene properties simultaneously.

\section{Problem Formulation}
\label{sec:SIRFS}

We call our problem formulation for recovering intrinsic scene properties from a single image of a (masked) object  ``shape, illumination, and reflectance from shading'', or ``SIRFS''.  SIRFS can be thought of as an extension of classic shape-from-shading models \cite{HornThesis} in which not only shape, but reflectance and illumination are unknown. Conversely, SIRFS can be framed as an ``intrinsic image'' technique for recovering shading and reflectance, in which shading is parametrized by a model of shape and illumination. The SIRFS problem formulation is:
\begin{eqnarray}
\displaystyle \underset{R, Z, L}{\operatorname{maximize}} && P(R)P(Z)P(L) \nonumber \\
\operatorname{subject\;to} && I = R + S(Z, L)
\label{Eq_SIRFS1}
\end{eqnarray}
Where $R$ is a log-reflectance image, $Z$ is a depth-map, and $L$ is a spherical-harmonic model of illumination \cite{ramamoorthi2001}. $Z$ and $R$ are ``images'' with the same dimensions as $I$, and $L$ is a vector parametrizing the illumination. $S(Z,L)$ is a ``rendering engine'' which linearizes $Z$ into a set of surface normals, and produces a log-shading image from those surface normals and $L$ (see Appendix A for a thorough explanation).  $P(R)$, $P(Z)$, and $P(L)$ are priors on reflectance, shape, and illumination, respectively, whose likelihoods we wish to maximize subject to the constraint that the log-image $I$ is equal to a rendering of our model $R + S(Z,L)$. We can simplify this problem formulation by reformulating the maximum-likelihood aspect as minimizing a sum of cost functions (by taking the negative log of $P(R)P(Z)P(L)$) and by absorbing the constraint and removing $R$ as a free parameter. This gives us the following unconstrained optimization problem:
\begin{equation}
\displaystyle \underset{Z, L}{\operatorname{minimize}} \quad g(I - S(Z, L)) + f(Z) + h(L) 
\label{Eq_SIRFS2}
\end{equation}
where $g(R)$, $f(Z)$, and $h(L)$ (Sections~\ref{sec:reflectance}, \ref{sec:shape}, and \ref{sec:illumination}, respectively) are cost functions for reflectance, shape, and illumination respectively, which we will refer to as our ``priors'' on these scene properties \footnote{Throughout this paper we use the term ``prior'' loosely. We refer to loss functions or regularizers on $Z$, $A$, and $L$ as ``priors''  because they often have an interpretation as the negative log-likelihood of some probability density function. We refer to minimizing entropy as a ``prior'', which is again an abuse of terminology. Our occluding contour ``prior'' and our external observation ``prior'' require first observing the silhouette of an object or some external observation of shape, respectively, and are therefore posteriors, not priors.}. Solving this problem (Section \ref{sec:optimization}) corresponds to searching for the least costly (or most likely) explanation $\{ Z, R, L \}$ for image $I$.

\section{ Priors on Reflectance }

\label{sec:reflectance}

Our prior on reflectance consists of three components: 1) An assumption of piecewise constancy, which we will model by minimizing the local variation of log-reflectance in a heavy-tailed fashion. 2) An assumption of parsimony of reflectance --- that the palette of colors with which an entire image was painted tends to be small --- which we model by minimizing the global entropy of log-reflectance. 3) An ``absolute'' prior on reflectance which prefers to paint the scene with some colors (white, gray, green, brown, etc) over others (absolute black, neon pink, etc), thereby addressing color constancy. Formally, our reflectance prior $g(A)$ is a weighted combination of three costs:
\begin{eqnarray}
\label{eq:APrior}
g(R) = \lambda_s g_s(R) + \lambda_e g_e(R) + \lambda_a g_a(R)
\end{eqnarray}
where $g_s(R)$ is our smoothness prior, $g_e(R)$ is our parsimony prior, and $g_a(R)$ is our ``absolute'' prior. The $\lambda$ multipliers are learned through cross-validation on the training set.

Our smoothness and parsimony priors are on the differences of log-reflectance, which makes them equivalent to priors on the ratios of reflectance. This makes intuitive sense, as reflectance is defined as a ratio of reflected light to incident light, but is also crucial to the success of our algorithm: Consider the reflectance-map $\rho$ implied by log-image $I$ and log-shading $S(Z,L)$, such that $\rho = \exp(I - S(Z,L))$. If we were to manipulate $Z$ or $L$ to increase $S(Z,L)$ by some constant $\alpha$ across the entire image, then $\rho$ would be divided by $\exp(\alpha)$ across the entire image, which would accordingly decrease the differences between pixels of $\rho$. Therefore, if we placed priors on the differences of reflectance it would be possible to trivially satisfy our priors by manipulating shape or illumination to increase the intensity of the shading image. However, in the log-reflectance case $R = I - S(Z,L)$,  increasing all of $S$ by $\alpha$ (increasing the brightness of the shading image) simply decreases all of $R$ by $\alpha$, and does not change the differences between log-reflectance values (it would, however, affect our absolute prior on reflectance). Priors on the differences of log-albedo are therefore invariant to scaling of illumination or shading, which means they behave similarly in well-lit regions as in shadowed regions, and cannot be trivially satisfied.

\subsection{Smoothness}

\begin{figure}[t!]
	\centering
	\subfigure[\scriptsize univariate/grayscale GSM]{
		\resizebox{\fourwidth}{!}{\includegraphics{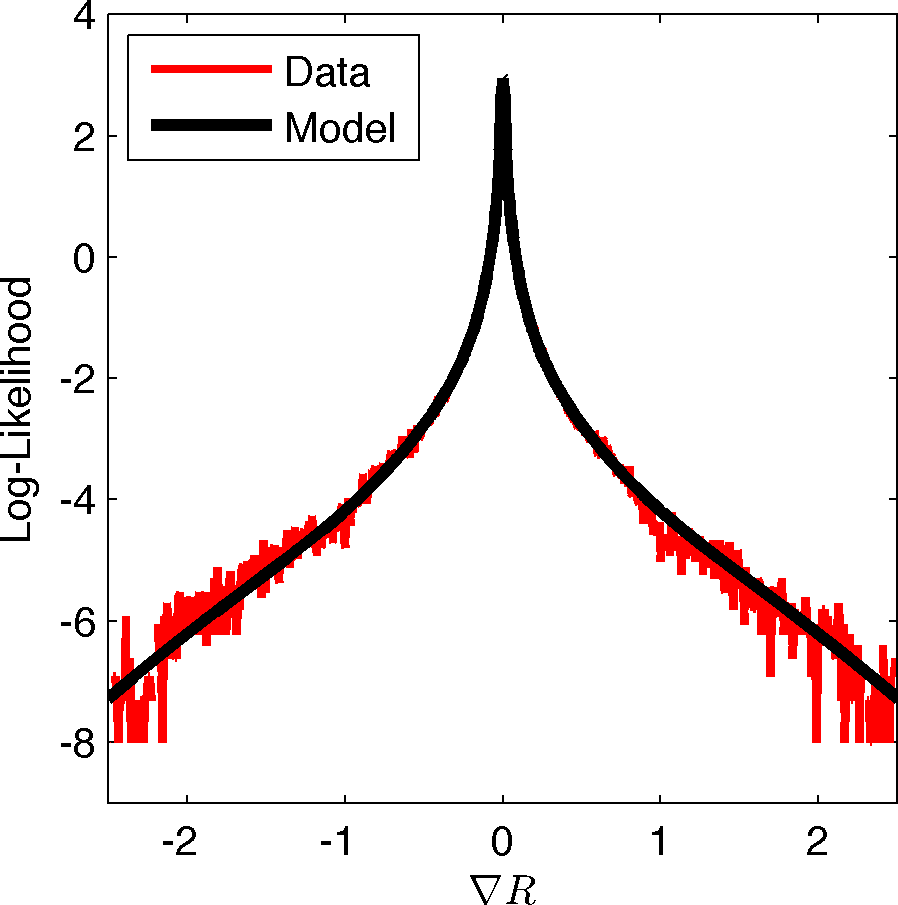}}
		\label{fig:ReflectanceGSM_gray}
		}
	\subfigure[\scriptsize multivariate/color GSM]{
		\resizebox{\fourwidth}{!}{\includegraphics{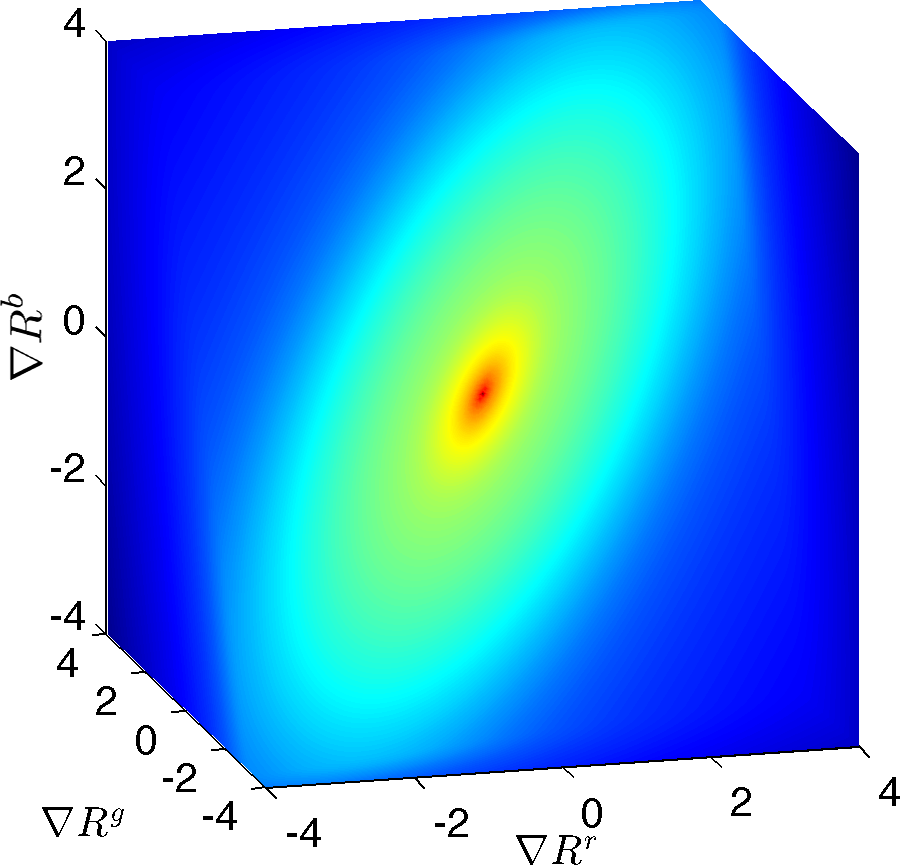}}
		\label{fig:ReflectanceGSM_color}
		}
	\caption{ 
	Our smoothness prior on log-reflectance is a univariate Gaussian scale mixture on the differences between nearby reflectance pixels for grayscale images, or a multivariate GSM for color images. These distribution prefers nearby reflectance pixels to be similar, but its heavy tails allow for rare non-smooth discontinuities. Our multivariate color model captures the correlation between color channels, which means that chromatic variation in log-reflectance lies further out in the tails, making it more likely to be ignored during inference.
	 \label{fig:ReflectanceGSM}}
\end{figure}

The reflectance images of natural objects tend to be piecewise constant  --- or equivalently, variation in reflectance images tends to be small and sparse. This is the insight that underlies the Retinex algorithm  \cite{grosse09intrinsic,Retinex,HornLightness}, and informs more recent intrinsic images work  \cite{ShenYJL11,Tappen,Gehler2011}. 

Our prior on grayscale reflectance smoothness is a multivariate Gaussian scale mixture (GSM) placed on the differences between each reflectance pixel and its neighbors. We will maximize the likelihood of $R$ under this model, which corresponds to minimizing the following cost function:
\begin{equation}
g_s(R) = \sum_i \sum_{j \in N(i)} c\left( R_i - R_j \, ;  \boldsymbol\alpha_R, \, \boldsymbol\sigma_R  \right)
\label{eq:Rsmooth1}
\end{equation}
Where $N(i)$ is the $5 \times 5$ neighborhood around pixel $i$,  $R_i - R_j$ is a the difference in log-RGB from pixel $i$ to pixel $j$, and $c\left( \cdot \, ;  \boldsymbol\alpha, \, \boldsymbol\sigma \right)$ is the negative log-likelihood of a discrete univariate Gaussian scale mixture (GSM), parametrized by $\boldsymbol\alpha$ and $\boldsymbol\sigma$, the mixing coefficients and standard deviations, respectively, of the Gaussians in the mixture:
\begin{equation}
c(x; \boldsymbol\alpha, \boldsymbol\sigma) = \displaystyle -\log \sum_{j = 1}^M \alpha_{j} \mathcal{N} \left( x \, ;  0, \sigma_{j}^2\right)
\label{eq:GSM}
\end{equation}
We set the mean of the GSM is $0$, as the most likely reflectance image under our model should be flat. We set $M = 40$ (the GSM has $40$ discrete Gaussians), and $\boldsymbol\alpha_R$ and $\boldsymbol\sigma_R$ are trained on reflectance images in our training set using expectation-maximization. The log-likelihood of our learned model can be seen in Figure~\ref{fig:ReflectanceGSM_gray}.

\begin{figure}[!]
	\centering
	\subfigure[\scriptsize some $R$]{
		\resizebox{\sixwidth}{!}{\includegraphics{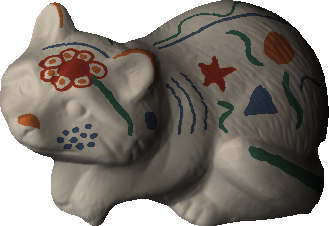}}
		\label{fig:ColorEdge_b}
		}
	\subfigure[\scriptsize $g_s(R)$ (cost)]{
		\resizebox{\sixwidth}{!}{\includegraphics{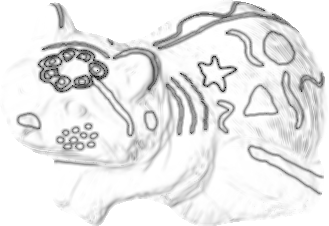}}
		}
	\subfigure[\scriptsize $\nabla g_s(R)$ (influence)]{
		\resizebox{\sixwidth}{!}{\includegraphics{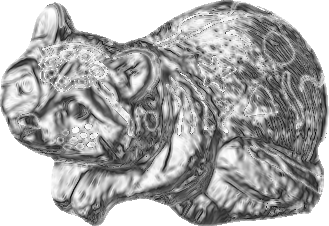}}
		}
	\caption{
	Here we have a color reflectance image $R$, and its cost and influence (derivative of cost) under our multivariate GSM smoothness prior. Strong, colorful edges, such as those caused by reflectance variation, are very costly, while small edges, such as those caused by shading, are less costly. But in terms of influence --- the gradient of cost with respect to each pixel --- we see an inversion: because sharp edges lie in the tails of the GSM, they have little influence, while shading variation has great influence. This means that during inference our model attempts to explain shading (small, achromatic variation) in the image by varying shape, while explaining sharp or chromatic variation by varying reflectance.
	 \label{fig:ColorEdge}}
\end{figure}

Gaussian scale mixtures have been used previously to model the heavy-tailed distributions found in natural images \cite{Portilla03imagedenoising}, for the purpose of denoising or inpainting. Effectively, using this family of distributions gives us a log-likelihood which looks like a smooth, heavy-tailed spline which decreases monotonically with distance from $0$. Because it is monotonically decreasing, the cost of log-reflectance variation increases with the magnitude of variation, but because the distribution is heavy tailed, the influence of variation (the derivative of log-likelihood) is strongest when variation is small (that is, when variation resembles shading) and weaker when variation is large. This means that our model prefers a reflectance image that is mostly flat but occasionally varies heavily, but abhors a reflectance image which is constantly varying slightly. This behavior is similar to that of the Retinex algorithm, which operates by shifting strong gradients to the reflectance image and weak gradients to the shading image.

To extend our model to color images, we simply extend our smoothness prior to a multivariate Gaussian scale mixture
\begin{equation}
g_s(R) = \sum_i \sum_{j \in N(i)} C\left( R_i - R_j \, ;  \boldsymbol\alpha_R, \, \boldsymbol\sigma_R, \mathrm{\Sigma_R}  \right)
\end{equation}
Where $R_i - R_j$ is now a $3$-vector of the log-RGB differences, $\boldsymbol\alpha$ are mixing coefficients, $\boldsymbol\sigma$ are the scalings of the Gaussians in the mixture, and $\mathrm{\Sigma}$ is the covariance matrix of the entire GSM (shared among all Gaussians of the mixture). 
\begin{eqnarray}
C(\mathbf{x} \, ; \boldsymbol\alpha, \boldsymbol\sigma, \mathrm\Sigma) = -\log \sum_{j=1}^M \alpha_j \, \mathcal{N} \left(  \mathbf{x} \, ;  \mathbf{0}, \sigma_j \, \mathrm{\Sigma}  \right)
\end{eqnarray}
We set $M = 40$ (the GSM has $40$ discrete Gaussians), and we train $\boldsymbol\alpha_R$, $\boldsymbol\sigma_R$, and $\Sigma_R$ on color reflectance images in our training set (we train a distinct model from the grayscale smoothness model). The log-likelihood of our learned model, and the training data used to learn that model, can be seen in Figure~\ref{fig:ReflectanceGSM_color}.

In color images, variation in reflectance tends to manifest itself in both the luminance and chrominance of an image (white transitioning to blue, for example) while shading, assuming the illumination is mostly white, primarily affects the luminance of an image (light blue transitioning to dark blue, for example).  Past work has exploited this insight by building specialized models that condition on the chrominance variation of the input image~\cite{grosse09intrinsic,HornLightness,ShenYJL11,Tappen,Gehler2011}.  By placing a multivariate prior over differences in reflectance, we are able to capture the correlation of the different color channels, which implicitly encourages our model to explain chromatic variation using reflectance and achromatic variation using shading without the need for any hand-crafted heuristics. See Figure~\ref{fig:ColorEdge} for a demonstration of this effect. Our model places more-colorful edges further into the tails of the distribution, thereby reducing their influence. Again, this is similar to color variants of the Retinex algorithm \cite{grosse09intrinsic} which uses the increased chrominance of an edge as a heuristic for it being a reflectance edge. But this approach (which is common among intrinsic image algorithms) of using image chrominance as a substitute for reflectance chrominance means that these techniques fail when faced with non-white illumination, while our model is robust to non-white illumination.

\subsection{Parsimony}
\label{sec_entropy}

\begin{figure}[!]
	\centering
	\begin{tabular}{ccc}
	\resizebox{\sixwidth}{!}{
		\includegraphics{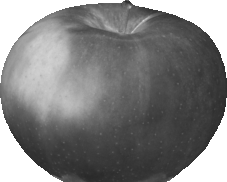}
	}
	&
	\resizebox{\sixwidth}{!}{
		\includegraphics{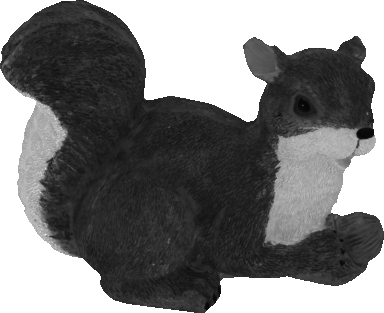}
	}
	&
	\resizebox{0.65in}{!}{
		\includegraphics{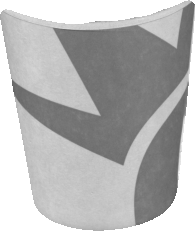}
	}\\
	\resizebox{\sixwidth}{!}{
		\includegraphics{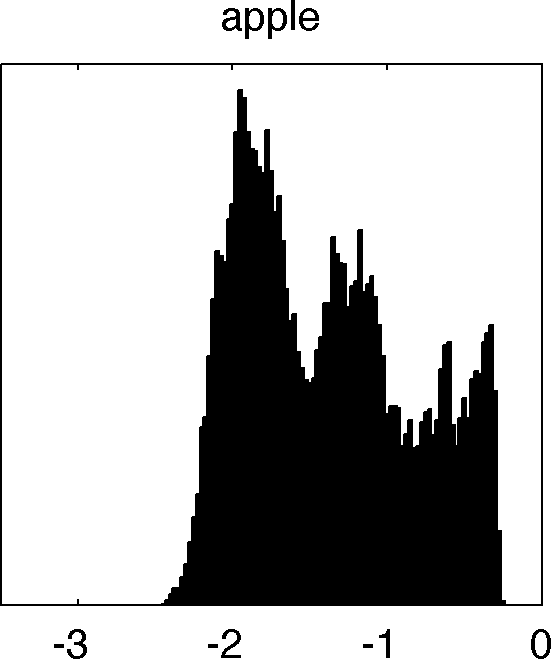}
	}
	&
	\resizebox{\sixwidth}{!}{
		\includegraphics{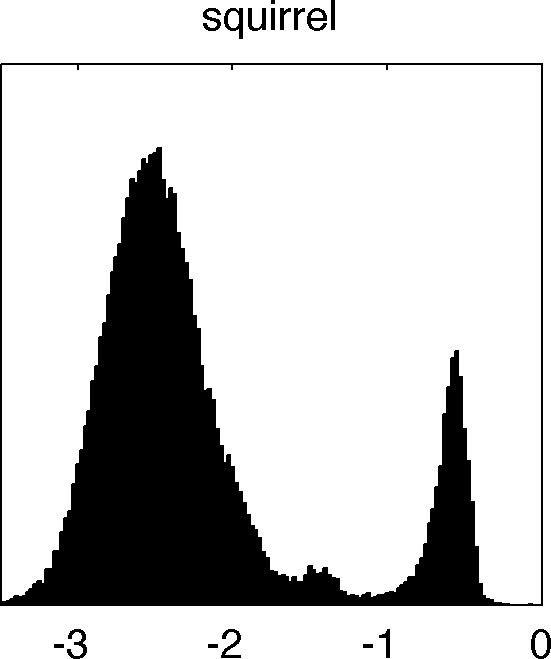}
	}
	&
	\resizebox{\sixwidth}{!}{
		\includegraphics{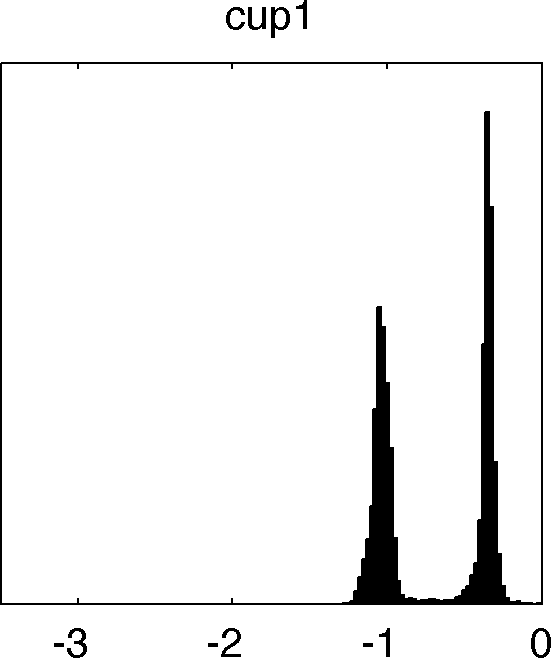}
	}
	\end{tabular}
	\caption{Three grayscale log-reflectance images from our dataset and their marginal distributions. Log-reflectance in an image tend to be grouped around certain values, or equivalently, these distributions tend to be low-entropy.
 \label{fig:Ahists}}
\end{figure}

In addition to piece-wise smoothness, the second property we expect from reflectance images is for there to be a small number of reflectances in an image --- that the palette with which an image was painted be small. As a hard constraint, this is not true: even in painted objects, there are small variations in reflectance. But as a soft constraint, this assumption holds. In Figure~\ref{fig:Ahists} we show the marginal distribution of grayscale log-reflectance for three objects in our dataset. Though the man-made "cup1" object shows the most clear peakedness in its distribution, natural objects like "apple" show significant clustering. 

We will therefore construct a prior which encourages parsimony -- that our representation of the reflectance of the scene be economical and efficient, or ``sparse''. This is effectively a instance of Occam's razor, that one should favor the simplest possible explanation. We are not the first to explore global parsimony priors on reflectance: 
different forms of this idea have been used in intrinsic images techniques~\cite{Gehler2011}, photometric stereo~\cite{alldrin2007resolving}, shadow removal~\cite{Finlayson:2009:EMS:1569442.1569449}, and color representation \cite{Omer2004}.  We use the quadratic entropy formulation of  \cite{Principe98learningfrom} to minimize the entropy of log-reflectance, thereby encouraging parsimony. Formally, our parsimony prior for reflectance is:
\begin{eqnarray}
g_e(R) &=& \displaystyle -\log \left( \frac{1}{Z} \sum_{i=1}^N \sum_{j=1}^N \exp\left( -\frac{(R_{i} - R_{j})^2}{4\sigma_R^2}\right) \right)  \nonumber \\
Z &=& N^2 \sqrt{4 \pi \sigma^2}
\label{eq:REPrior}
\end{eqnarray}
This is quadratic entropy (a special case of R\'enyi entropy) for a set of points $\mathbf{x}$ assuming a Parzen window (a Gaussian kernel density estimator, with a bandwidth of $\sigma_R$) \cite{Principe98learningfrom}. Effectively, this is a ``soft'' and differentiable generalization of Shannon entropy, computed on a set of real values rather than a discrete histogram. By minimizing this quantity, we encourage all pairs of reflectance pixels in the image to be similar to each other. However, minimizing this entropy does not force all pixels to collapse to one value, as the ``force'' exerted by each pair falls off exponentially with distance --- it is robust to outliers. This prior effectively encourages Gaussian ``clumps'' of reflectance values, where the Gaussian clumps have standard deviations of roughly $\sigma_R$.

\begin{figure}[!]
	\centering
	\subfigure[\scriptsize No parsimony]{
		\resizebox{\sixwidth}{!}{\includegraphics{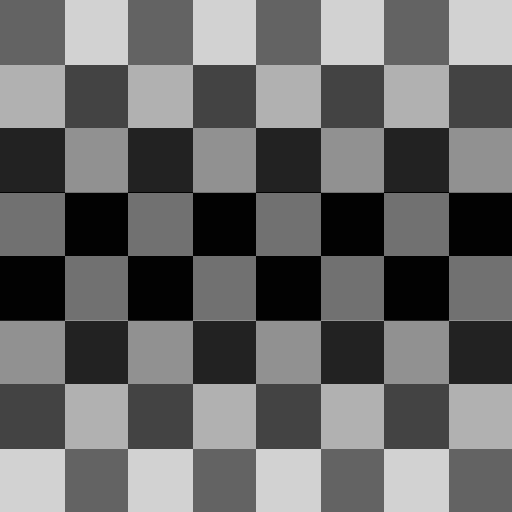}}
		\label{fig:checkerboard_fade}
		}
	\subfigure[\scriptsize No smoothness ]{
		\resizebox{\sixwidth}{!}{\includegraphics{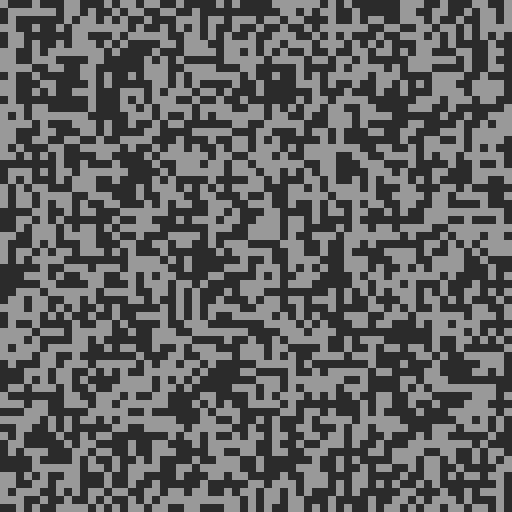}}
		\label{fig:checkerboard_shuffle}
		}
	\subfigure[\scriptsize Both ]{
		\resizebox{\sixwidth}{!}{\includegraphics{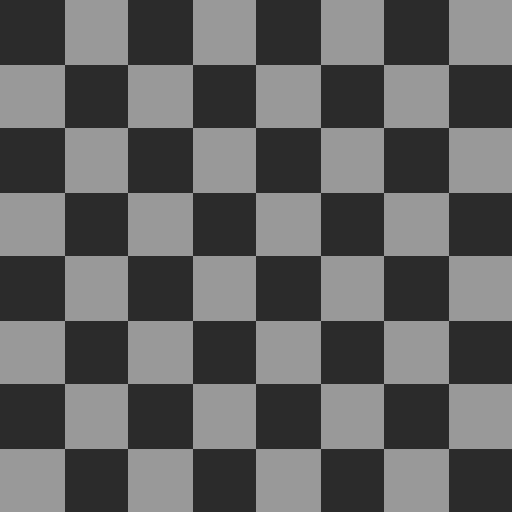}}
		\label{fig:checkerboard_normal}
		}
	\caption{A demonstration of the importance of both our smoothness and parsimony priors on reflectance. Using only a smoothness prior, as in  \ref{fig:checkerboard_fade}, allows for reflectance variation across disconnected regions. Using only the parsimony prior, as in  \ref{fig:checkerboard_shuffle}, encourages reflectance to take on a small number of values, but does not encourage it to form large piecewise-constant regions. Only by using the two priors in conjunction, as in \ref{fig:checkerboard_normal}, does our model correctly favor a normal, paint-like checkerboard configuration.
 \label{fig:checkerboard}}
\end{figure}

At first glance, it may seem that this global parsimony prior is redundant with our local smoothness prior: Encouraging piecewise smoothness seems like it should cause entropy to be minimized indirectly. This is often true, but there are common situations in which both of these priors are necessary. For example, if two regions are separated by a discontinuity in the image then optimizing for local smoothness will never cause the reflectance on both sides of the discontinuity to be similar. Conversely, simply minimizing global entropy may force reflectance to take on a small number of values, but need not produce large piecewise-smooth regions. The merit of using both priors in conjunction is demonstrated in Figure~\ref{fig:checkerboard}.

\begin{figure}[b!]
	\centering
	\subfigure[\scriptsize Correct ]{
		\begin{tabular}{c}
		\resizebox{0.75in}{!}{\includegraphics{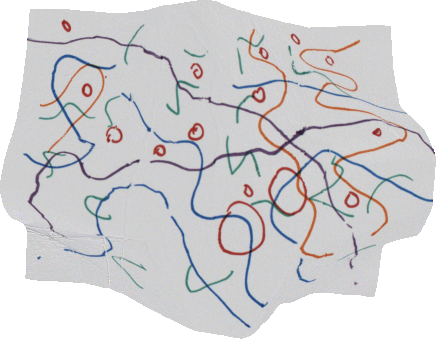}} \\
		\resizebox{0.8in}{!}{\includegraphics{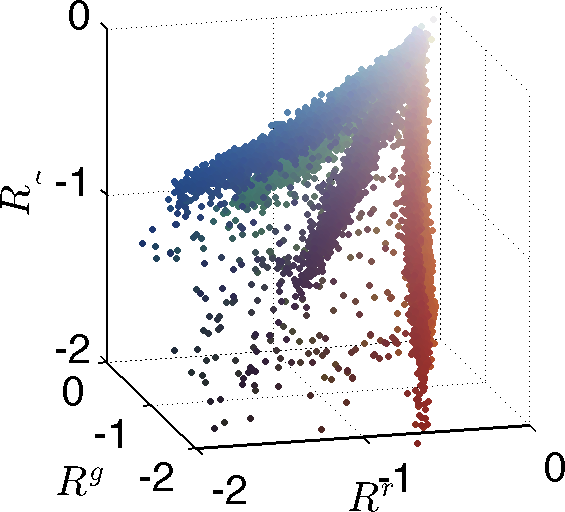}}
		\end{tabular}
		}
	\subfigure[\scriptsize Wrong Shape]{
		\begin{tabular}{c}
		\resizebox{0.75in}{!}{\includegraphics{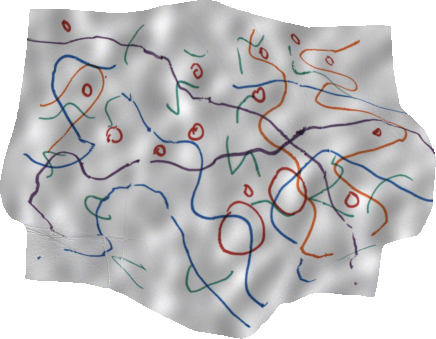}} \\
		\resizebox{0.8in}{!}{\includegraphics{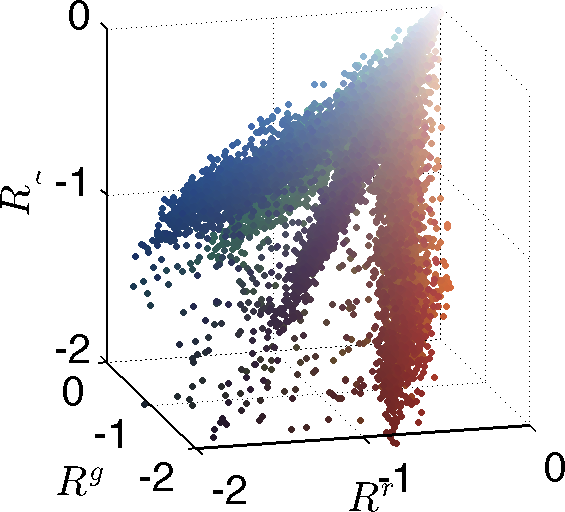}}
		\end{tabular}
		}
	\subfigure[\scriptsize Wrong Light]{
		\begin{tabular}{c}
		\resizebox{0.75in}{!}{\includegraphics{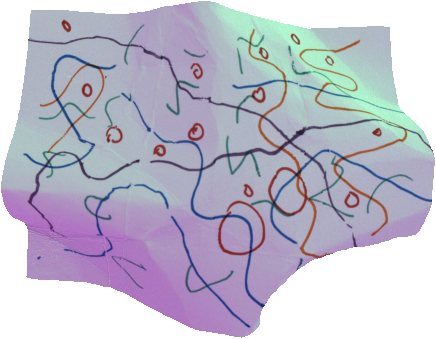}} \\
		\resizebox{0.8in}{!}{\includegraphics{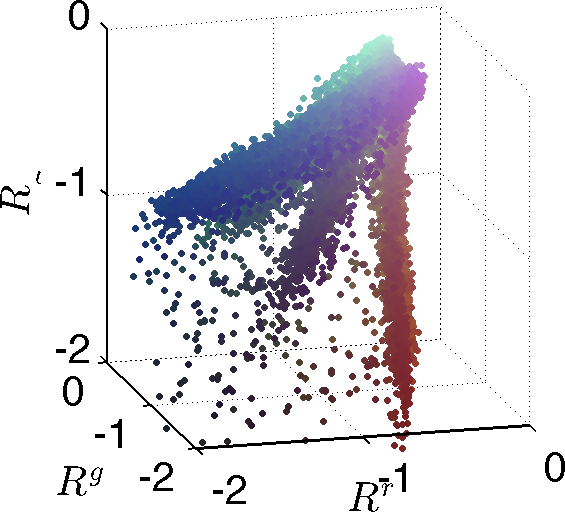}}
		\end{tabular}
		}
	\caption{ Some reflectance images and their corresponding log-RGB scatterplots. Mistakes in estimating shape or illumination produce shading-like or illumination-like errors in the inferred reflectance, causing the log-RGB distribution of the reflectance to be ``smeared'', and causing entropy (and therefore cost) to increase.
	 \label{fig:ColorEntropySmear}}
\end{figure}

Generalizing our grayscale parsimony prior to color reflectance images requires generalizing our entropy model to higher dimensionalities. A naive extension of this one-dimensional entropy model to three dimensions is not sufficient for our purposes: The RGB channels of natural reflectance images are highly correlated, causing a naive ``isotropic'' high-dimensional entropy measure to work poorly. To address this, we pre-compute a whitening transformation from log-reflectance images in the training set, and compute an isotropic entropy measure in this whitened space during inference, which gives us an anisotropic entropy measure.  Formally, our cost function is quadratic entropy in the space of whitened log-reflectance:
\begin{equation}
g_e(R) \!\! =\ \!\!  -\log\left( \frac{1}{Z}  \sum_{i=1}^N \sum_{j=1}^N \exp\left( - \frac{\norm{\mathrm{W_R} ( R_{i} - R_{j} )}_2^2}{4\sigma_R^2}   \right) \right)
\label{eq:REPrior_color}
\end{equation}
Where $\mathrm{W_R}$ is the whitening transformation learned from reflectance images in our training set, as follows: Let $\mathrm{X}$ be a $3 \times n$ matrix of the pixels in the reflectance images in our training set. We compute the matrix $\mathrm{\Sigma} = \mathrm{X} \mathrm{X}^\mathrm{T}$, take its eigenvalue decomposition $\mathrm{\Sigma} = \mathrm{\Phi} \mathrm{\Lambda} \mathrm{\Phi}^\mathrm{T}$, and from that construct the whitening\footnote{Our whitening transformation of reflectance is not strictly correct, as we do not first center the data by subtracting the mean. This was done both for mathematical and computational convenience, and because the origin of the space of log-reflectance (absolute white) is arguable the most reasonable choice for the ``center'' of our data.} transformation $\mathrm{W_R} = \mathrm{\Phi} \mathrm{\Lambda}^{1/2} \mathrm{\Phi}^\mathrm{T}$. The bandwidth of the Parzen window is $\sigma_R$, which determines the scale of the clusters produced by minimizing this entropy measure, and is tuned through cross-validation (independently of the same variable for the grayscale case). See Figure~\ref{fig:ColorEntropySmear} for a motivation of this model.

Naively computing this quadratic entropy measure requires calculating the difference between all $N$ log-reflectance values in the image with all other $N$ log-reflectance values, making it quadratically expensive in $N$ to compute naively. In Appendix B we describe an accurate linear-time algorithm for approximating this quadratic entropy and its gradient, based on the bilateral grid \cite{Chen2007}.

\subsection{Absolute Ref{}lectance}
\label{sec_absolute}

The previously described priors were imposed on \emph{relative} properties of reflectance: the differences between nearby or not-nearby pixels. We must impose an additional prior on \emph{absolute} reflectance: the raw value of each pixel in the reflectance image. Without such a prior (and the prior on illumination presented in Section~\ref{sec:illumination}) our model would be equally pleased to explain a gray pixel in the image as gray reflectance under gray illumination as it would nearly-black reflectance under extremely-bright illumination, or blue reflectance under yellow illumination, etc.

This sort of prior is fundamental to color-constancy, as most basic white-balance or auto-contrast/brightness algorithms can be viewed as minimizing a similar sort of cost: the gray-world assumption penalizes reflectance for being non-gray, the white-world assumption penalizes reflectance for being non-white, and gamut-based models penalize reflectance for lying outside of a gamut of previously-seen reflectances. We experimented with variations or combinations of these types of models, but found that what worked best was using a regularized smooth spline to model the log-likelihood of log-reflectance values.

\begin{figure}[t!]
	\centering
	\subfigure[\scriptsize Training data and PDF]{
		\resizebox{\fourwidth}{!}{\includegraphics{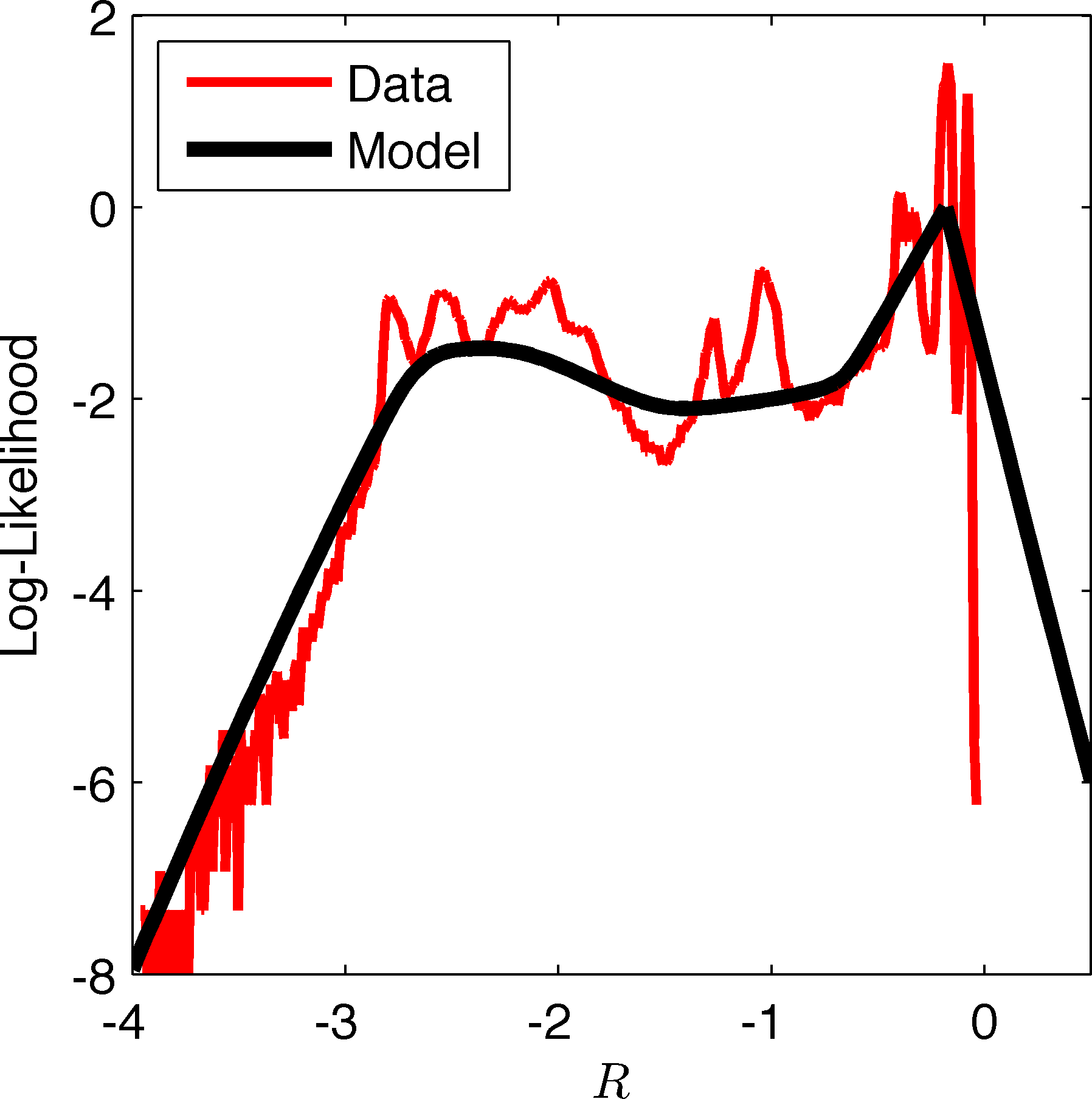}}
		 \label{fig:ReflectanceGray1}
	}
	\subfigure[\scriptsize Samples]{
		\resizebox{\fourwidth}{!}{\includegraphics{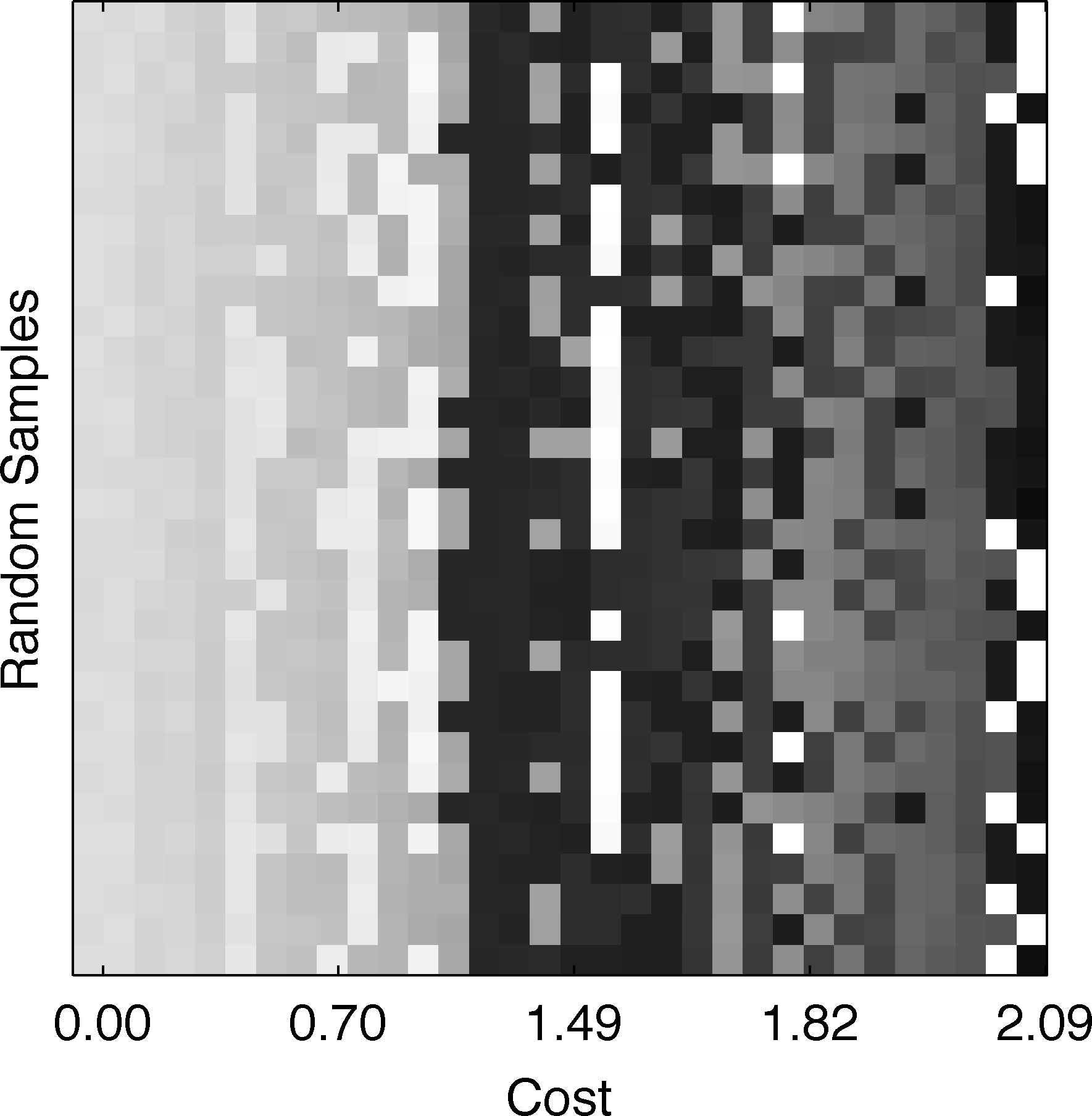}}
		 \label{fig:ReflectanceGray2}
	}
	\caption{A visualization of our ``absolute'' prior on grayscale reflectance, trained on the MIT Intrinsic Images dataset \cite{grosse09intrinsic}. In \ref{fig:ReflectanceGray1} we have the log-likelihood of our density model, and the data on which it was trained. In \ref{fig:ReflectanceGray2} we have samples from our model, where the $x$ axis is sorted by cost ($y$ axis is random).
	 \label{fig:ReflectanceGray}}
\end{figure}

For grayscale images, we use a 1D spline, which we have fit to log-reflectance images in the training set as follows:
\begin{equation}
\underset{\mathbf{f}}{\operatorname{minimize}} \quad \mathbf{f}^\mathrm{T} \mathbf{n}  + \log \left( \sum_{i}\exp \left( -\mathbf{f}_i  \right) \right)  +  \lambda\sqrt{ (\mathbf{f}^{''})^2 + \epsilon^2} 
\label{eq:GraySpline}
\end{equation}
Where $\mathbf{f}$ is our spline, which determines the non-normalized negative log-likelihood (cost) assigned to every reflectance, $\mathbf{n}$ is a 1D histogram of log-reflectance in our training data, and $\mathbf{f}^{''}$ is the second derivative of the spline, which we robustly penalize ($\epsilon$ is a small value added in to make our regularization differentiable everywhere). Minimizing the sum of the first two terms is equivalent to maximizing the likelihood of the training data (the second term is the log of the partition function for our density estimation), and minimizing the third term causes the spline to be piece-wise smooth. The smoothness multiplier $\lambda$ is tuned through cross-validation. A visualization of our prior can be found in Figure~\ref{fig:ReflectanceGray}.

During inference, we maximize the likelihood of the grayscale reflectance image $R$ by minimizing its cost under our learned model:
\begin{eqnarray}
g_a(R) = \sum_i \mathbf{f}(R_i) 
\end{eqnarray}
where $\mathbf{f}(R_i)$ is the value of $\mathbf{f}$ at $R_i$, the log-reflectance at pixel $i$, which we computed using linear interpolation (so that this cost is differentiable).

\begin{figure}[t!]
	\centering
	\subfigure[\scriptsize Training data]{
		\resizebox{\sixwidth}{!}{\includegraphics{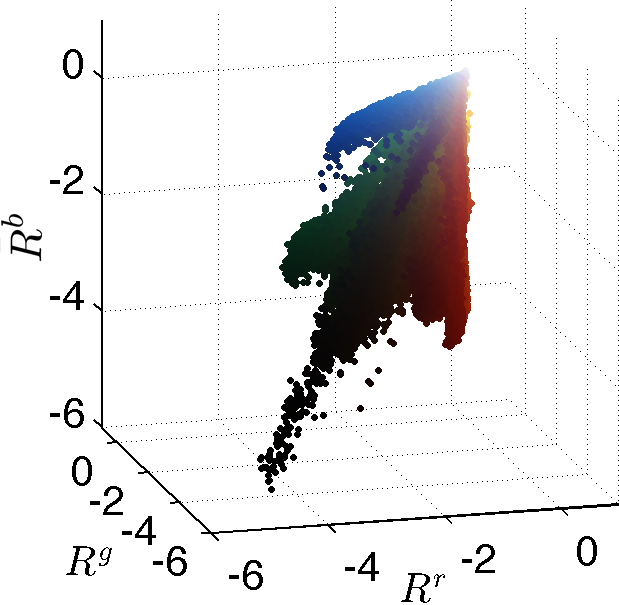}}
		 \label{fig:ReflectanceColor1}
	}
	\subfigure[\scriptsize PDF]{
		\resizebox{\sixwidth}{!}{\includegraphics{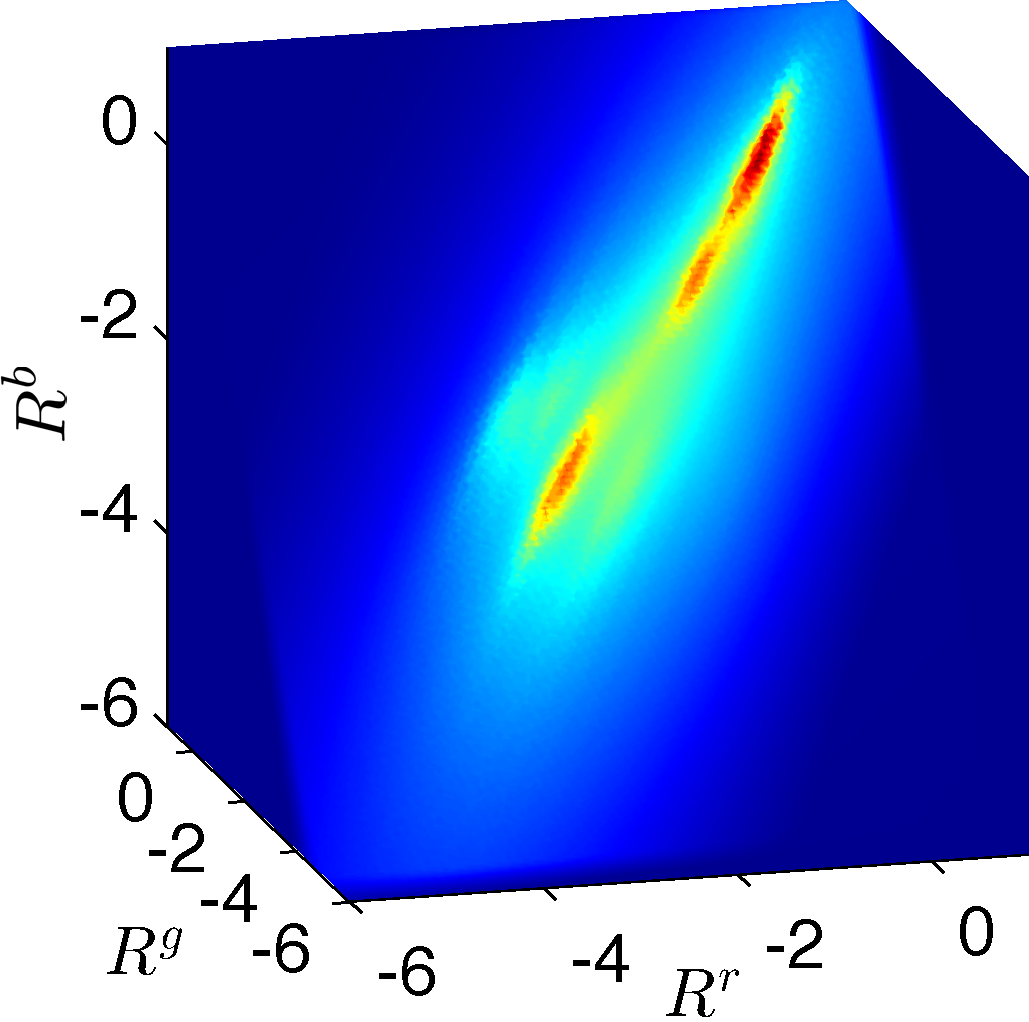}}
		 \label{fig:ReflectanceColor2}
	}
	\subfigure[\scriptsize Samples]{
		\resizebox{\sixwidth}{!}{\includegraphics{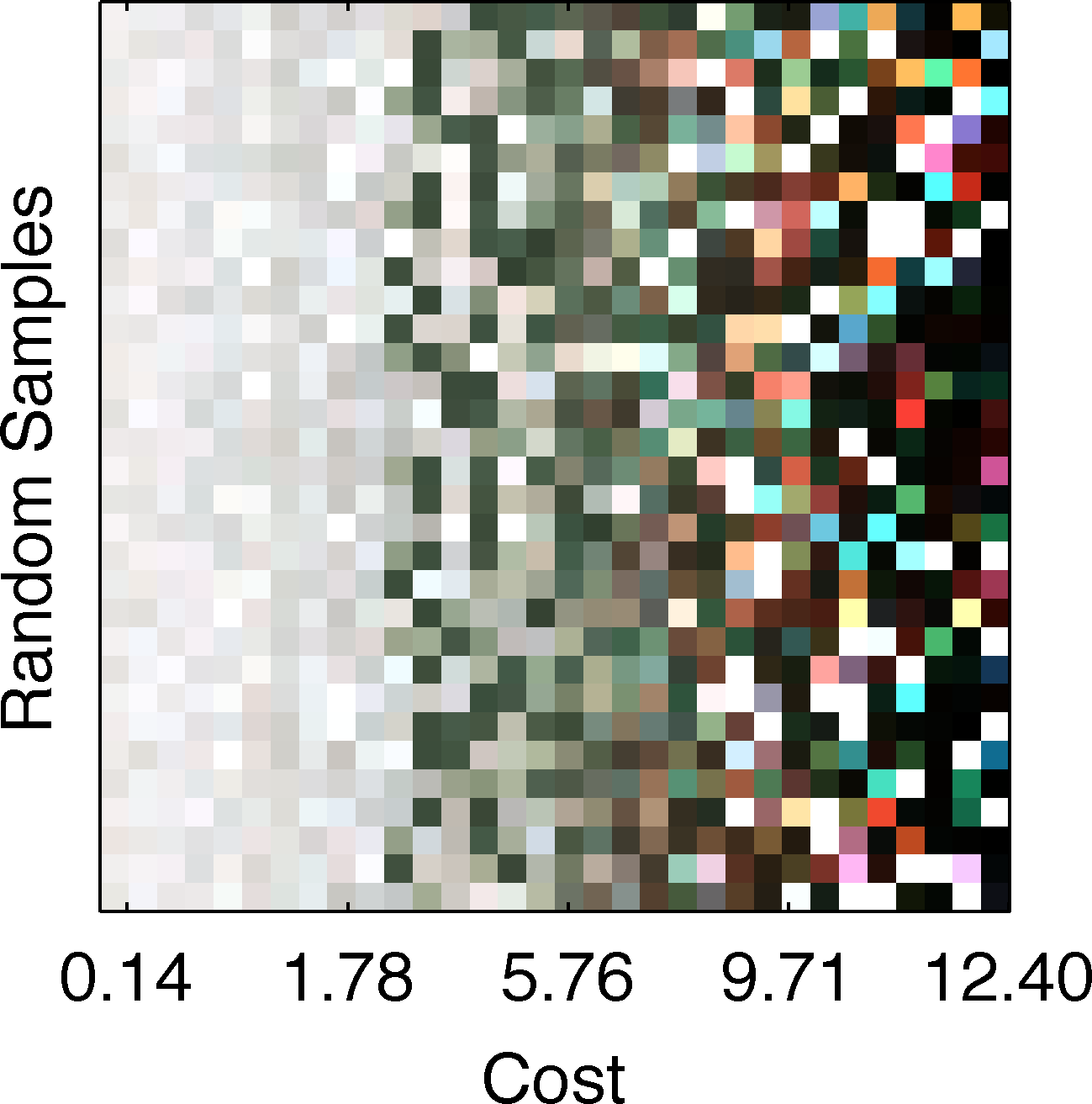}}
		 \label{fig:ReflectanceColor3}
	}
	\subfigure[\scriptsize Training data]{
		\resizebox{\sixwidth}{!}{\includegraphics{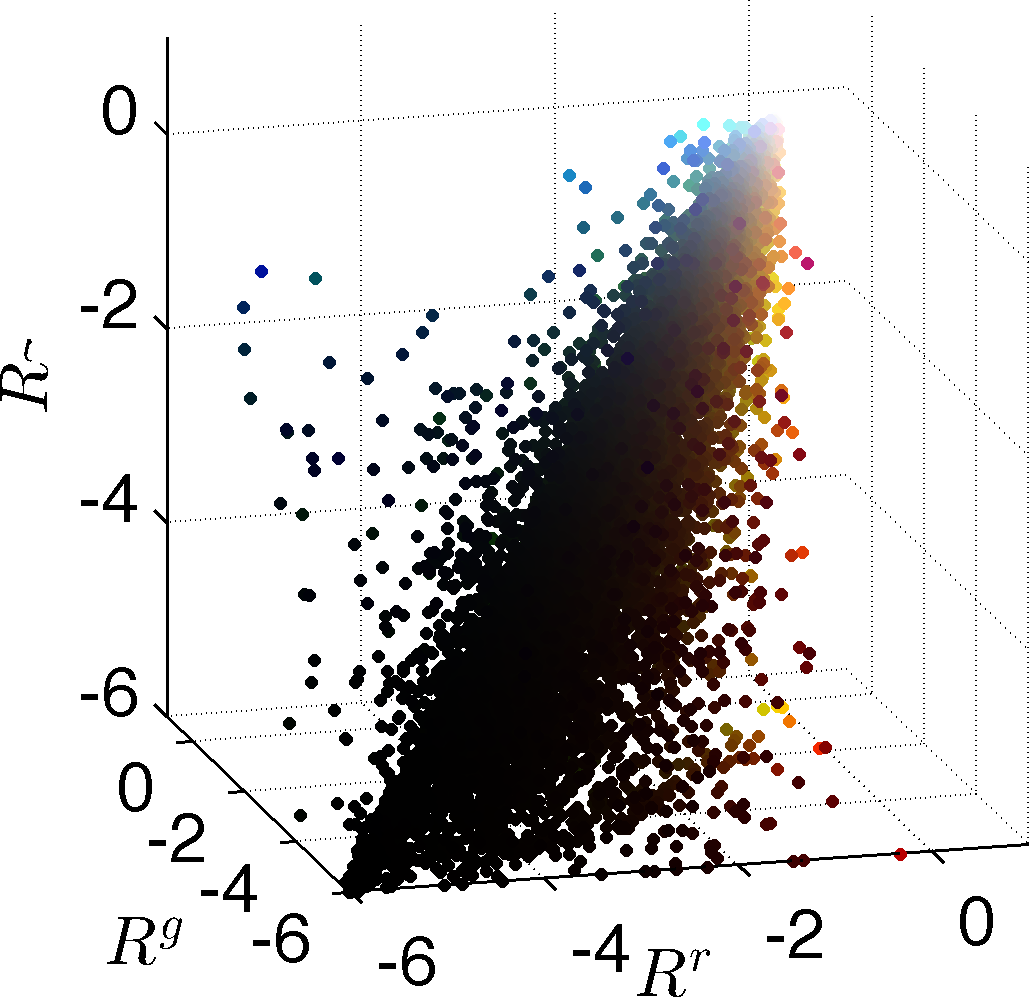}}
		 \label{fig:ReflectanceColor4}
	}
	\subfigure[\scriptsize PDF]{
		\resizebox{\sixwidth}{!}{\includegraphics{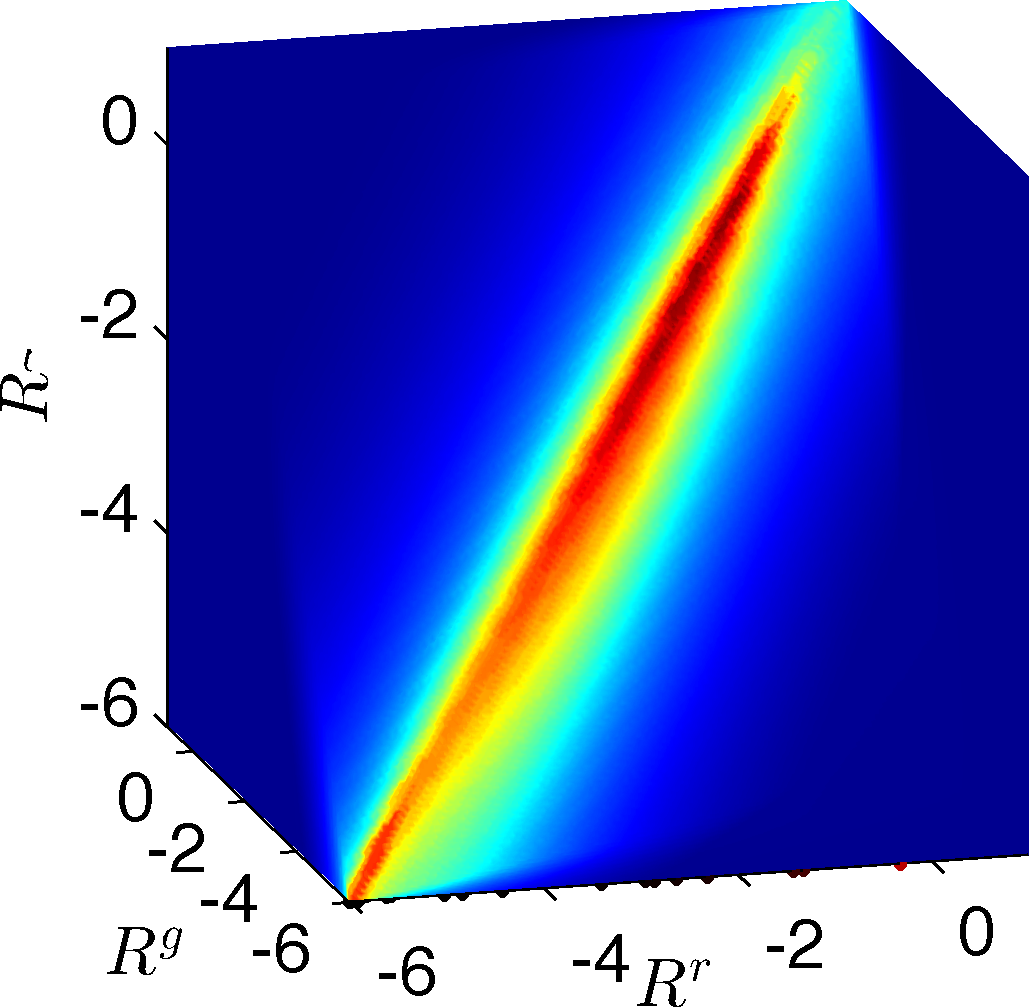}}
		 \label{fig:ReflectanceColor5}
	}
	\subfigure[\scriptsize Samples]{
		\resizebox{\sixwidth}{!}{\includegraphics{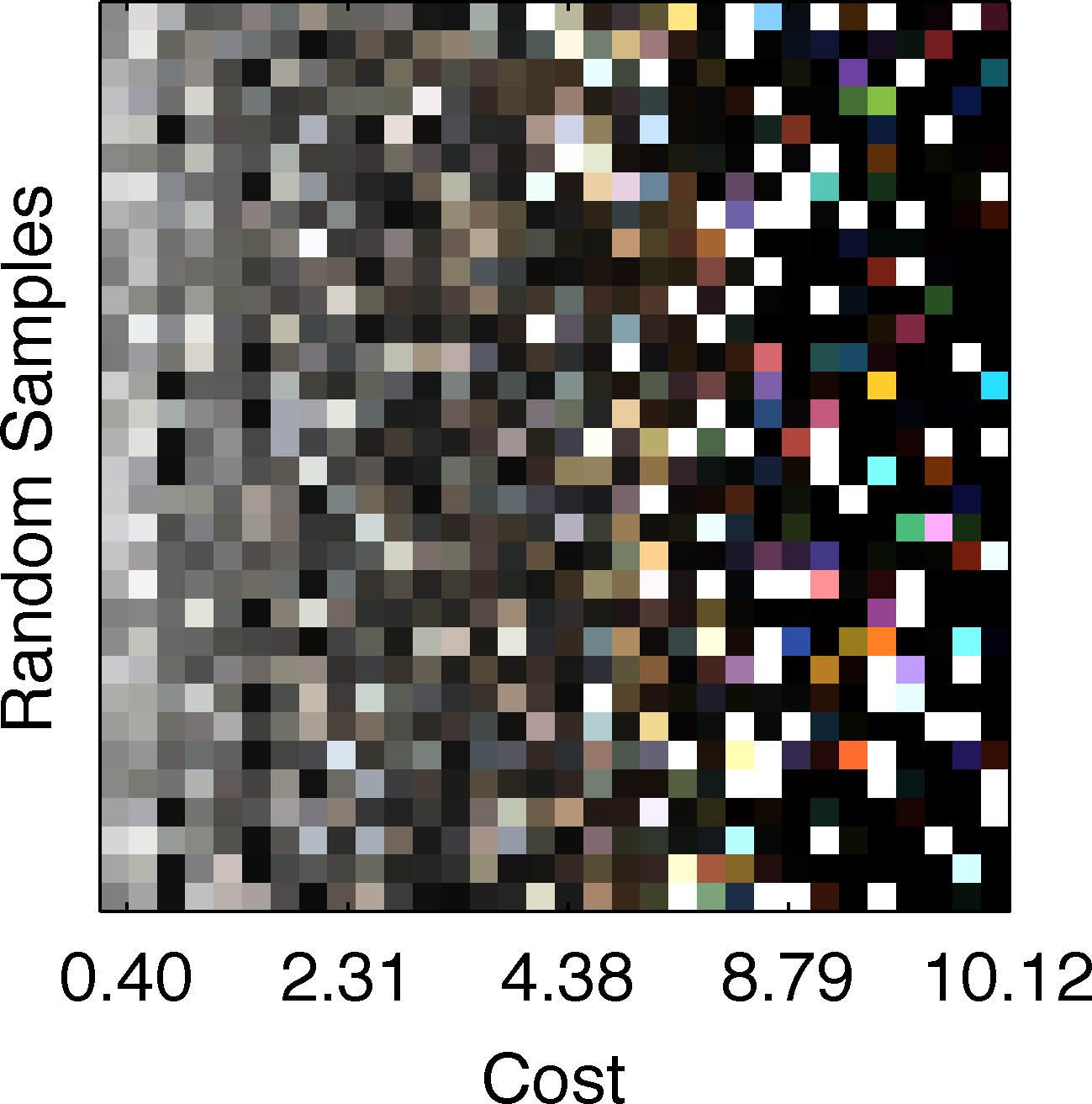}}
		 \label{fig:ReflectanceColor6}
	}
	\caption{A visualization of our ``absolute'' prior on color reflectance. We train two versions of our prior, one on the MIT Intrinsic Images dataset \cite{grosse09intrinsic} that we use in our experiments (top row) and one on the OpenSurfaces dataset for comparison \cite{bell13opensurfaces} (bottom row). In the first-column we have the log-RGB reflectance pixels in our training set, and in the second column we have a visualization of the 3D spline PDF that we fit to that data. In the third column we have samples from the PDF, where the $x$ axis is sorted by cost ($y$ axis is random). For both datasets, our model prefers less saturated, more earthy or subdued colors, and abhors brightly lit neon-like colors or very dark colors --- the high-cost reflectances often do not even look like paint, but instead appear glowing and luminescent. 
	 \label{fig:ReflectanceColor}}
\end{figure}

To generalize this model to color reflectance images, we simply use a 3D spline, trained on whitened log-reflectance pixels in our training set. Formally, to train our model we minimize the following:
\begin{equation}
\resizebox{3.3 in}{!}{$
\displaystyle \underset{\mathrm{F}}{\operatorname{minimize}} \, <\!\mathrm{F}, \mathrm{N}\!>   + \log \left( \sum_{i}\exp \left(-\mathrm{F}_{i} \right) \right) +  \lambda\sqrt{J( \mathrm{F})+ \epsilon^2}  \nonumber
$}\end{equation}
\begin{equation}
J(\mathrm{F}) = \mathrm{F}_{xx}^2 + \mathrm{F}_{yy}^2 + \mathrm{F}_{zz}^2 + 2 \mathrm{F}_{xy}^2 +  2 \mathrm{F}_{yz}^2 + 2 \mathrm{F}_{xz}^2 
\label{eq:ColorSpline}
\end{equation}
Where $\mathrm{F}$ is our 3D spline describing cost, $\mathrm{N}$ is a 3D histogram of the whitened log-RGB reflectance in our training data, and $J(\cdot)$ is a smoothness penalty (the thin-plate spline smoothness energy, made more robust by taking its square root). The smoothness multiplier $\lambda$ is tuned through cross-validation. As in our parsimony prior, we use whitened log-reflectance to address the correlation between channels, which is necessary as our smoothness term is isotropic. A visualization of our prior can be seen in Figure~\ref{fig:ReflectanceColor}.

During inference, we maximize the likelihood of the color reflectance image $R$ by minimizing its cost under our learned model:
\begin{eqnarray}
g_a(R) = \sum_i \mathrm{F}(\mathrm{W_R} R_i) 
\end{eqnarray}
where $\mathrm{F}(\mathrm{W_R} R_i)$ is the value of $\mathrm{F}$ at the coordinates specified by the 3-vector $\mathrm{W} R_i$, the whitened reflectance at pixel $i$ ($\mathrm{W_R}$ is the same as in Section~\ref{sec_entropy}). To make this function differentiable, we compute $\mathrm{F}(\cdot)$ using trilinear interpolation.

We trained our absolute color prior on the MIT Intrinsic Images dataset \cite{grosse09intrinsic}, and used that learned model in all experiments shown in this paper. However, the MIT dataset is very small and this absolute prior contains very many parameters (hundreds, in contrast to our other priors which are significantly more constrained), which suggests that we may be overfitting to the small set of reflectances in the MIT dataset. To address this concern, we trained an additional version of our absolute prior on the color reflectances in the OpenSurfaces dataset \cite{bell13opensurfaces}, which is a huge and varied dataset that is presumably a more accurate representation of real-world reflectances. Both models can be seen in Figure~\ref{fig:ReflectanceColor}, where we see that the priors we learn for each dataset are somewhat different, but that both prefer lighter, desaturated reflectances. We ran some additional experiments using our OpenSurfaces model instead of our MIT model (not presented in this paper), and found that the outputs of each model were virtually indistinguishable. This is a testament to the robustness of our model, and suggests that we are not overfitting to the color reflectances in the MIT dataset.

\section{ Priors on Shape }
\label{sec:shape}

Our prior on shape consists of three components: 1) An assumption of smoothness (that shapes tend to bend rarely), which we will model by minimizing the variation of mean curvature. 2) An assumption of isotropy of the orientation of surface normals (that shapes are just as likely to face in one direction as they are another) which reduces to a well-motivated ``fronto-parallel'' prior on shapes. 3) An prior on the orientation of the surface normal near the boundary of masked objects, as shapes tend to face outward at the occluding contour. Formally, our shape prior $f(Z)$ is a weighted combination of four costs:
\begin{equation}
f(Z) = \lambda_k f_k(Z) + \lambda_i f_i(Z)  + \lambda_c f_c(Z)
\end{equation}
where $f_k(Z)$ is our smoothness prior, $f_i(Z)$ is our isotropy prior, and $f_c$ is our bounding contour prior. The $\lambda$ multipliers are learned through cross-validation on the training set.

Most of our shape priors are imposed on intermediate representations of shape, such as mean curvature or surface normals. This requires that we compute these intermediate representations from a depth map, calculate the cost and the gradient of cost with respect to those intermediate representations, and backpropagate the gradients back onto the shape. In the appendix we explain in detail how to efficiently compute these quantities and backpropagate through them.

\subsection{ Smoothness}
\label{sec:zsmooth}

There has been much work on modeling the statistics of natural shapes  \cite{HuangRange,Woodford}, with one overarching theme being that regularizing some function of the second derivatives of a surface is effective. However, this past work has severe issues with invariance to out-of-plane rotation and scale. Working within differential geometry, we present a shape prior based on the variation of mean curvature, which allows us to place smoothness priors on $Z$ that are invariant to rotation and scale.

\begin{figure}[t!]
	\centering
	\subfigure[\scriptsize some shape $Z$]{
		\resizebox{\fourwidth}{!}{\includegraphics{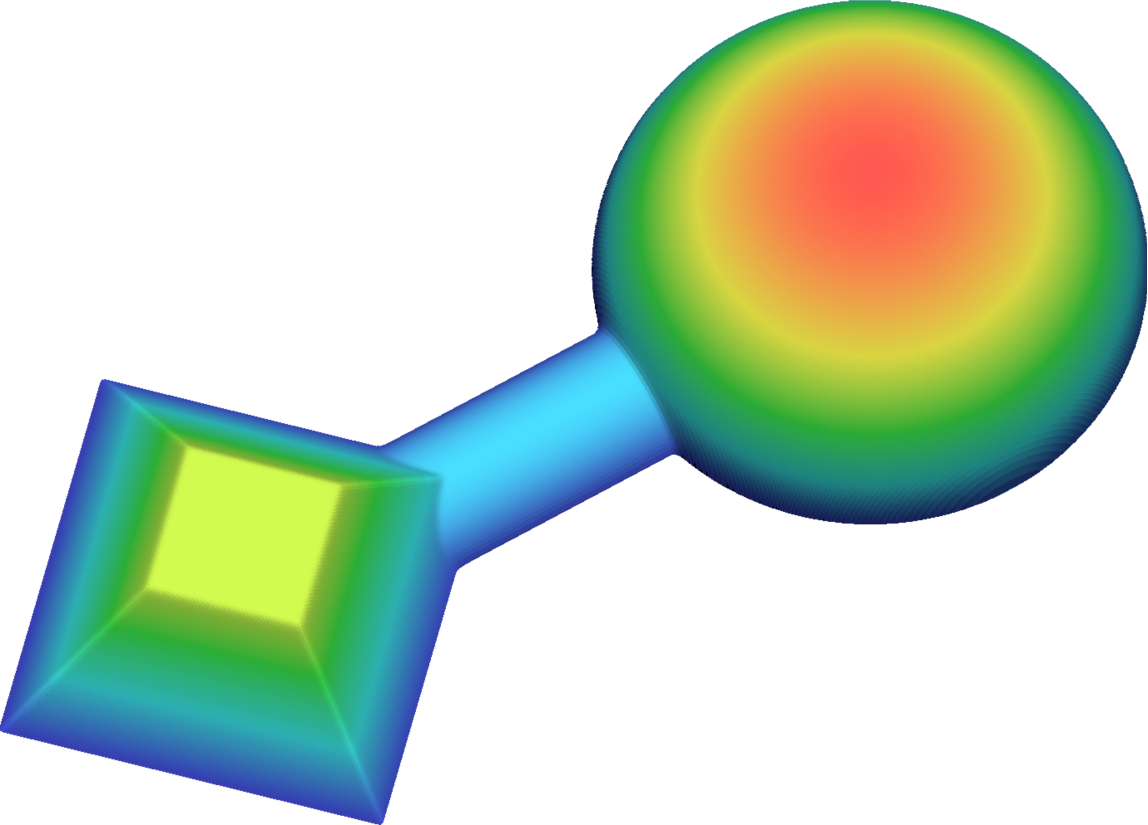}}
		 \label{fig:exampleKZ1}
	}
	\subfigure[\scriptsize mean curvature $H(Z)$]{
		\resizebox{\fourwidth}{!}{\includegraphics{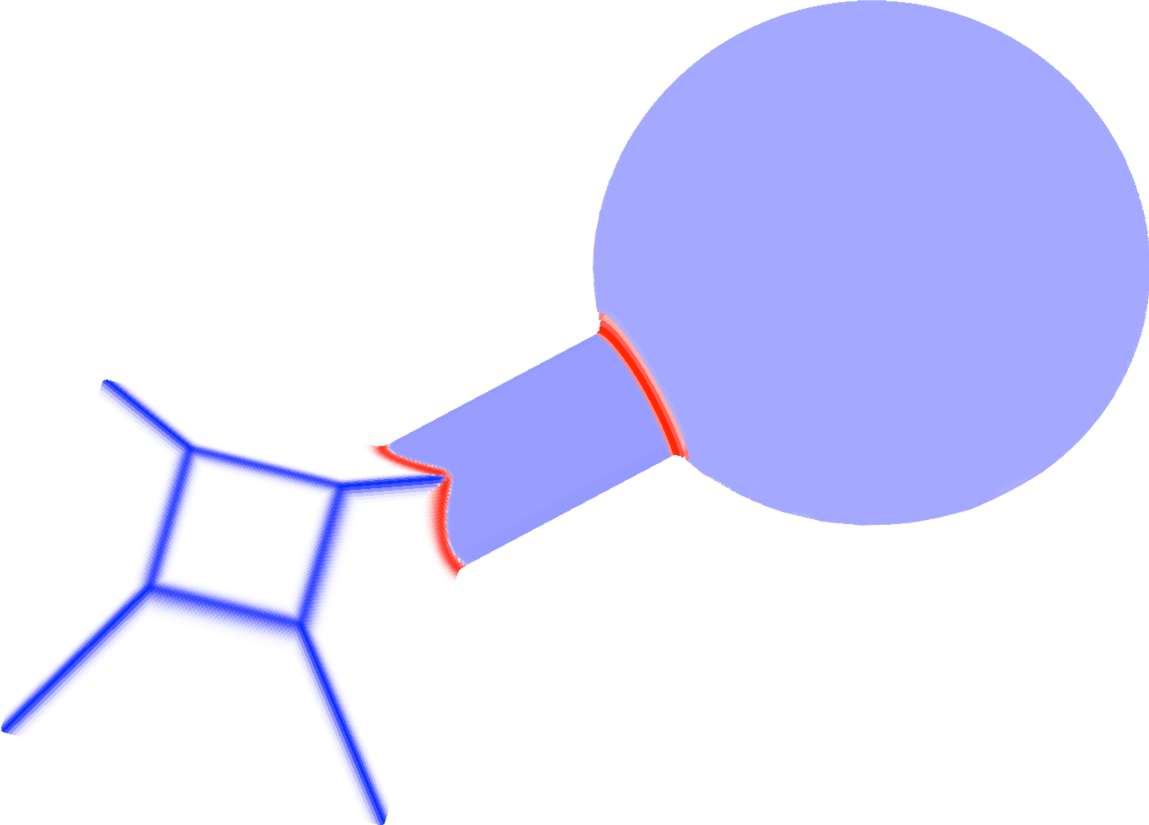}}
		 \label{fig:exampleKZ2}
	}
	\caption{A visualization of a shape and its mean curvature (blue = positive, red = negative, white = 0). Planes and soap films have $0$ mean curvature, spheres and cylinders have constant mean curvature, and mean curvature varies where shapes bend.
	 \label{fig:exampleKZ}}
\end{figure}

To review: mean curvature is the divergence of the normal field. Planes and soap films have $0$ mean curvature everywhere, spheres and cylinders have constant mean curvature everywhere, and the sphere has the smallest total mean curvature among all convex solids with a given surface area \cite{Hilbert56}. See Figure~\ref{fig:exampleKZ} for a visualization. Mean curvature is a measure of curvature in \emph{world coordinates}, not image coordinates, so (ignoring occlusion) the marginal distribution of $H(Z)$ is invariant to out-of-plane rotation of $Z$ --- a shape is just as likely viewed from one angle as from another.  In comparison, the Laplacian of $Z$ and the second partial derivatives of $Z$ can be made large simply due to foreshortening, which means that priors placed on these quantities \cite{Woodford} would prefer certain shapes simply due to the angle from which those shapes are observed --- clearly undesirable.

But priors on raw mean curvature are not scale-invariant. Were we to minimize $| H(Z) |$, then the most likely shape under our model would be a plane, while spheres would be unlikely. Were we to minimize $| H(Z) - \alpha|$ for some constant $\alpha$, then the most likely shape under our model would be a sphere of a certain radius, but larger or smaller spheres, or a resized image of the same sphere, would be unlikely. Clearly, such scale sensitivity is an undesirable property for a general-purpose prior on natural shapes. Inspired by previous work on minimum variation surfaces \cite{Moreton}, we place priors on the local variation of mean curvature. The most likely shapes under such priors are surfaces of constant mean curvature, which are well-studied in geometry and include soap bubbles and spheres of any size (including planes). Priors on the variation of mean curvature, like priors on raw mean curvature, are invariant to rotation and viewpoint, as well as concave/convex inversion.

\begin{figure}[t!]
	\centering
	\subfigure[\scriptsize Smoothness]{
		\resizebox{\fourwidth}{!}{\includegraphics{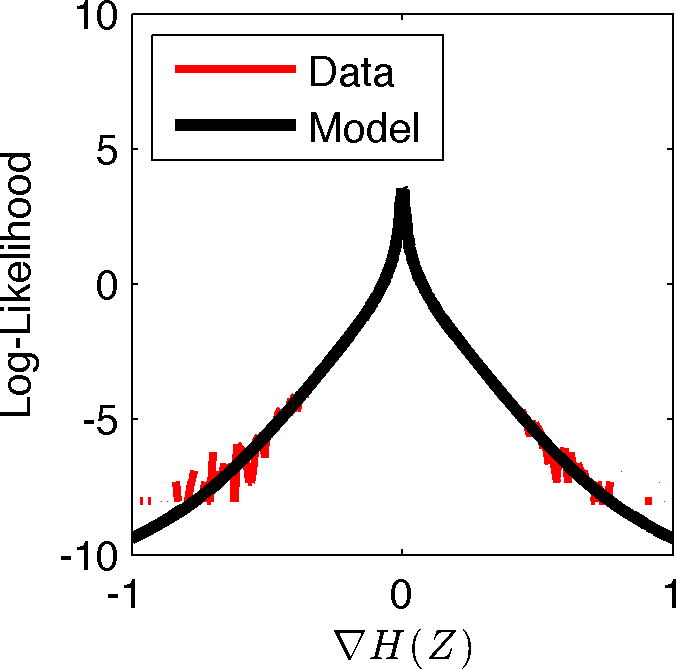}}
		 \label{fig:SmoothPDF}
	}
	\subfigure[\scriptsize Samples]{
		\resizebox{\fourwidth}{!}{\includegraphics{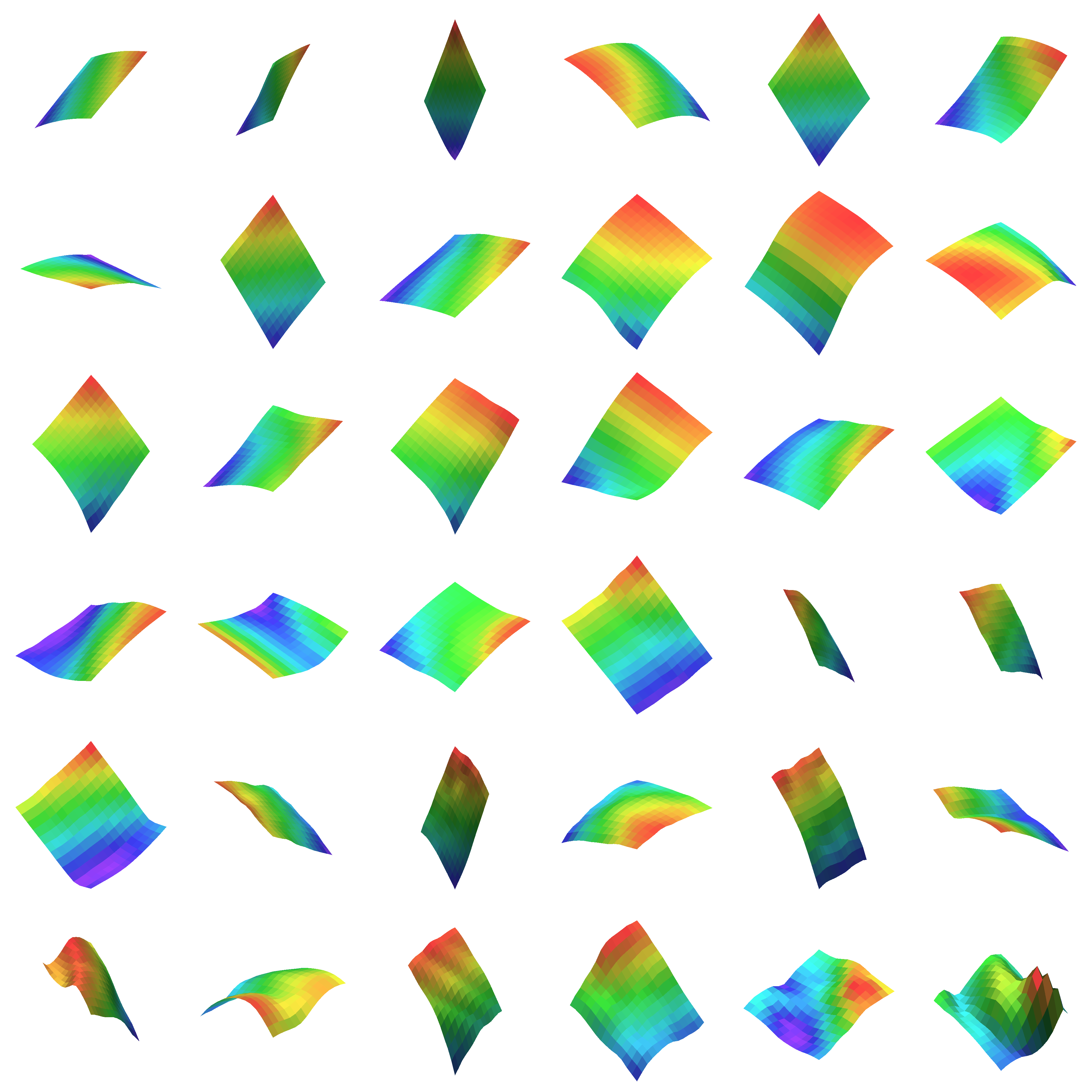}}
		 \label{fig:SmoothExamples}
	}
	\caption{To encourage shapes to be smooth, we model the variation in mean curvature of shapes using a Gaussian scale mixture, shown in \ref{fig:SmoothPDF}. In \ref{fig:SmoothExamples} we show patches of shapes in our training data, sorted from least costly (upper left) to most costly (lower right). Likely shapes under our model look like soap-bubbles, and unlikely shapes look contorted.
	 \label{fig:ShapeSmooth}}
\end{figure}

Mean curvature is defined as the average of principle curvatures: $H = \frac{1}{2}(\kappa_1 + \kappa_2)$. It can be approximated on a surface using filter convolutions that approximate first and second partial derivatives, as shown in \cite{BeslJain}.
\begin{equation}
H(Z)  = \frac{\left(1 + Z_x^2 \right) Z_{yy} -2 Z_x Z_y Z_{xy} + \left(1 + Z_y^2 \right) Z_{xx}}{2 \left(1 + Z_x^2 + Z_y^2 \right)^{3/2}}
\label{eq:HZ}
\end{equation}
In Appendix C we detail how to calculate and differentiate $H(Z)$ efficiently. Our smoothness prior for shapes is a Gaussian scale mixture on the local variation of the mean curvature of $Z$:
\begin{eqnarray}
f_k(Z) = \sum_i \sum_{j \in N(i)} c\left( H(Z)_i - H(Z)_j \, ;  \boldsymbol\alpha_k, \, \boldsymbol\sigma_k  \right)
\end{eqnarray}
Notation is similar to Equation~\ref{eq:Rsmooth1}: $N(i)$ is the $5 \times 5$ neighborhood around pixel $i$,  $H(Z)$ is the mean curvature of shape $Z$, and $H(Z)_i - H(Z)_j$ is the difference between the mean curvature at pixel $i$ and pixel $j$. $c\left( \cdot \, ;  \boldsymbol\alpha, \, \boldsymbol\sigma \right)$ is defined in Equation~\ref{eq:GSM}, and is the negative log-likelihood (cost) of a discrete univariate Gaussian scale mixture (GSM), parametrized by $\boldsymbol\alpha$ and $\boldsymbol\sigma$, the mixing coefficients and standard deviations, respectively, of the Gaussians in the mixture. The mean of the GSM is $0$, as the most likely shapes under our model should be smooth. We set $M = 40$ (the GSM has $40$ discrete Gaussians), and $\boldsymbol\alpha_k$ and $\boldsymbol\sigma_k$ are learned from our training set using expectation-maximization. The log-likelihood of our learned model can be seen in Figure~\ref{fig:SmoothPDF}, and the likelihoods it assigns to different shapes can be visualized in Figure~\ref{fig:SmoothExamples}. The learned GSM is very heavy tailed, which encourages shapes to be mostly smooth, and occasionally very non-smooth --- or equivalently, our prior encourages shapes to bend rarely. 

\subsection{Surface Isotropy}
\label{sec:isotropy}

Our second prior on shapes is motivated by the observation that shapes tend to be oriented isotropically in space. That is, it is equally likely for a surface to face in any direction. This assumption is not valid in many settings, such as man-made environments (which tend to be composed of floors, walls, and ceilings) or outdoor scenes (which are dominated by the ground-plane). But this assumption is more true for generic objects floating in space, which tend to resemble spheres (whose surface orientations are truly isotropic) or sphere-like shapes --- though there is often a bias on the part of photographers towards imaging the front-faces of objects. Despite its problems, this assumption is still effective and necessary.

Intuitively, one may assume that imposing this isotropy assumption requires no effort: if our prior assumes that all surface orientations are equally likely, doesn't that correspond to a constant cost for all surface orientations? However, this ignores the fact that once we have observed a surface in space, we have introduced a bias: observed surfaces are much more likely to face the observer ($N^z \approx 1$) than to be perpendicular to the observer ($N^z \approx 0$). We must therefore impose an isotropy prior to undo this bias.

\begin{figure}[t!]
	\centering
	\subfigure[\scriptsize An isotropic shape]{
		\resizebox{\fourwidth}{!}{\includegraphics{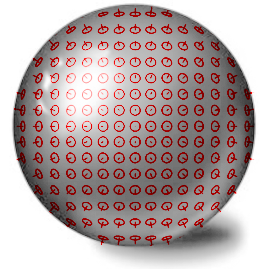}}
		\label{fig:ShapeIsotropy1}
	}
	\subfigure[\scriptsize Our isotropy prior]{
		\resizebox{\fourwidth}{!}{\includegraphics{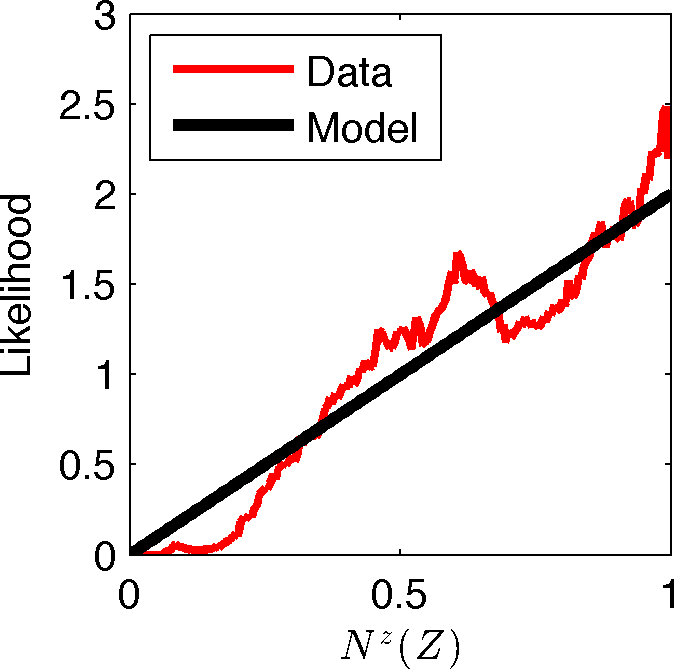}}
		\label{fig:ShapeIsotropy2}
	}
	\caption{We assume the surfaces of shapes to be isotropic --- equally likely to face in any orientation, like in a sphere. However, observing an isotropic shape imposes a bias, as observed surfaces are more likely to face the observer than to be perpendicular to the observer (as shown by the red gauge figure ``thumbtacks'' placed on the sphere in \ref{fig:ShapeIsotropy1}). We undo this bias by imposing a prior on $N^z$, shown in \ref{fig:ShapeIsotropy2}, which coarsely resembles our training data.
	 \label{fig:ShapeIsotropy}}
\end{figure}

We will derive our isotropy prior analytically. Assume surfaces are oriented uniformly, and that the surfaces are observed under orthogonal perspective with a view direction $(0, 0, -1)$. It follows that all $N^z$ (the $z$-component of surface normals, relative to the viewer) are distributed uniformly between 0 and 1.  Upon observation, these surfaces (which are assumed to have identical surface areas) have been foreshortened, such that the area of each surface in the image is $N^z$. Given the uniform distribution over $N^z$ and this foreshortening effect, the probability distribution over $N^z$ that we should expect at a given pixel in the image is proportional to $N^z$. Therefore, maximizing the likelihood of our uniform distribution over orientation in the world is equivalent to minimizing the following in the image:
\begin{eqnarray}
f_i(Z) = -\sum_{x,y} \log\left( N^z_{x,y}(Z) \right)
\end{eqnarray}
Where $N^z_{x,y}(Z)$ is the $z$-component of the surface normal of $Z$ at position $(x,y)$ (defined in Appendix A).

Though this was derived as an isotropy prior, the shape which maximizes the likelihood of this prior is not isotropic, but is instead (because of the nature of MAP estimation) a fronto-parallel plane. This gives us some insight into the behavior of this prior --- it serves to as a sort of ``fronto-parallel'' prior. This prior can therefore be thought of as combating the bas-relief ambiguity \cite{belhumeur1999bas} (roughly, that absolute scale and orientation are ambiguous), by biasing our shape estimation towards the fronto-parallel members of the bas-relief family.

Our prior on $N^z$ is shown in Figure~\ref{fig:ShapeIsotropy2} compared to the marginal distribution of $N^z$ in our training data. Our model fits the data well, but not perfectly. We experimented with learning distributions on $N^z$ empirically, but found that they worked poorly compared to our analytical prior. We attribute this to the aforementioned photographer's bias towards fronto-parallel surfaces, and to data sparsity when $N^z$ is close to $0$.

It is worth noting that $-\log\left(N^z\right)$ is proportional to the surface area of $Z$. Our prior on slant therefore has a helpful interpretation as a prior on minimal surface area: we wish to minimize the surface area of $Z$, where the degree of the penalty for increasing $Z$'s surface area happens to be motivated by an isotropy assumption. This notion of placing priors on surface area has been explored previously \cite{forsyth2011VSSA}, but not in the context of isotropy. And of course, this connection relates our model to the study of minimal surfaces in mathematics \cite{Hilbert56}, though this connection is somewhat tenuous as the fronto-parallel planes favored by our model are very different from classical minimal surfaces such as planes and soap films. 

\subsection{The Occluding Contour}
\label{sec:contours}

\begin{figure}[!]
	\centering
	\subfigure[\scriptsize A cropped object an its normals]{
		\resizebox{\fourwidth}{!}{\includegraphics{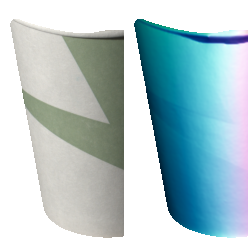}}
		 \label{fig:ShapeContour1}
	}
	\subfigure[\scriptsize Our occluding contour prior]{
		\resizebox{\fourwidth}{!}{\includegraphics{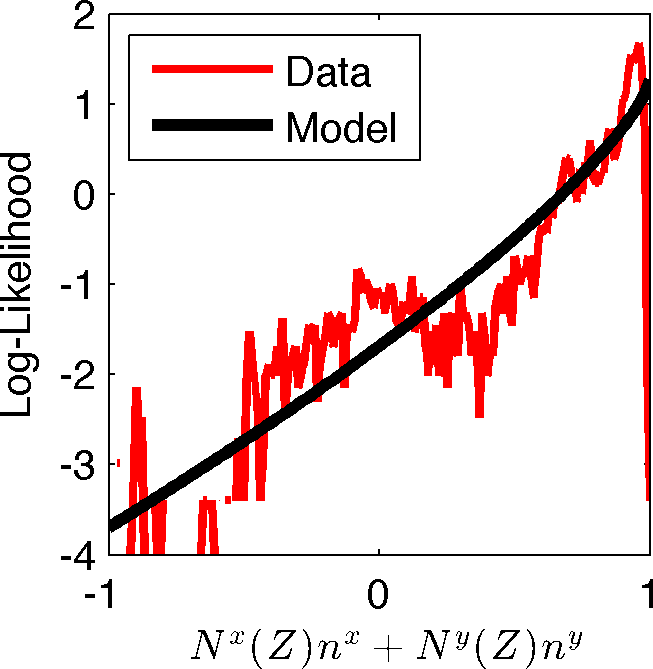}}
		 \label{fig:ShapeContour2}
	}
	\caption{In \ref{fig:ShapeContour1} we have an image and surface normals of a subset of a cup, in our dataset. The side of this cup are ``limbs'', points where the surface normal faces outward and is perpendicular to the occluding contour, while the top of the cup are ``edges'', sharp discontinuities where the surface is oriented arbitrarily. Our heavy-tailed prior over surface orientation at the occluding contour in \ref{fig:ShapeContour2} models the behavior of limbs, but is robust to the outliers caused by edges.
	 \label{fig:ShapeContour}}
\end{figure}

The occluding contour of a shape (the contour that surrounds the silhouette of a shape) is a powerful cue for shape interpretation \cite{koenderink1984does} which often dominates shading cues \cite{Mamassian}, and algorithms have been presented for coarsely estimating shape given contour information \cite{Brady83anextremum}. At the occluding contour of an object, the surface is tangent to all rays from the vantage point. Under orthographic projection (which we assume), this means the $z$-component of the normal is $0$, and the $x$ and $y$ components are determined by the contour in the image. In principle, this property is absolutely true, but in practice the occluding contour of a surface tends to be composed of limbs (points where the surface is tangent to rays from the vantage point, like the smooth side of a cylinder) and edges (an abrupt discontinuity of the surface, like the top of a cylinder or the edge of a piece of paper) \cite{malik1987interpreting}. See Figure~\ref{fig:ShapeContour1} for an example of a shape which contains both phenomena. Of course, this taxonomy is somewhat false --- all edges are limbs, but some are so small that they appear to be edges, and some are just small enough relative to the image resolution that the ``limb'' assumption begins to break down.

\begin{figure}[!]
	\centering
	\subfigure[\scriptsize occluding contour normals ]{
		\resizebox{\fourwidth}{!}{\includegraphics{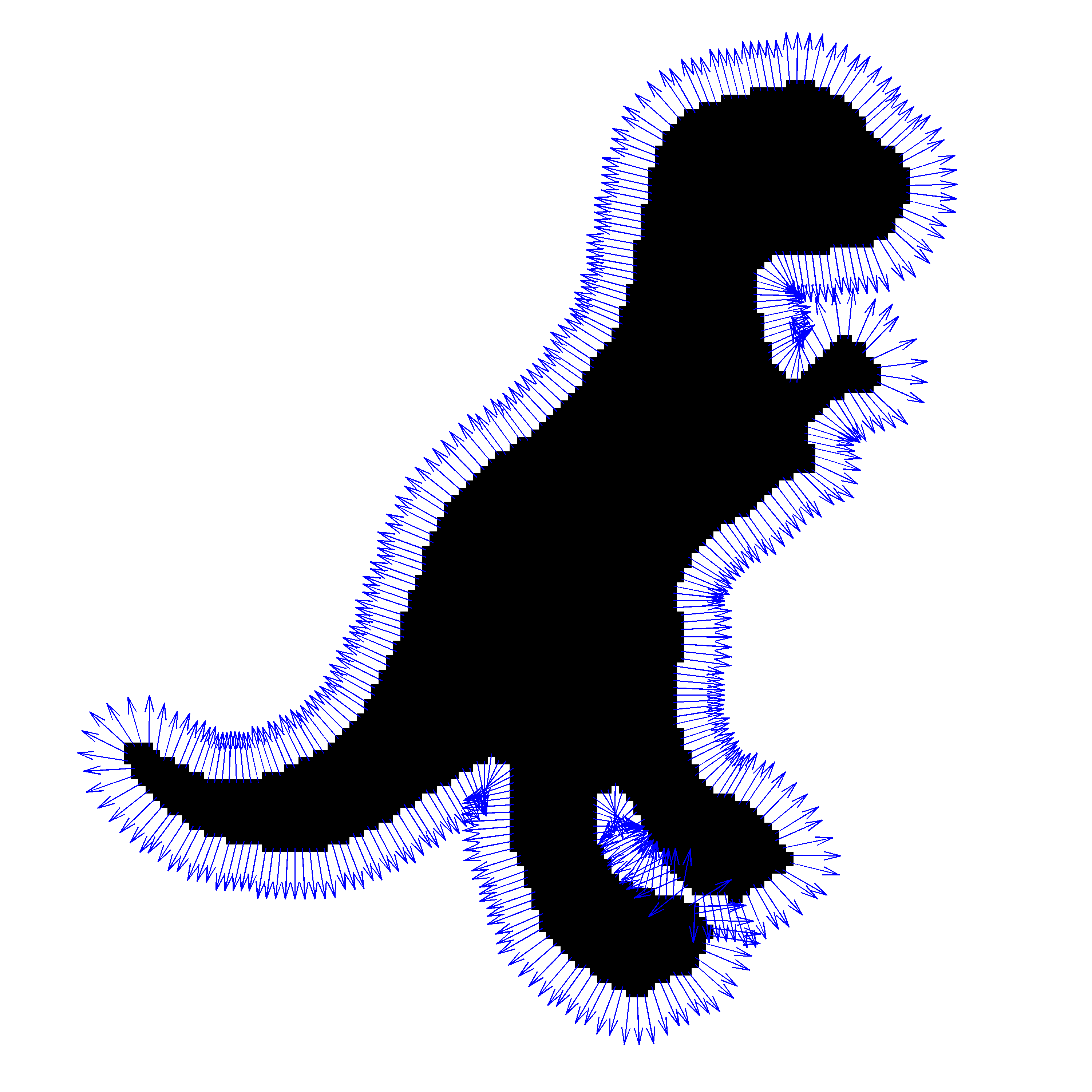}}
		 \label{fig:SFC1}
	}
	\subfigure[\scriptsize shape-from-contour output]{
		\resizebox{\fourwidth}{!}{\includegraphics{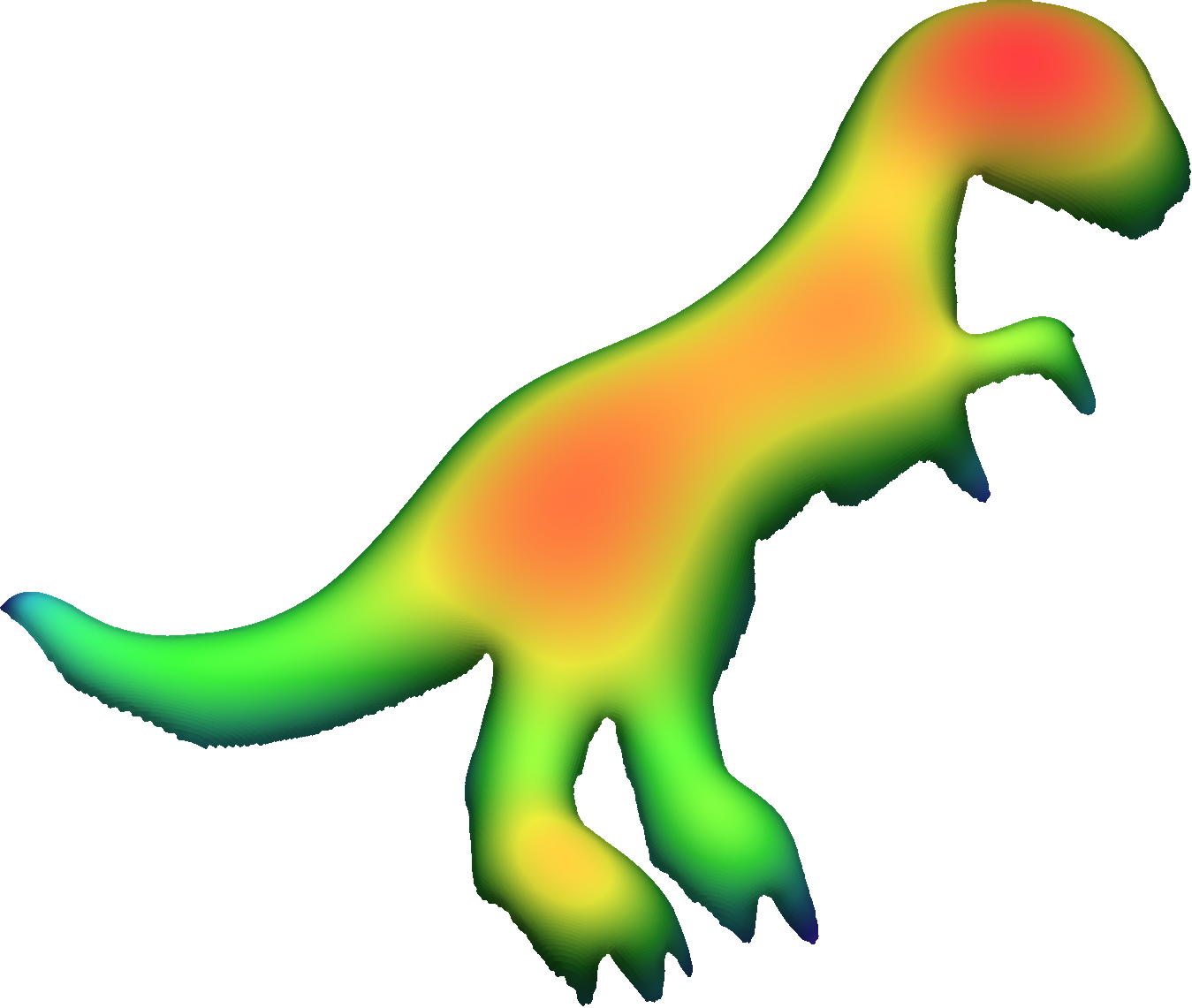}}
		 \label{fig:SFC2}
	}
	\caption{A subset of our model that includes only our priors on shape is equivalent to a shape-from-contour model. Given only the normals of the silhouette of the object in \ref{fig:SFC1}, we can produce the coarse estimate of the shape of the object in \ref{fig:SFC2}.
	 \label{fig:SFC}}
\end{figure}

We present a ``soft'' version of a limb constraint, one which captures the ``limb''-like behavior we expect to see but which can be violated by edges or small limbs. Because our dataset consists of masked objects, identifying the occluding contour $C$ is trivial (see Figure~\ref{fig:SFC1}). For each point $i$ on $C$, we estimate $n_{i}$, the local normal to the occluding contour in the image plane. Using those we regularize the surface normals in $Z$ along the boundary by minimizing the following loss:
\begin{eqnarray}
f_{c}(Z) = \sum_{ i \in C } \left( 1 - \left( N_{i}^x(Z) n_{i}^x + N_{i}^y(Z) n_{i}^y \right) \right)^{\gamma_c}
\end{eqnarray}
Where $N(Z)$ is the surface normal of $Z$, as defined in Appendix A. We set $\gamma_c = 0.75$, which fits the training data best, and which performs best in practice. The inner product of $n_i$ and $N_i$ (both of which are unit-norm) is $1$ when both vectors are oriented in the same direction, in which case the loss is 0. If the normals do not agree, then some cost is incurred. This cost corresponds to a heavy-tailed distribution (shown in Figure~\ref{fig:ShapeContour2}) which encourages the surface orientation to match the orientation of the occluding contour at limbs, allows surface normals to violate this assumption at edges.

This occluding-contour prior, when combined with our priors on smooth and isotropic shapes, allows us to easily define an ablation of our entire model that corresponds to a shape-from-contour algorithm: we simply optimize with respect to these shape priors, and ignore our priors on reflectance and illumination, thereby ignoring all but the silhouette of the input image. An example of the output of our shape-from-contour model can be seen in Figure~\ref{fig:SFC2}, and this model is evaluated quantitatively against our complete SIRFS model in Section~\ref{sec:experiments}.

\section{Priors over Illumination}
\label{sec:illumination}

Because illumination is unknown, we must regularize it during inference. Our prior on illumination is extremely simple: we fit a multivariate Gaussian to the spherical-harmonic illuminations in our training set. During inference, the cost we impose is the (non-normalized) negative log-likelihood under that model:
\begin{eqnarray}
h(L) = \lambda_L (L - \boldsymbol\mu_L)^\mathrm{T} \mathrm{\Sigma}_L^{-1}  (L - \boldsymbol\mu_L)
\end{eqnarray}
where $\boldsymbol\mu_L$ and $\mathrm{\Sigma}_L$ are the parameters of the Gaussian we learned, and $\lambda_L$ is the multiplier on this prior (learned on the training set). 

We use a spherical-harmonic (SH) model of illumination, so $L$ is a $9$ (grayscale) or $27$ (color, $9$ dimensions per RGB channel) dimensional vector. In contrast to traditional SH illumination, we parametrize log-shading rather than shading. This choice makes optimization easier as we don't have to deal with ``clamping'' illumination at $0$, and it allows for easier regularization, as the space of log-shading SH illuminations is surprisingly well-modeled by a simple multivariate Gaussian while the space of traditional SH illumination coefficients is not. 

\begin{figure}[t!]
	\centering
	\subfigure[\scriptsize ``Laboratory'' Data/Samples]{
		\resizebox{0.71in}{!}{\includegraphics{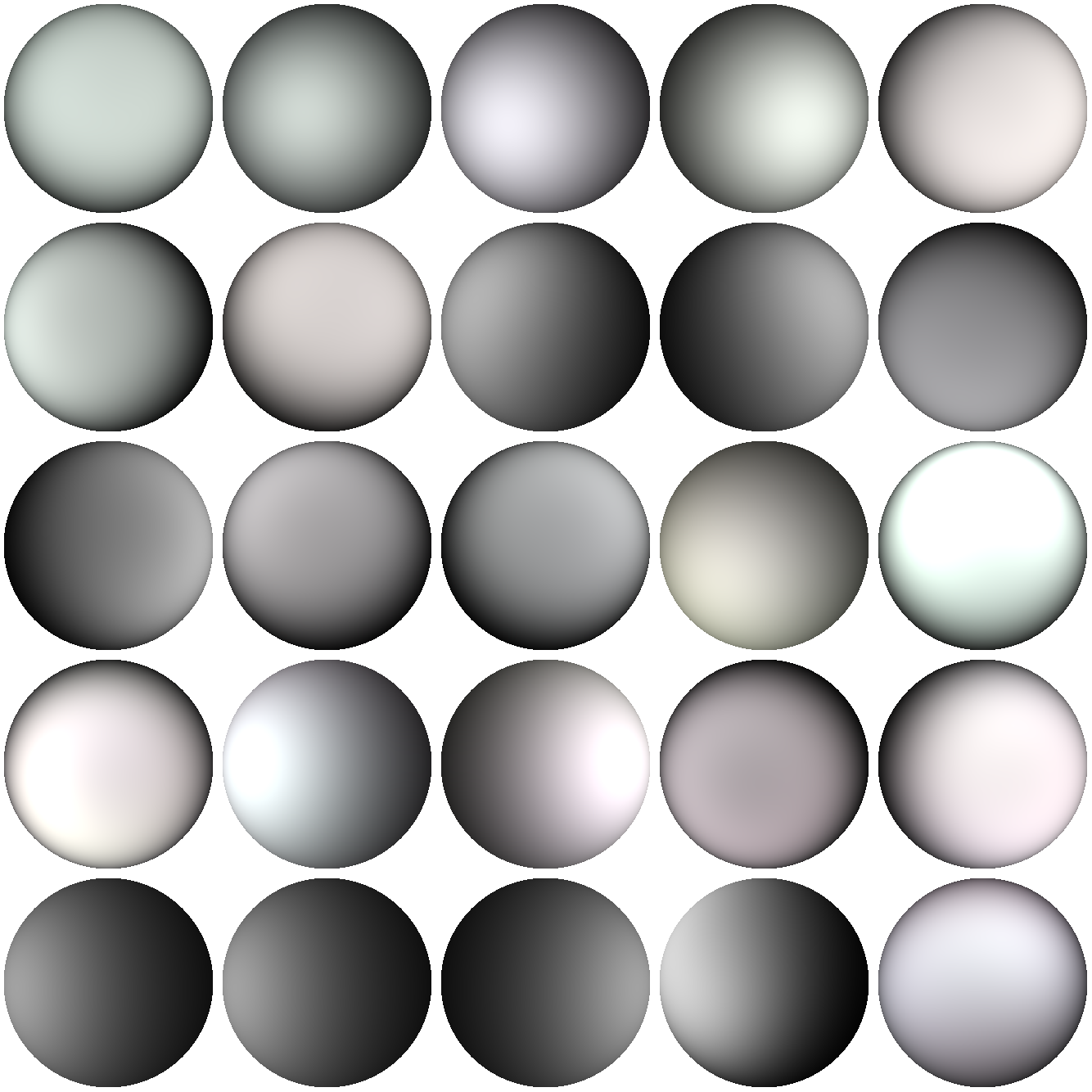}}
		\,
		\resizebox{0.71in}{!}{\includegraphics{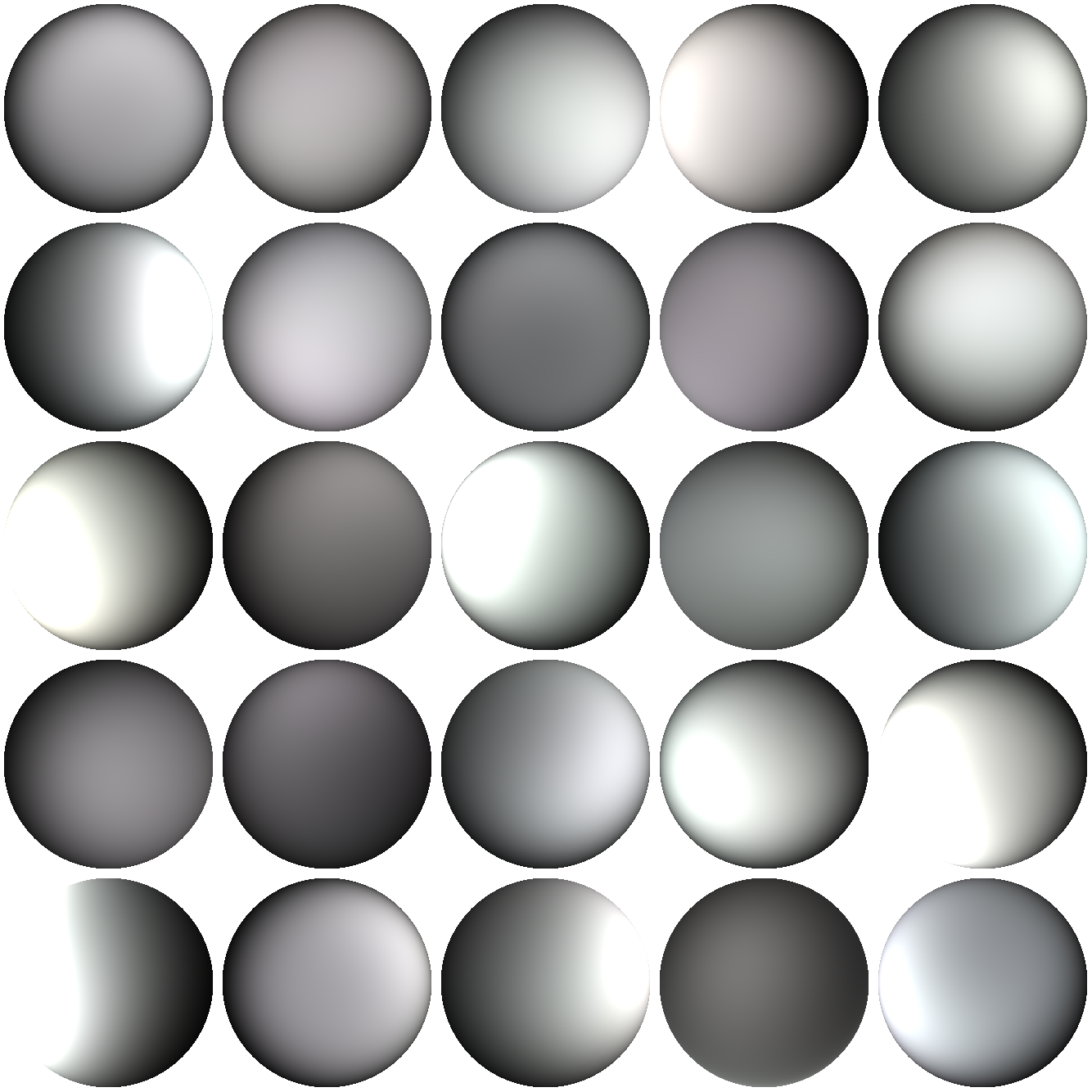}}
		 \label{fig:Lights_Lab}
	}
	\subfigure[\scriptsize ``Natural'' Data/Samples]{
		\resizebox{0.71in}{!}{\includegraphics{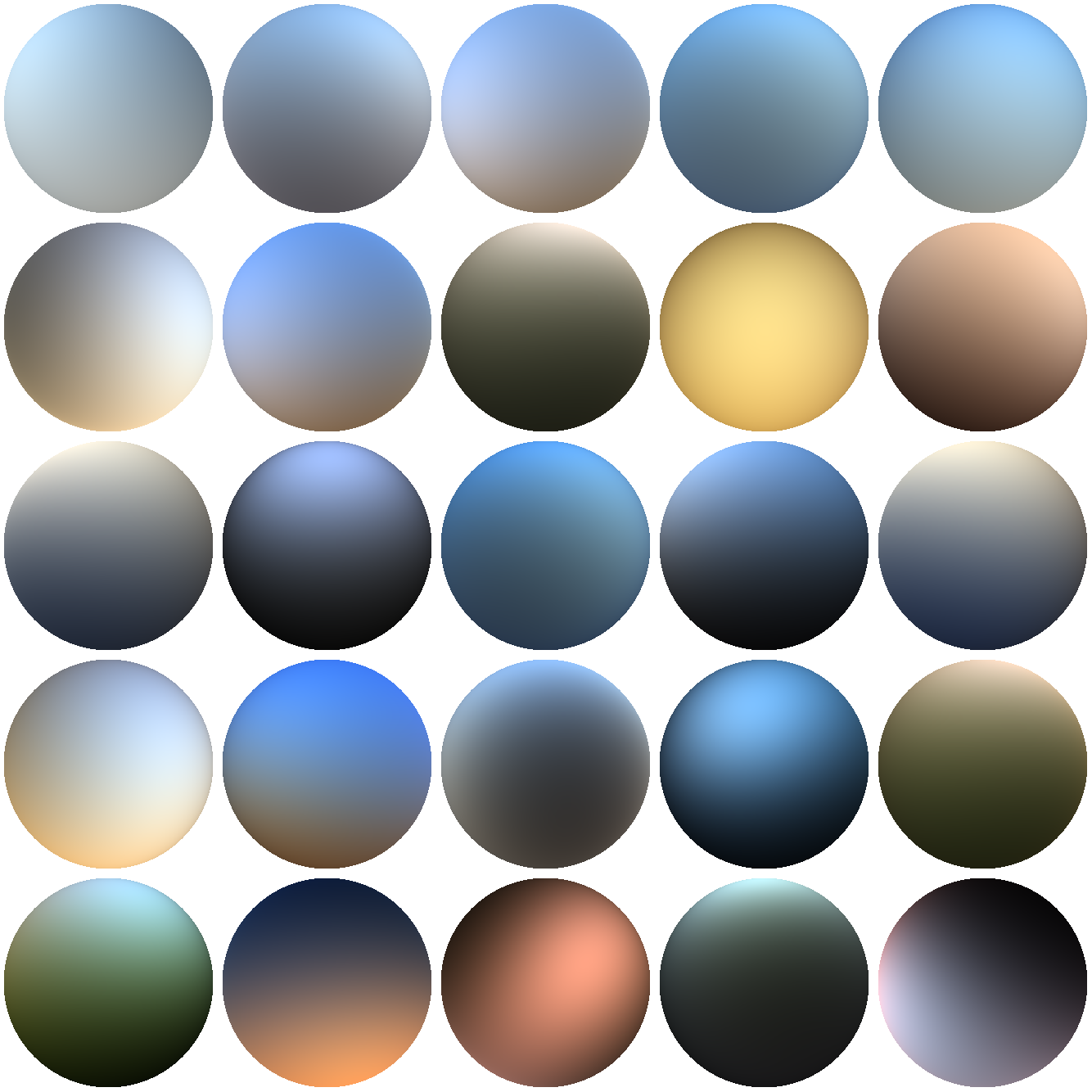}}
		\,
		\resizebox{0.71in}{!}{\includegraphics{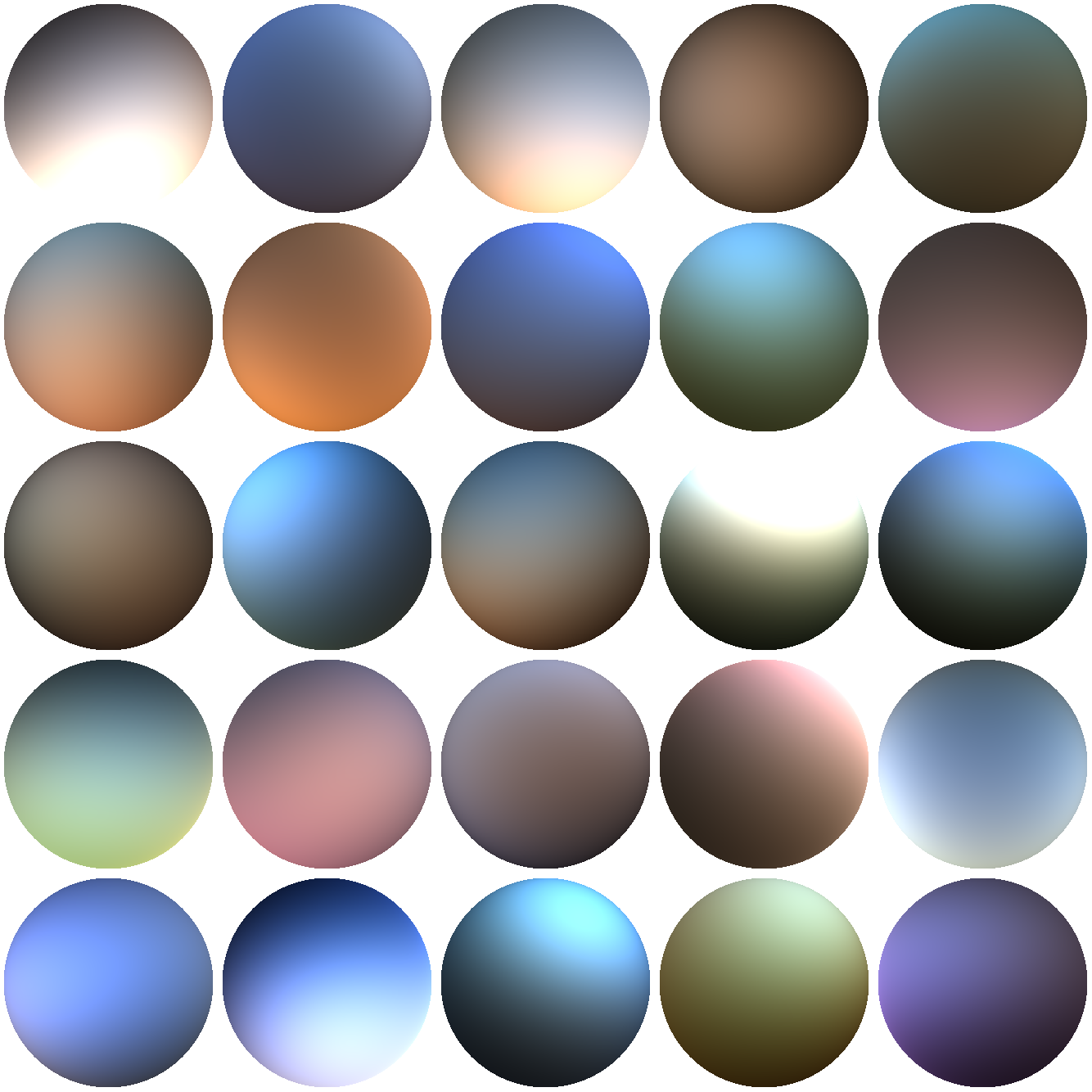}}
		 \label{fig:Lights_Nat}
	}
	\caption{
	We use two datasets:  the ``laboratory''-style illuminations of the MIT intrinsic images dataset~\cite{grosse09intrinsic} which are harsh, mostly-white, and well-approximated by point sources, and a dataset of ``natural'' illuminations, which are softer and much more colorful. Shown here are some illuminations from the training sets of our two datasets, and samples from a multivariate Gaussian fit to each training set (our illumination prior from Section~\ref{sec:illumination}), rendered on Lambertian spheres. In each visualization the illuminations are sorted from least costly (upper left) to most costly (lower right) according to either our ``Laboratory'' or ``Natural'' illumination priors.
	 \label{fig:Lights}}
\end{figure}

See Figure~\ref{fig:Lights} for examples of SH illuminations in our different training sets, as well as samples from our model. The illuminations in Figure~\ref{fig:Lights} come from two different datasets (see Section~\ref{sec:experiments}) for which we build two different priors. We see that our samples look similar to the illuminations in the training set, suggesting that our model fits the data well. The illuminations in these visualizations are sorted by their likelihoods under our priors, which allows us to build an intuition for what these illumination priors encourage. More likely illuminations tend to be lit from the front and are usually less saturated and more ambient, while unlikely illuminations are often lit from unusual angles and tend to exhibit strong shadowing and colors.

\section{Optimization}
\label{sec:optimization}

To estimate shape, illumination, and reflectance, we must solve the optimization problem in Equation~\ref{Eq_SIRFS2}. This is a challenging optimization problem, and naive gradient-based optimization with respect to $Z$ and $L$ fails badly. We therefore present an effective multi-scale optimization technique, which is similar in spirit to multigrid methods \cite{Terzopoulos}, but extremely general and simple to implement.  We will describe our technique in terms of optimizing $a(X)$, where $a(\cdot)$ is some loss function and $X$ is some signal.

Let us define $\Gaussian$, which constructs a Gaussian pyramid from a signal. Because Gaussian pyramid construction is a linear operation, we will treat $\Gaussian$ as a matrix. Instead of minimizing $a(X)$ directly, we minimize $b(Y)$, where $X = \Gaussian^\mathrm{T} Y$:
\begin{eqnarray}
&& \left[ \mathit{\ell}, \nabla_Y \mathit{\ell} \right] = b(Y) :  \\
&& \quad\quad X \gets \Gaussian^\mathrm{T} Y \textrm{\small // reconstruct signal} \nonumber \\
&& \quad\quad \left[ \mathit{\ell}, \nabla_X \mathit{\ell} \right] \gets a(X)  \textrm{ \small // compute loss \& gradient} \nonumber \\ 
&& \quad\quad \nabla_Y \mathit{\ell} \gets \Gaussian \nabla_X \textrm{\small // backpropagate gradient} \nonumber
\end{eqnarray}
We initialize $Y$ to a vector of all $0$'s, and then solve for $\hat{X} = \Gaussian^\mathrm{T} \left( {\operatorname{arg\,min}_{Y} b(Y)} \right)$ using L-BFGS. Any arbitrary gradient-based optimization technique could be used, but L-BFGS worked best in our experience.

The choice of the filter used in constructing our Gaussian pyramid is crucial.  We found that $4$-tap binomial filters work well, and that the choice of the magnitude of the filter dramatically affects multiscale optimization.  If the magnitude is small, then the coefficients of the upper levels of the pyramid are so small that they are effectively ignored, and optimization fails (and in the limit, a filter magnitude of $0$ reduces our model to single-scale optimization). Conversely, if the magnitude is large, then the coarse scales of the pyramid are optimized and the fine scales are ignored. The filter that we found worked best is: $\frac{1}{\sqrt{8}} [1, 3, 3, 1] $, which has twice the magnitude of the filter that would normally be used for Gaussian pyramids. This increased magnitude biases optimization towards adjusting coarse scales before fine scales, without preventing optimization from eventually optimizing fine scales. This filter magnitude does not appear to be universally optimal --- different tasks appear to have different optimal filter magnitudes. Note that this technique is substantially different from standard coarse-to-fine optimization, in that \emph{all} scales are optimized simultaneously. As a result, we find much lower minima than standard coarse-to-fine techniques, which tend to keep coarse scales fixed when optimizing over fine scales. Optimization is also much faster than comparable coarse-to-fine techniques.

To  optimizing Equation~\ref{Eq_SIRFS2} we initialize $Z$ and $L$ to $\vec{0}$ ( $L=\vec{0}$ is equivalent to an entirely ambient, white illumination) and optimize with respect to a vector that is a concatenation of $\Gaussian^\mathrm{T} Z$ and a whitened version of $L$. We optimize in the space of whitened illuminations because the Gaussians we learn for illumination mostly describe a low-rank subspace of SH coefficients, and so optimization in the space of unwhitened illumination is ill-conditioned. We pre-compute a whitening transformation for $\Sigma_L$ and $\mu_L$, and during each evaluation of the loss in gradient descent we unwhiten our whitened illumination, compute the loss and gradient, and backpropagate the gradient onto the whitened illumination. After optimizing Equation~\ref{Eq_SIRFS2} we have a recovered depth map $\hat Z$ and illumination $\hat L$, with which we calculate a reflectance image $\hat R = I - S(\hat Z, \hat L) $. When illumination is known, $L$ is fixed. Optimizing to near-convergence (which usually takes a few hundred iterations) for a $1$-$2$ megapixel grayscale image takes $1$-$5$ minutes on a 2011 Macbook Pro, using a straightforward Matlab/C implementation. Optimization takes roughly twice as long if the image is color. See Appendix E for a description of some methods we use to make the evaluation of our loss function more efficient.

We use this same multiscale optimization scheme with L-BFGS to solve the optimization problems in Equations \ref{eq:GraySpline} and \ref{eq:ColorSpline}, though we use different filter magnitudes for the pyramids. Naive single-scale optimization for these problems works poorly.

\section{Experiments}
\label{sec:experiments}

Quantitatively evaluating the accuracy of our model is challenging, as there are no pre-existing datasets with ground-truth shape, surface normals, shading, reflectance, and illumination. Thankfully, the MIT Intrinsic Images dataset \cite{grosse09intrinsic} provides ground-truth shading and reflectance for 20 objects (one object per image), and includes many additional images of each object under different illumination conditions. Given this, we have created the MIT-Berkeley Intrinsic Images dataset, an augmented version of the MIT Intrinsic Images dataset in which we have used photometric stereo on the additional images of each object to estimate the shape of each object and the spherical harmonic illumination for each image. An example object in our dataset can be seen in Figure~\ref{fig:MITImages}, and the appendix contains additional images and details of our photometric stereo algorithm. In all of our experiments, we use the following test-set: cup2, deer, frog2, paper2, pear, potato, raccoon, sun, teabag1, turtle. The other 10 objects are used for training.

An additional difficulty in evaluation is the choice of error metrics. Constructing error metrics for specific intrinsic scene properties such as a depth map or a reflectance image is challenging, as naive choices such as mean-squared-error often correspond very poorly with the perceptual salience of an error. Additionally, constructing a single error metric that describes all errors in each intrinsic scene property is difficult. We therefore present six different error metrics that have been designed to capture different kinds of important errors for each intrinsic scene property:  $\ZMAE$ is the shift-invariant absolute error between the estimated shape and the ground-truth shape. $\NMAE$ is the mean error between our estimated normal field and ground-truth normal field, in radians. $\SMSE$ and $\AMSE$ are the scale-invariant mean-squared-error of our recovered shading and reflectance, respectively. $\mitMSE$ is the error metric introduced in conjunction with the MIT intrinsic images dataset \cite{grosse09intrinsic}, which measures a locally scale-invariant error for both reflectance and shading\footnote{The authors of \cite{grosse09intrinsic} refer to this error metric to as ``LMSE'', but we will call it $\mitMSE$ to minimize confusion with $\lightMSE$}. $\lightMSE$ is the scale-invariant MSE of a rendering of our recovered illumination on a sphere, relative to a rendering of the ground-truth illumination. To summarize these individual error metrics, we report an ``average'' error metric, which is the geometric mean of the previous six error metrics.  For each error metric and the average metric, we report the geometric mean of error across the test-set images. The use of the geometric mean prevents the average error from being dominated by individual error metrics with large dynamic ranges, or by particularly challenging images. See Appendix F for a thorough explanation of our choice of error metrics.

Though the MIT dataset has a great deal of variety in terms of the kinds of objects used, the illumination in the dataset is very ``laboratory''-like --- lights are white, and are placed at only a few locations relative to the object. See Figure~\ref{fig:Lights_Lab} for examples of these ``laboratory'' illuminations. In contrast, natural illuminations exhibit much more color and variety: the sun is yellow, outdoor shadows are often tinted blue, man-made illuminants have different colors, and indirect illumination from colored objects may cause very colorful illuminations. To acquire some illumination models that are more representative of the variety seen in the natural world, we took all of the environment maps from the sIBL Archive\footnote{\url{http://www.hdrlabs.com/sibl/archive.html}}, expanded that set of environment maps by shifting and mirroring them and varying their contrast and saturation (saturation is only ever decreased, never increased) and produced spherical harmonic illuminations from the resulting environment maps. After removing similar illuminations, the illuminations were split into training and test sets. See Figure~\ref{fig:Lights_Nat} for examples of these ``natural'' illuminations. Each object in the MIT dataset was randomly assigned an illumination (such that training illuminations were assigned to training objects, etc), and each object was re-rendered under its new illumination, using that object's ground-truth shape and reflectance. We will refer to this new pseudo-synthetic dataset of naturally illuminated objects as our ``natural'' illumination dataset, and we will refer to the original MIT images as the ``laboratory'' illumination dataset. From our experience applying our model to real-world images, these ``natural'' illuminations appear to be much more representative of the sort of illumination we see in uncontrolled environments, though the dataset is heavily biased towards more colorful illuminations. We attribute this to a photographer's bias towards ``interesting'' environment maps in the sIBL Archive.

\begin{table}[b!]
\begin{center}
\resizebox{3.3in}{!}{\Large
\begin{tabular}{ c@{\,}l  |  c   c     c    c    c    c  |  c  } 
\multicolumn {9}{ c }{\Huge I. Grayscale Images, Laboratory Illumination}\\
\multicolumn {2}{ c }{ Algorithm } & $\ZMAE$ & $\NMAE$ & $\SMSE$ & $\AMSE$ & $\mitMSE$ & \multicolumn {1}{ c }{$\lightMSE$} & \multicolumn {1}{ c }{Avg.} \\ 
\hline
(1) &  Naive Baseline & $25.56$ & $0.7223$ & $0.0571$ & $0.0426$ & $0.0353$ & $0.0484$ & $0.2061$\\ 
(2) &  Retinex \cite{grosse09intrinsic,HornLightness} + SFS & $67.15$ & $0.8342$ & $0.0311$ & $0.0265$ & $0.0289$ & $0.0484$ & $0.2002$\\ 
(3) &  Tappen \etal \cite{Tappen} + SFS & $41.96$ & $0.7413$ & $0.0354$ & $0.0252$ & $0.0285$ & $0.0484$ & $0.1835$\\ 
(4) &  Shen \etal \cite{ShenYJL11} + SFS & $45.57$ & $0.8293$ & $0.0493$ & $0.0427$ & $0.0436$ & $0.0484$ & $0.2348$\\ 
\hline
(A) &  SIRFS  & $31.00$ & $0.5343$ &  \cellcolor{Yellow} $\bf{ 0.0156 } $ &  \cellcolor{Yellow} $\bf{ 0.0177 } $ &  \cellcolor{Yellow} $\bf{ 0.0209 } $ &  \cellcolor{Yellow} $\bf{ 0.0103 } $ &  \cellcolor{Yellow} $\bf{ 0.0998 } $\\ 
(B) &  SIRFS, no R-smoothness & $27.25$ & $0.5361$ & $0.0267$ & $0.0255$ & $0.0290$ & $0.0152$ & $0.1279$\\ 
(C) &  SIRFS, no R-parsimony & $23.53$ & $0.4862$ & $0.0224$ & $0.0261$ & $0.0228$ & $0.0167$ & $0.1170$\\ 
(D) &  SIRFS, no R-absolute & $24.02$ & $0.5023$ & $0.0190$ & $0.0201$ & $0.0222$ & $0.0122$ & $0.1037$\\ 
(E) &  SIRFS, no Z-smoothness & $29.05$ & $0.5783$ & $0.0241$ & $0.0227$ & $0.0337$ & $0.0125$ & $0.1254$\\ 
(F) &  SIRFS, no Z-isotropy & $98.07$ & $0.7560$ & $0.0200$ & $0.0198$ & $0.0268$ & $0.0104$ & $0.1419$\\ 
(G) &  SIRFS, no Z-contour & $34.29$ & $0.7676$ & $0.0208$ & $0.0207$ & $0.0232$ & $0.0231$ & $0.1351$\\ 
(H) &  SIRFS, no L-gaussian & $26.75$ & $0.5929$ & $0.0270$ & $0.0212$ & $0.0327$ & $0.1940$ & $0.1964$\\ 
(I) &  SIRFS, no Z-multiscale & $25.58$ & $0.7233$ & $0.0571$ & $0.0426$ & $0.0353$ & $0.0414$ & $0.2009$\\ 
(J) &  SIRFS, no L-whitening & $33.93$ & $0.5837$ & $0.0207$ & $0.0208$ & $0.0256$ & $0.0119$ & $0.1171$\\ 
(K) &  Shape-from-Contour &  \cellcolor{Yellow} $\bf{ 18.96 }$ &  \cellcolor{Yellow} $\bf{ 0.4192 } $ & $0.0571$ & $0.0426$ & $0.0353$ & $0.0484$ & $0.1791$\\ 
&&&&&&& \\ 
(S) &  shape observation & $4.83$ & $0.1952$ &  -  &  -  &  -  &  -  &  - \\ 
(A$+$S) &  SIRFS + shape observation & $3.72$ & $0.2414$ & $0.0128$ & $0.0176$ & $0.0210$ & $0.0096$ & $0.0586$\\ 
(A$+$L) &  SIRFS + known illumination & $27.32$ & $0.4944$ & $0.0175$ & $0.0179$ & $0.0225$ &  -  &  - \\ 
\multicolumn {9}{ c }{}\\
\multicolumn {9}{ c }{}\\
\multicolumn {9}{ c }{\Huge II. Color Images, Laboratory Illumination}\\
\multicolumn {2}{ c }{ Algorithm } & $\ZMAE$ & $\NMAE$ & $\SMSE$ & $\AMSE$ & $\mitMSE$ & \multicolumn {1}{ c }{$\lightMSE$} & \multicolumn {1}{ c }{Avg.} \\ 
\hline
(1) &  Naive Baseline & $25.56$ & $0.7223$ & $0.0577$ & $0.0455$ & $0.0354$ & $0.0489$ & $0.2092$\\ 
(2) &  Retinex \cite{grosse09intrinsic,HornLightness} + SFS & $85.34$ & $0.8056$ & $0.0204$ & $0.0186$ & $0.0163$ & $0.0489$ & $0.1658$\\ 
(3) &  Tappen \etal \cite{Tappen} + SFS & $41.96$ & $0.7413$ & $0.0361$ & $0.0379$ & $0.0347$ & $0.0489$ & $0.2040$\\ 
(4) &  Shen \etal \cite{ShenYJL11} + SFS & $55.95$ & $0.8529$ & $0.0528$ & $0.0458$ & $0.0398$ & $0.0489$ & $0.2466$\\ 
(5) &  Gehler \etal \cite{Gehler2011} + SFS & $53.36$ & $0.6844$ & $0.0106$ & $0.0101$ & $0.0131$ & $0.0489$ & $0.1166$\\ 
\hline
(A) &  SIRFS  & $19.24$ &  \cellcolor{Yellow} $\bf{ 0.3914 } $ &  \cellcolor{Yellow} $\bf{ 0.0064 } $ &  \cellcolor{Yellow} $\bf{ 0.0098 } $ &  \cellcolor{Yellow} $\bf{ 0.0125 } $ & $0.0096$ &  \cellcolor{Yellow} $\bf{ 0.0620 } $\\ 
(B) &  SIRFS, no R-smoothness & $19.23$ & $0.4046$ & $0.0125$ & $0.0163$ & $0.0214$ & $0.0092$ & $0.0824$\\ 
(C) &  SIRFS, no R-parsimony & $19.45$ & $0.4312$ & $0.0096$ & $0.0149$ & $0.0140$ & $0.0091$ & $0.0731$\\ 
(D) &  SIRFS, no R-absolute & $22.98$ & $0.4288$ & $0.0085$ & $0.0113$ & $0.0135$ & $0.0095$ & $0.0704$\\ 
(E) &  SIRFS, no Z-smoothness & $19.28$ & $0.4367$ & $0.0114$ & $0.0116$ & $0.0219$ &  \cellcolor{Yellow} $\bf{ 0.0088 } $ & $0.0773$\\ 
(F) &  SIRFS, no Z-isotropy & $84.08$ & $0.7013$ & $0.0117$ & $0.0128$ & $0.0160$ & $0.0103$ & $0.1063$\\ 
(G) &  SIRFS, no Z-contour & $32.59$ & $0.7351$ & $0.0103$ & $0.0146$ & $0.0173$ & $0.0444$ & $0.1186$\\ 
(H) &  SIRFS, no L-gaussian & $20.81$ & $0.4631$ & $0.0199$ & $0.0140$ & $0.0183$ & $0.1272$ & $0.1358$\\ 
(I) &  SIRFS, no Z-multiscale & $25.62$ & $0.7237$ & $0.0574$ & $0.0453$ & $0.0353$ & $0.0401$ & $0.2022$\\ 
(J) &  SIRFS, no L-whitening & $24.96$ & $0.4766$ & $0.0106$ & $0.0156$ & $0.0188$ & $0.0138$ & $0.0894$\\ 
(K) &  Shape-from-Contour &  \cellcolor{Yellow} $\bf{ 18.96 }$ & $0.4192$ & $0.0577$ & $0.0455$ & $0.0354$ & $0.0489$ & $0.1818$\\ 
&&&&&&& \\ 
(S) &  shape observation & $4.83$ & $0.1952$ &  -  &  -  &  -  &  -  &  - \\ 
(A$+$S) &  SIRFS + shape observation & $3.40$ & $0.2126$ & $0.0070$ & $0.0111$ & $0.0153$ & $0.0063$ & $0.0420$\\ 
(A$+$L) &  SIRFS + known illumination & $18.58$ & $0.3761$ & $0.0076$ & $0.0120$ & $0.0146$ &  -  &  - \\ 
\multicolumn {9}{ c }{}\\
\multicolumn {9}{ c }{}\\
\multicolumn {9}{ c }{\Huge III. Color Images, Natural Illumination}\\
\multicolumn {2}{ c }{ Algorithm } & $\ZMAE$ & $\NMAE$ & $\SMSE$ & $\AMSE$ & $\mitMSE$ & \multicolumn {1}{ c }{$\lightMSE$} & \multicolumn {1}{ c }{Avg.} \\ 
\hline
(1) &  Naive Baseline & $25.56$ & $0.7223$ & $0.0283$ & $0.0266$ & $0.0125$ & $0.0371$ & $0.1364$\\ 
(2) &  Retinex \cite{grosse09intrinsic,HornLightness} + SFS & $26.76$ & $0.5851$ & $0.0174$ & $0.0174$ & $0.0083$ & $0.0371$ & $0.1066$\\ 
(3) &  Tappen \etal \cite{Tappen} + SFS & $53.87$ & $0.7255$ & $0.0255$ & $0.0280$ & $0.0268$ & $0.0371$ & $0.1740$\\ 
(4) &  Gehler \etal \cite{Gehler2011} + SFS & $37.66$ & $0.6398$ & $0.0162$ & $0.0150$ & $0.0105$ & $0.0371$ & $0.1149$\\ 
\hline
(A) &  SIRFS  & $28.21$ & $0.4057$ & $0.0055$ & $0.0046$ &  \cellcolor{Yellow} $\bf{ 0.0036 } $ & $0.0103$ & $0.0469$\\ 
(B) &  SIRFS, no R-smoothness & $28.41$ & $0.4192$ & $0.0061$ & $0.0057$ & $0.0062$ & $0.0104$ & $0.0546$\\ 
(C) &  SIRFS, no R-parsimony & $28.90$ & $0.4184$ & $0.0073$ & $0.0064$ & $0.0041$ & $0.0107$ & $0.0540$\\ 
(D) &  SIRFS, no R-absolute & $20.63$ &  \cellcolor{Yellow} $\bf{ 0.3538 } $ & $0.0068$ & $0.0058$ & $0.0039$ & $0.0091$ &  \cellcolor{Yellow} $\bf{ 0.0466 } $\\ 
(E) &  SIRFS, no Z-smoothness & $24.68$ & $0.4441$ & $0.0087$ & $0.0062$ & $0.0095$ & $0.0099$ & $0.0618$\\ 
(F) &  SIRFS, no Z-isotropy & $50.49$ & $0.4015$ &  \cellcolor{Yellow} $\bf{ 0.0046 } $ &  \cellcolor{Yellow} $\bf{ 0.0039 } $ & $0.0037$ &  \cellcolor{Yellow} $\bf{ 0.0086 } $ & $0.0475$\\ 
(G) &  SIRFS, no Z-contour & $41.27$ & $0.7036$ & $0.0094$ & $0.0083$ & $0.0062$ & $0.0256$ & $0.0843$\\ 
(H) &  SIRFS, no L-gaussian & $20.22$ & $0.3937$ & $0.0100$ & $0.0088$ & $0.0075$ & $0.0483$ & $0.0796$\\ 
(I) &  SIRFS, no Z-multiscale & $25.64$ & $0.7205$ & $0.0279$ & $0.0279$ & $0.0124$ & $0.0291$ & $0.1316$\\ 
(J) &  SIRFS, no L-whitening & $51.74$ & $0.9430$ & $0.0140$ & $0.0106$ & $0.0066$ & $0.0777$ & $0.1246$\\ 
(K) &  Shape-from-Contour &  \cellcolor{Yellow} $\bf{ 19.55 }$ & $0.4253$ & $0.0283$ & $0.0266$ & $0.0125$ & $0.0371$ & $0.1194$\\ 
&&&&&&& \\ 
(S) &  shape observation & $4.83$ & $0.1952$ &  -  &  -  &  -  &  -  &  - \\ 
(A$+$S) &  SIRFS + shape observation & $3.17$ & $0.1471$ & $0.0034$ & $0.0032$ & $0.0030$ & $0.0049$ & $0.0206$\\ 
(A$+$L) &  SIRFS + known illumination & $10.28$ & $0.1957$ & $0.0018$ & $0.0014$ & $0.0022$ &  -  &  - \\ 
\end{tabular}
}
\vspace{1mm}
\caption{We evaluate SIRFS on three different variants of our dataset, and we compare SIRFS to several baseline techniques, several ablations, and two extensions in which additional information is provided.
\label{table:results}}
\end{center} 
\end{table}

Given our dataset, we will evaluate our model on the task of recovering all intrinsic scene properties from a single image of a masked object, under three different conditions: I: the input is a grayscale image and the illumination is ``laboratory''-like, II: the input is a color image and the illumination is ``laboratory''-like, and III: the input is a color image and the illumination is ``natural''. For all tasks, we use the same training/test split, and for each task we tune a different set of hyperparameters on the training set ($\lambda_s, \lambda_e, \lambda_a, \sigma_R, \lambda_k, \lambda_i, \lambda_c,$ and $\lambda_L$), and fit a different prior on illumination (as in Section~\ref{sec:illumination}). Hyperparameters are tuned using coordinate descent to minimize our ``average'' error metric for the training set.  For each task, we compare SIRFS against several intrinsic images algorithms (meant to decompose an image into shading and reflectance components), upon which we've run a shape-from-shading algorithm on the shading image. For the sake of a generous comparison, the SFS algorithm uses our shape priors, which boosts each baseline's performance (detailed in Appendix H).  We also compare against a ``naive'' algorithm, which is a baseline in which $Z=\vec{0}$ and $L=\vec{0}$. Because the intrinsic image baselines do not estimate illumination, we use $L=\vec{0}$ as their prediction. We were forced to use different baseline techniques for different tasks, as some baselines do not have code available for running on new imagery, and some code that was designed for color images crashes when run on grayscale images. 

We also compare against several ablations of our model in which components have been removed: models B-H omit priors by simply setting their $\lambda$ hyperparameters to $0$, and models I and J omit our multiscale optimization over $Z$ and our whitened optimization over $L$ respectively. Model K is a shape-from-contour technique, in which only our shape-priors are non-zero and $L=\vec{0}$, so the only effective input to the model is the silhouette of the object (for this baseline, the hyperparameters have been completely re-tuned on the training set). We also compare against two extensions of SIRFS: model A$+$L, in which the ground-truth illumination is known (and fixed during optimization), and model A$+$S, in which we provide a blurry version of the ground-truth shape (convolved with a Gaussian kernel with $\sigma=30$)  as input. See Appendix D for a description of an additional shape ``prior'' we use to incorporate the external shape observation for this one variant of our model. Model S shows the performance of just the blurry ground-truth shape provided as input to model A$+$S, for reference.  The performance of SIRFS relative to some of these baselines and extensions can be see in Table~\ref{table:results}, in Figure~\ref{fig:MITImages} and in the appendices. 

\begin{figure}[!]
	\centering
	\subfigure[\scriptsize Achromatic illumination]{
		\resizebox{\eightwidth}{!}{\includegraphics{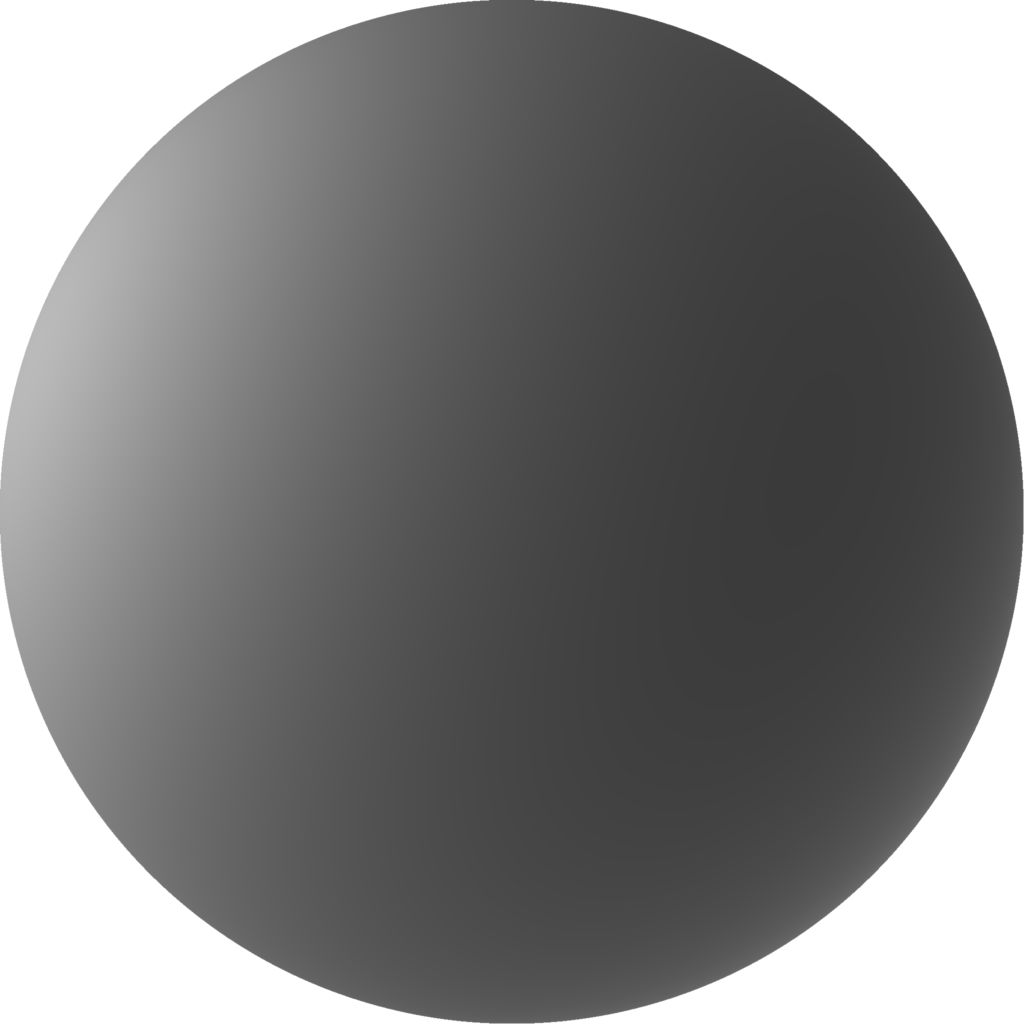}}
		\resizebox{\eightwidth}{!}{\includegraphics{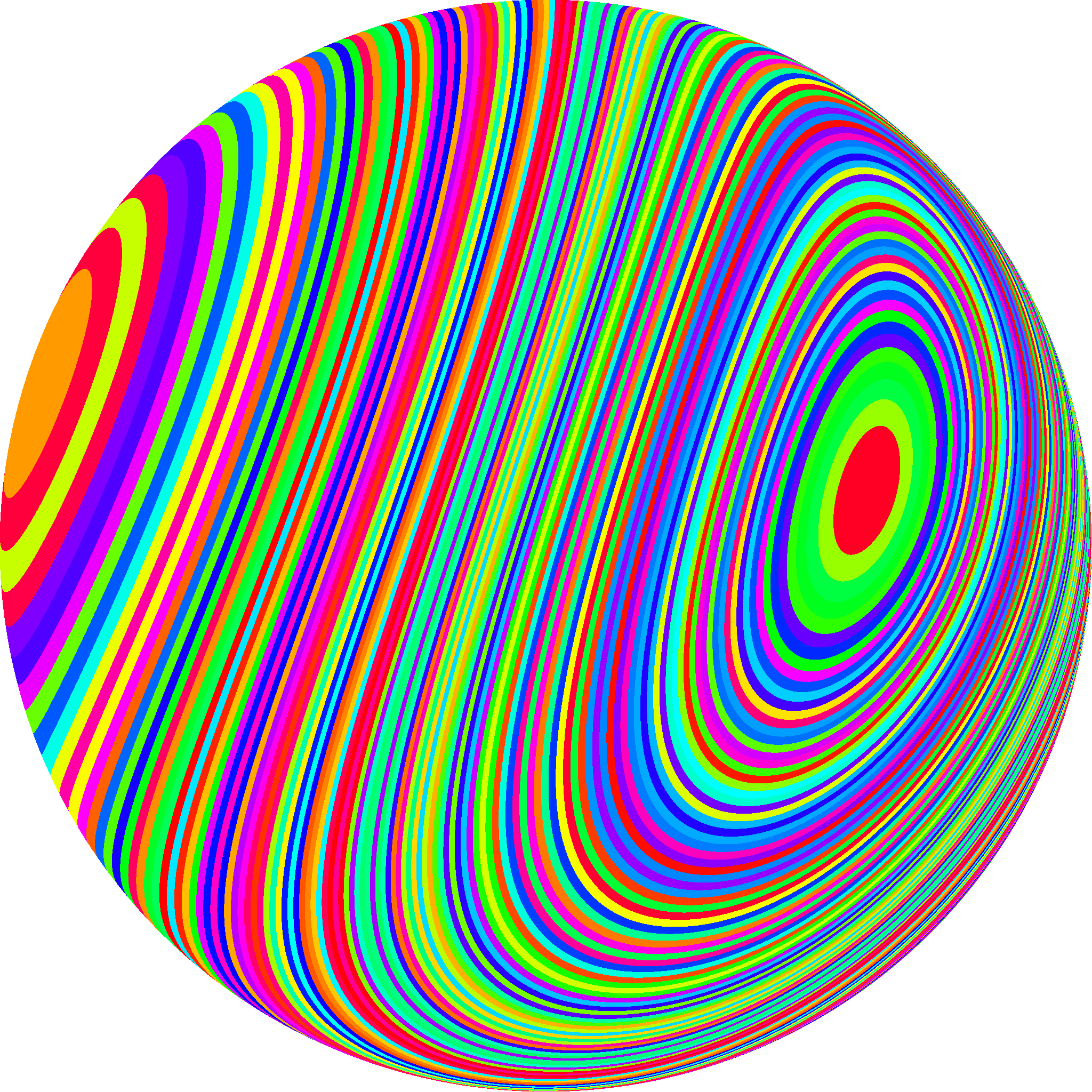}}
		}
	\subfigure[\scriptsize Chromatic illumination]{
		\resizebox{\eightwidth}{!}{\includegraphics{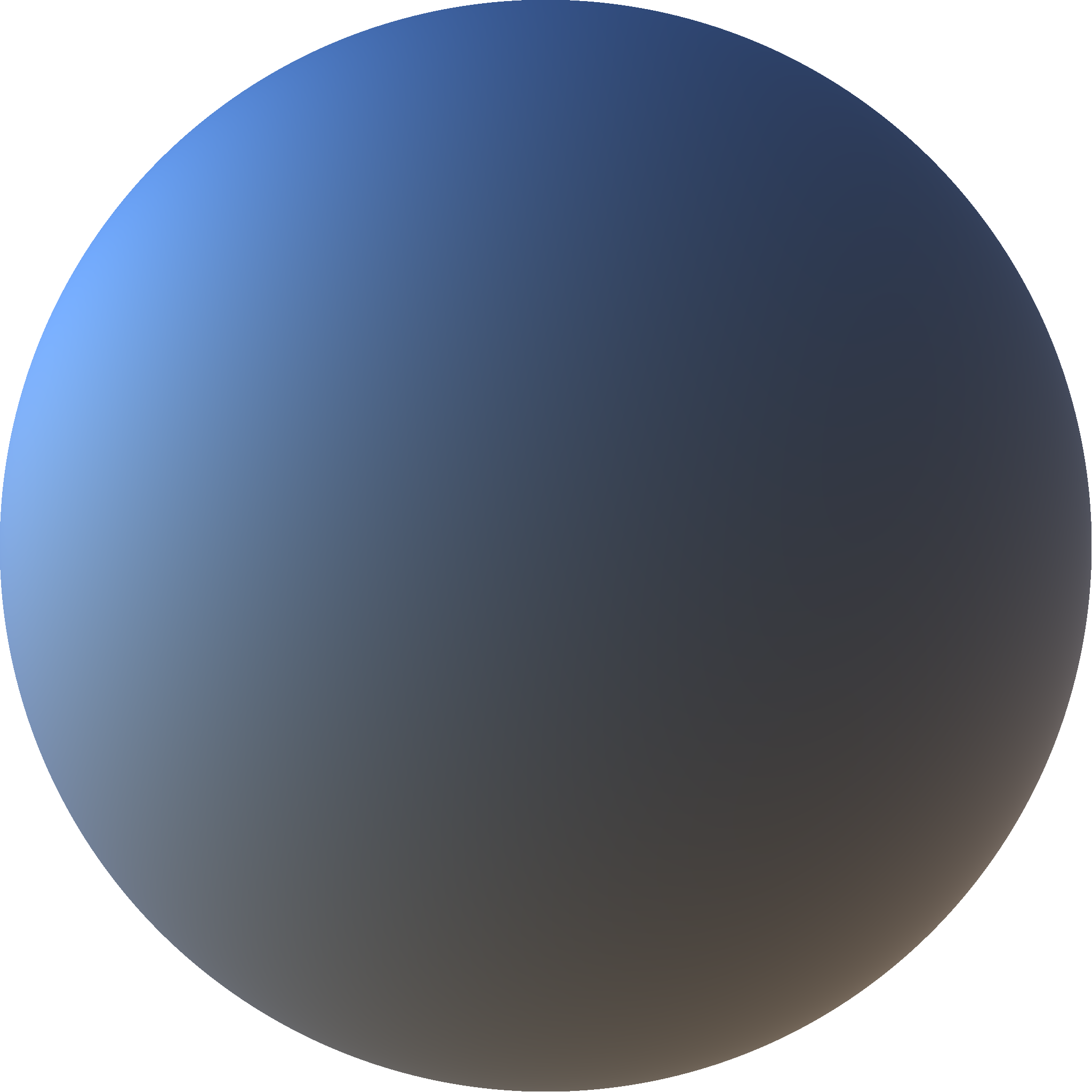}}
		\resizebox{\eightwidth}{!}{\includegraphics{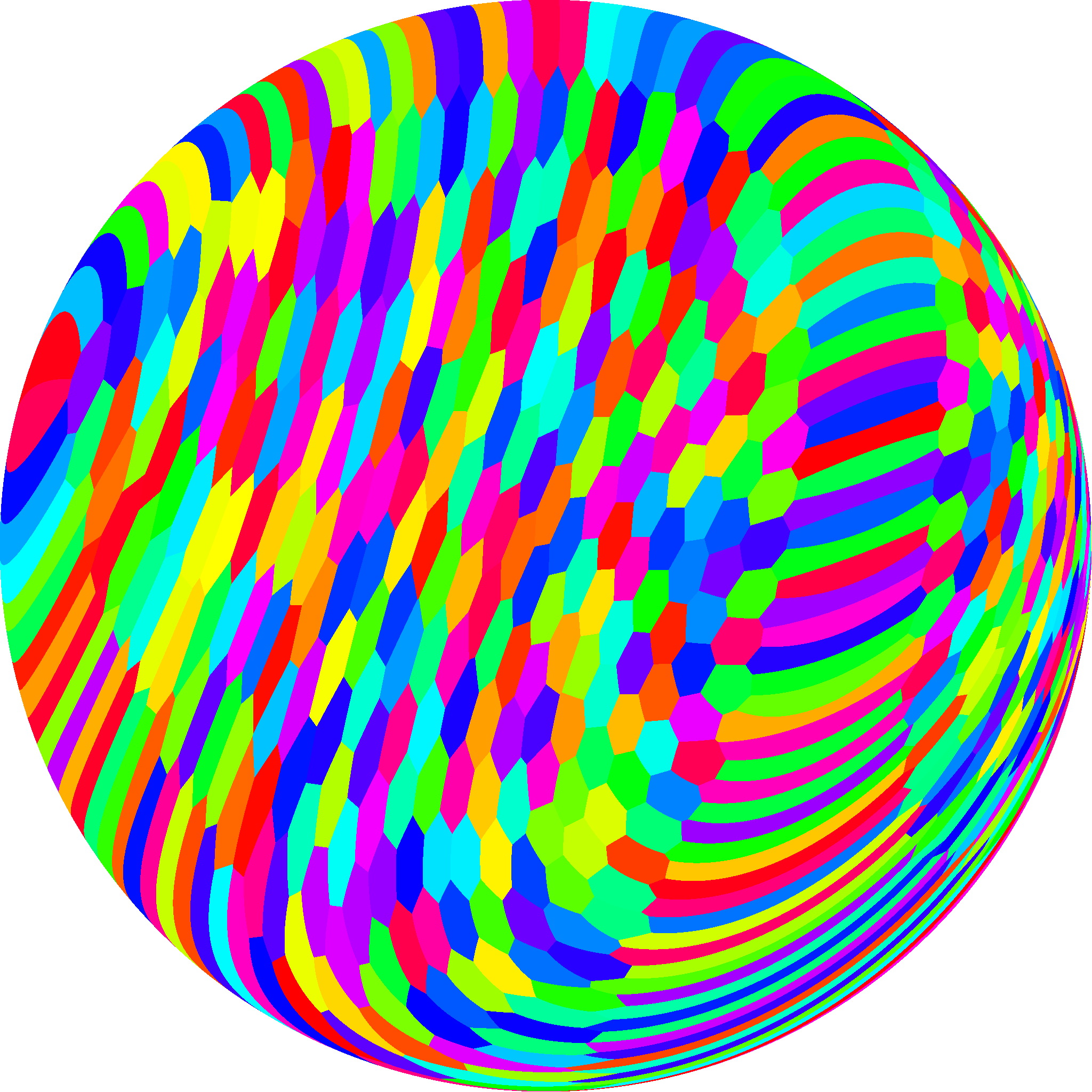}} 
		}
	\caption{ Chromatic illumination dramatically helps shape estimation. Achromatic isophotes (K-means clusters of log-RGB values) are very elongated, while chromatic isophotes are usually more tightly localized.  Therefore, under achromatic lighting a very wide range of surface orientations appear similar, but under chromatic lighting only similar orientations appear similar.
 \label{fig:Lkmeans}}
\end{figure}

\begin{figure*}[!]
	\centering
	\begin{tabular}{ c@{\,}|@{\,}c@{\,}c@{\,}c@{\,}c }
	\subfigure[\small Input Image]{
		\resizebox{\fivewidth}{!}{\includegraphics{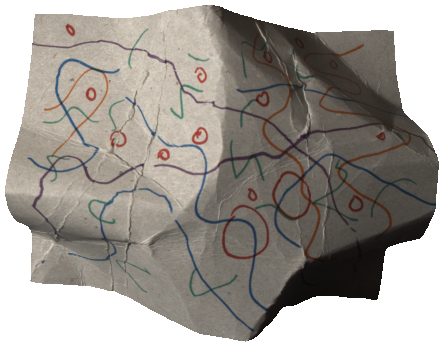}}
		} &
	\subfigure[\small Modified shape]{
		\resizebox{\fivewidth}{!}{\includegraphics{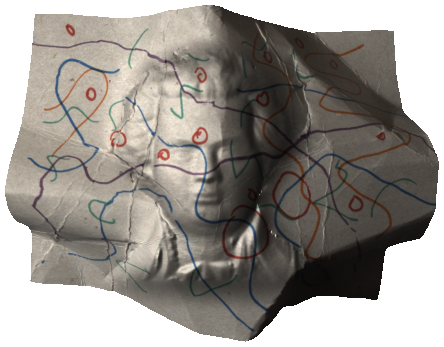}}
		} &
	\subfigure[\small Modified reflectance]{
		\resizebox{\fivewidth}{!}{\includegraphics{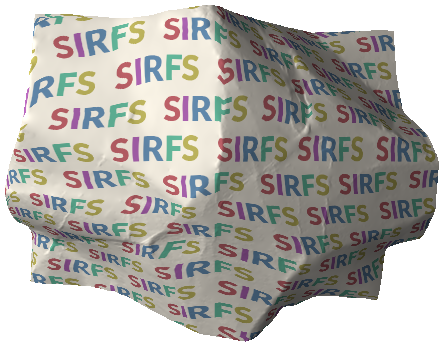}}
		} &
	\subfigure[\small Modified light]{
		\resizebox{\fivewidth}{!}{\includegraphics{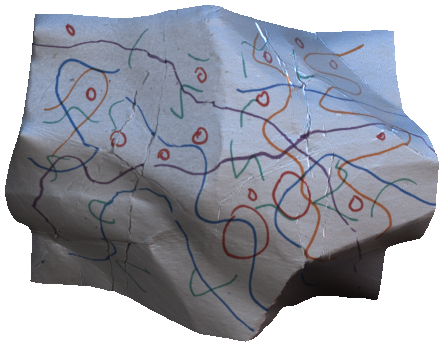}}
		} &
	\subfigure[\small Modified orientation]{
		\resizebox{\fivewidth}{!}{\includegraphics{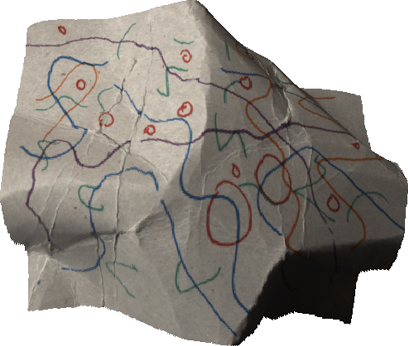}}
		}
	\end{tabular}
	\caption{Our system has obvious graphics applications. Given only a single image, we can estimate an object's shape, reflectance, or illumination, modify any of those three scene properties (or simply rotate the object), and then re-render the object.
 \label{fig:graphics}}
\end{figure*}

From Table~\ref{table:results}, we see that SIRFS outperforms all baseline techniques. For grayscale images, the improvement is substantial: our error is roughly half that of the best technique. For color images under ``laboratory'' illumination, our recovered shading and reflectance images are only slightly better than those of the best-performing intrinsic image technique \cite{Gehler2011}, but our recovered shape and surface normals are significantly better, demonstrating the value of a unified technique over a piecewise system that first does intrinsic images, and then does shape from shading. For color images under ``natural'' illumination, SIRFS outperforms all baseline models by a very large margin, it is the only model that can reason well about color illumination and (implicitly) color shading. From our ablation study, we see that each prior contributes positively to performance, though the improvement we get from each prior is greater in the grayscale case than in the color/natural case. This makes sense, as color images under natural illumination contain much more information than in grayscale images, and so the ``likelihood'' dominates our priors during inference. Our ablation study also shows that our multiscale optimization is absolutely critical to performance. Surprisingly, our shape-from-contour baseline performs very well in terms of our shape/normal error metrics. This is probably just a reflection of the fact that all models are bad at absolute shape reconstruction, due to the inherent ambiguity in shape-from-shading, and so the overly-smooth shape predicted by the shape-from-contour model, by virtue of being smooth and featureless, has a low error relative to the more elaborate depth maps produced by other models. Of course, the shape-from-contour model performs poorly on all other error metrics, as we would expect. This analysis of the inherent difficulty of shape estimation is further demonstrated by model A$+$S, which includes external shape information, and which therefore performs much better in terms of our shape/normal error metrics, but surprisingly performs similarly to model A (basic SIRFS) in terms of all other error metrics. From the performance of model A$+$L we see that knowing the illumination of the scene a-priori does not help much when the illumination is laboratory-like, but helps a great deal when the illumination is ``natural'' --- which makes sense, as more-varied illumination simply makes the reconstruction task more difficult. One surprising conclusion we can draw is that, though the intrinsic image baselines perform worse in the presence of ``natural'' illumination, SIRFS actually performs \emph{better} in natural illumination, as it can exploit color illumination to better disambiguate between shading and reflectance (Figure~\ref{fig:ColorEdge}), and produce higher-quality shape reconstructions (Figure~\ref{fig:Lkmeans}). This finding is consistent with recent work regarding shape-from-shading under natural illumination \cite{johnson11sfs}. However, we should mention that some of the improved performance in the natural illumination task may be due to the fact that the images are pseudo-synthethic (their shading images were produced using our spherical-harmonic rendering) and so they are Lambertian and contain no cast shadows.

In Figure~\ref{fig:graphics}, we demonstrate a simple graphics application using the output of our model, for a color image under laboratory illumination. Given just the output of our model from a single image, we can synthesize novel images in which the shape, reflectance, illumination, or orientation of the object has been changed. The output is not perfect --- the absolute shape is often very incorrect, as we saw in Table~\ref{table:results}, which is due to the inherent ambiguity and difficulty in estimating shape from shading. But such shape errors are usually only visible when rotating the object, and this inherent ambiguity in shape perception often works in our favor when only manipulating reflectance, illumination, or fine-scale shape --- low-frequency errors in shape-estimation made by our model are often not noticed by human observers, because both the model and the human are bad at using shading to estimate coarse shape.

\subsection{Real-World Images}

\begin{figure*}[t!]
	\centering
    \resizebox{6.5in}{!}{\includegraphics{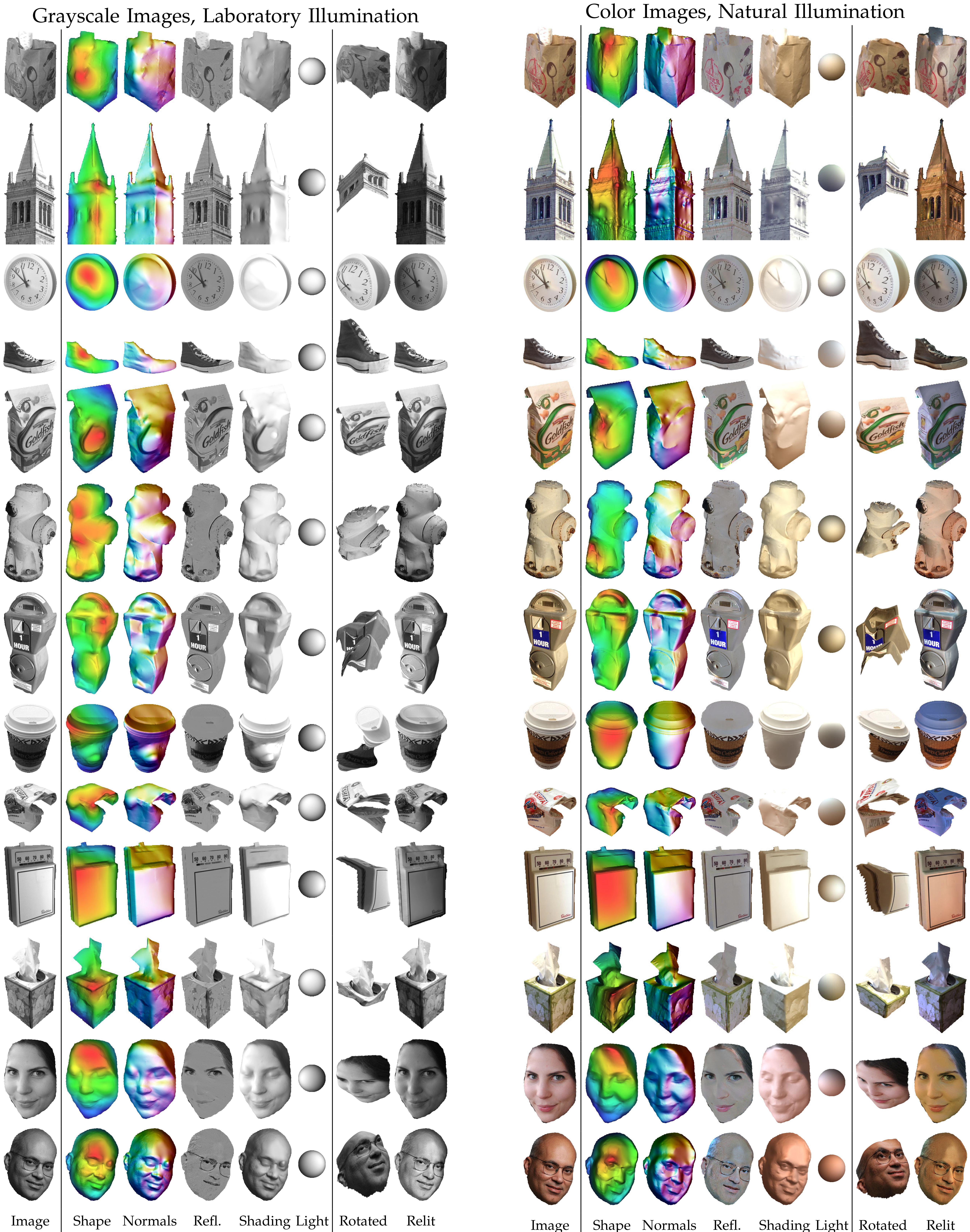}}
	\caption{Our model produces reasonable results on real, manually cropped images of objects. Here are images of arbitrary objects in uncontrolled illumination environments which were downloaded or taken on consumer cameras. For each image, we have the output of our model, and two renderings of our recovered model: one in which we rotate the object, and one in which we relight the object. We run our algorithm on a grayscale version of the image (left), and on the original color image (right). For the color images, we use our ``natural'' illumination model. The code and parameters used for these images are the same as in all other experiments.
	 \label{fig:RealImages}}
\end{figure*}

Though our model quantitatively performs very well on the MIT-Berkeley Intrinsic Images dataset, this dataset is not very representative of the variety of natural objects in the world --- materials are very Lambertian, many reflectances are very synthetic-looking, and illumination is not very varied. We therefore present an additional experiment in which we ran our model on arbitrary masked images of natural objects. We acquired many images (some with an iPhone camera, some with a DSLR, some downloaded from the internet), manually cropped the object in the photo, and used them as input to our model. In Figure~\ref{fig:RealImages} we visualize the output of our model: the recovered shape, normals, reflectance, shading, and illumination, a synthesized view of the object from a different angle, and a synthesized rendering of the object using a different (randomly generated) illumination. We did two experiments: one in which we used a grayscale version of the input image and our laboratory illumination model, and one in which we used the color input image and our natural illumination model. We use the same code and hyperparameters for all images in the two constituent tasks, where our hyperparameters are identical to those used in the previous experiments with the MIT-Berkeley Intrinsic Images dataset.

We see that our model is often able to produce extremely compelling shading and reflectance images, and qualitatively correct illumination. Our recovered shape and surface normals are often somewhat wrong, as evidenced by the new synthesized views of each object, but our ``relit'' objects are often very compelling. The most common mistakes made in shading/reflectance estimation are usually due to our model assuming that the dominant color of the object is due to illumination, not reflectance (such as in the two pictures of faces) which we believe is due to biases in our training data towards white reflectances and colorful illumination. 

\section{Conclusion}
\label{sec:conclusion}

We have presented SIRFS, a model which takes as input a single (masked) image of an object, and produces as output a reasonable estimate of the shape, surface normals, reflectance, shading, and illumination which produced that image. At the core of SIRFS is a series of priors on shape, reflectance, and illumination: surfaces tend to be isotropic and bend infrequency, reflectance images tend to be piecewise smooth and low-entropy, and illumination tends to be natural. Given these priors and our multiscale optimization technique, we can infer the most-likely explanation of a single image subject to our priors and the constraint that the image be explained.  Our unified approach to this problem outperforms all previous solutions to its constituent problems of shape-from-shading and intrinsic image recovery on our challenging dataset, and produces reasonable results on arbitrary masked images of real-world objects in uncontrolled environments. This suggests that the shape-from-shading and intrinsic images problem formulations may be fundamentally limited, and attention should be refocused towards developing models that jointly reason about shape and illumination in addition to shading and reflectance.

But of course, our model has some limitations. Because shading is an inherently poor cue for low-frequency shape estimation \cite{belhumeur1999bas,koenderinkConstancy} our model often makes mistakes in coarse shape estimation. To address this, we have presented a method for incorporating some external observation of shape, such as one from a stereo algorithm or a depth sensor, and we have demonstrated that by incorporating some low-frequency external shape observation (such as what a stereo algorithm or a depth sensor would provide) we can produce high-quality shape estimates. We assume that materials are Lambertian, which is often a reasonable approximation but can causes problems for objects with specularities. Thankfully, because of the modular nature of our algorithm, our simple Lambertian rendering engine can easily be replaced by a more sophisticated model. We assume that images consist of single, masked objects, while real-world natural scenes contain severe occlusion and support relationships.  We also assume illumination is global, and we ignore illumination issues such as cast shadows, mutual illumination, or other sources of spatially-varying illumination \cite{forsyth2011VSSA,Forsyth}. To address these two issues of occlusion and spatially-varying illumination in natural scenes, we have investigated into the interplay between SIRFS and segmentation techniques, by generalizing SIRFS to a mixture model of shapes and lights which are embedded in a  soft segmentation of a scene \cite{Barron2013A}. Another limitation of our technique is that our priors on shape and reflectance are independent of the category of object present in the scene. We see this as a strength of our model, as it means that our priors are general enough to generalize across object categories, but presumably an extension of our model which uses object recognition techniques to produce class-specific priors should perform better.

\ifCLASSOPTIONcompsoc
  \section*{Acknowledgments}
\else
  \section*{Acknowledgment}
\fi

J.B. was supported by NSF GRFP and ONR MURI N00014-10-10933. Thanks to Trevor Darrell, Bruno Olshausen, David Forsyth, Bill Freeman, Ted Adelson, and Estee Schwartz.

\ifCLASSOPTIONcaptionsoff
  \newpage
\fi

\bibliographystyle{IEEEtran}
\bibliography{IEEEabrv,SIRFS.bib}

\begin{appendices}

\section{Linearization and Rendering}
\label{sec:Linearization}

Here we will detail how to calculate $S(Z,L)$ (the log-shading of some depth map $Z$ subject to some spherical harmonic illumination $L$) and its analytical derivative efficiently, for the purpose of calculating $A$ and backpropagating losses on $A$ back onto $Z$ . First, we convert $Z$ into a set of surface normals:
\begin{eqnarray}
N^x = \frac{Z * h_3^x}{B }, \quad N^y = \frac{Z * h_3^y}{B}, \quad N^z = \frac{1}{B } \nonumber \\
B = \sqrt{1 + \left(Z * h_3^x \right)^2 + \left(Z * h_3^y \right)^2 }
\end{eqnarray}
where $*$ is convolution. We also compute the following:
\begin{eqnarray}
  F_{11} &=& \left(1 - N^x \times N^x \right) \times N^z \nonumber \\
  F_{22} &=& \left(1 - N^y \times N^y \right) \times N^z \nonumber \\
  F_{13} &=& -\left( N^x \times N^z \times N^z \right) \nonumber \\
  F_{23} &=& -\left( N^y \times N^z \times N^z \right) \nonumber\\
  F_{12} &=&  -\left( N^x \times N^y \times N^z \right) 
\end{eqnarray}
where $\times$ is component-wise multiplication of two images.
Let us look at the surface normal at one pixel: $\mathbf{n}_i = [N^x_i, N^y_i, N^z_i]^\mathrm{T}$. Rendering that point with spherical harmonics is:
\begin{eqnarray}
S(\mathbf{n}_i, L ) &=& [\mathbf{n}_i; 1]^\mathrm{T} \mathrm{M} [\mathbf{n}_i; 1] \\
\mathrm{M} &=& \left[ \begin{array}{cccc}
c_1 L_9 &  c_1 L_5 &  c_1 L_8 &  c_2 L_4 \\
c_1 L_5 & -c_1 L_9 &  c_1 L_6 &  c_2 L_2 \\
c_1 L_8 &  c_1 L_6 &  c_3 L_7 &  c_2 L_3 \\
c_2 L_4 &  c_2 L_2 &  c_2 L_3 &  c_4 L_1 - c_5 L_7
      \end{array} \right] \nonumber
\end{eqnarray}
\begin{eqnarray}
c_1 = 0.429043  & c_2 = 0.511664 \nonumber \\
c_3 = 0.743125  & c_4 = 0.886227 & c_5 = 0.247708  \nonumber 
\end{eqnarray}
Note that $S(\mathbf{n}_i, L )$ is the \emph{log}-shading at pixel $i$, not the shading. This is different from the traditional usage of spherical harmonic illumination. Directly modeling log-shading makes optimization easier by guaranteeing that shading is greater than $0$ without needing to clamp shading at $0$, as is normally done. The gradient of the log-shading at this point with respect to the surface normal is:
$$
B_i = \nabla_{\mathbf{n}_i} S(\mathbf{n}_i, L ) =  2\mathbf{n}_i^\mathrm{T} \mathrm{M}[: , 1:3]
$$
Where $B$ is a three-channel image, where $B^x$ is the gradient of $S$ with respect to $N^x$, etc. Given the log-shading, we can infer what the log-albedo at this point must be:
\begin{eqnarray}
A_i = I_i - S(\mathbf{n}_i, L)
\end{eqnarray}
After calculating $g(A)$ and $\nabla_{A} g(A)$, we can backpropagate the gradient onto $Z$ as follows:
\begin{eqnarray}
D_S &=& -\nabla_A g(A) \\
D_x &=& B^x \times F_{11} + B^y \times F_{12}  + B^z \times F_{13} \nonumber \\
D_y &=& B^x \times F_{12}  +  B^y \times F_{22} + B^z \times F_{23}  \nonumber \\
\nabla_Z g(A) &=& (D_S \times D_x) \star h_3^x + (D_S \times D_y) \star h_3^y \nonumber
\end{eqnarray}
where $\times$ is component-wise multiplication of two images and $\star$ is cross-correlation.

Let us construct the matrix $J$, the Jacobian matrix of all partial derivatives of $S$ with respect to $L$, which is a $n$ by $9$ matrix (where $n$ is the number of pixels in $S$), where row $i$ is:
\begin{eqnarray}
J_i = \big[c4, 2c_2N_i^y, 2c_2N_i^z, 2c_2N_i^x, 2c_1N_i^xN_i^y, 2c_1 N_i^yN_i^z, \nonumber \\
c_3 N_i^zN_i^z - c_5, 2c_1N_i^xN_i^z, c_1(N_i^xN_i^x - N_i^yN_i^y)\big] \nonumber
\end{eqnarray}
We can use this matrix to backpropagate the gradient of the loss with respect to $S$ onto $L$, as follows:
\begin{eqnarray}
\nabla_L g(A) &=& J^\mathrm{T} D_S
\end{eqnarray}

We have described how to linearize a depth map, compute a log-shading image of that linearization with respect to a grayscale spherical-harmonic model of illumination, and backpropagate a gradient with respect to that shading image onto the depth map and the illumination model. To do the same for a color image, we simply do the same procedure three times --- though for efficiency's sake we need only linearize the depth map once.

\section{Efficient Quadratic Entropy}
\label{sec:quadratic}

Here we will detail a novel method for calculating the quadratic entropy measure introduced in \cite{Principe98learningfrom}, which we use in our parsimony prior on log-reflectance. Let $\mathbf{x}$ be a vector, $N$ is the length of $\mathbf{x}$, and $\sigma$ is the  bandwidth parameter (the width of the Gaussian bump around each element of $\mathbf{x}$). Then the quadratic entropy of $\mathbf{x}$ under the Parzen window defined by $\mathbf{x}$ and $\sigma$ is defined as:
\begin{eqnarray}
H(\mathbf{x}) &=& -\log \left( \frac{1}{Z} \sum_{i=1}^N \sum_{j=1}^N \exp\left( -\frac{(\mathbf{x}_i - \mathbf{x}_j)^2}{4\sigma^2}\right) \right) \label{eq:naiveV} \nonumber\\
Z &=& N^2 \sqrt{4 \pi \sigma^2}
\end{eqnarray}
Note that we will use $H(\cdot)$ here to describe entropy, rather than mean curvature. Our first insight is that this can be re-expressed as a function on a histogram of $\mathbf{x}$. Let $W$ be the bin-width of the histogram of $\mathbf{x}$, let $M$ be the number of bins, and let $\mathbf{n}_a$ be the count of $\mathbf{x}$ in bin $a$. Then:
\begin{equation}
H(\mathbf{n}) = -\log \left( \sum_{a=1}^M \mathbf{n}_a  \sum_{b=1}^M \frac{\mathbf{n}_b}{Z} \exp\left( -\frac{ W^2 (a-b)^2}{4 \sigma^2}  \right) \right)
\end{equation}
Though the computation complexity of this formulation is still quadratic with respect to $M$, if the histogram is constructed such that many datapoints fall in the same bin this formulation can be much more efficient in practice. Our second insight is that this can be expressed as a convolution of $\mathbf{n}$ with a small Gaussian filter.  Let $\mathbf{g}$ be a Gaussian filter:
\begin{equation}
\mathbf{g}_d = \frac{1}{Z} \exp\left(  -\frac{W^2 d^2 }{ 4\sigma^2 } \right)
\end{equation}
Where $d$ is distance from the center. With this, we can rewrite $H(\mathbf{n})$ as follows:
\begin{equation}
H(\mathbf{n}) = -\log \left( \mathbf{n}^\mathrm{T} (\mathbf{n} * \mathbf{g}) \right)
\end{equation}
Where $*$ is convolution. This quantity is extremely efficient to compute, provided that the lengths of $\mathbf{n}$ and $\mathbf{g}$ are small, which is true provided that the range of $\mathbf{x}$ is not much larger than $\sigma$, which is generally true in practice.

This formulation also allows us to easily compute the gradient of $V(\mathbf{n})$ with respect to $\mathbf{n}$:
\begin{equation}
\nabla H(\mathbf{n}) = \left( \frac{-2 }{ \mathbf{n}^\mathrm{T} (\mathbf{n} * \mathbf{g})}  \right) (\mathbf{n} * \mathbf{g})
\end{equation}
Histogramming is a non-smooth operation, making this approximation to entropy not differentiable with respect to $\mathbf{x}$. However, if instead of standard histogramming we use linear interpolation to construct $\mathbf{n}$, then the gradient with respect to $\mathbf{x}$ is non-zero and can be calculated easily.

Let $R_L$ and $R_U$ define the bounds on the range of the bins, with $R_U = \mathrm{min}(\mathbf{x})$ and $R_L = \mathrm{max}(\mathbf{x})$. The fenceposts assigned to datapoint $x_i$ are $b_L$ and $b_U$, where $b_L$ is the largest fencepost below it, and $b_U$ is the smallest fencepost above it:
\begin{eqnarray}
b_L = \lfloor (x_i - R_L) / W \rfloor, \quad b_U = b_L + 1
\end{eqnarray}
 $x_i$ will be assigned to those bins according to these weights:
 \begin{eqnarray}
 w_L = (x_i - b_L)/W, \quad  w_U = 1-w_L
 \end{eqnarray}
When adding $x_i$ to the histogram, we just add these two weights to the appropriate bins:
\begin{eqnarray}
\mathbf{n}_L = \mathbf{n}_L + w_L, \quad
\mathbf{n}_U = \mathbf{n}_U + w_U
\end{eqnarray}
The partial derivatives of the histogram with respect to $x_i$ are simple:
\begin{equation}
\frac{\partial \mathbf{n}_L }{ \partial x_i} = -\frac{1}{W}, \quad \frac{\partial \mathbf{n}_U }{ \partial x_i} = \frac{1}{W}
\end{equation}
With this, we can construct the Jacobian $J$ of $\mathbf{n}$ with respect to $\mathbf{x}$, which is a $M$ by $N$ sparse matrix. With this, we can calculate the gradient of $H$ with respect to $\mathbf{x}$:
\begin{eqnarray}
\nabla H(\mathbf{x}) \approx J^\mathrm{T} \nabla H(\mathbf{n})
\end{eqnarray}

This approximation to quadratic entropy is, in practice, extremely efficient and extremely accurate. Other techniques exist for computing approximations to this quantity, most notably the fast Gauss transform and the improved fast Gauss transform. Also, $H(\mathbf{x})$ could be computed exactly using the naive formulation in Equation~\ref{eq:naiveV}. The naive formulation is completely intractable, as the computation complexity is $O(N^2)$. The FGT-based algorithms are $O(N \log N)$, and provide no efficient way to compute  $\nabla H(\mathbf{x})$, which makes those algorithms impossible to use in our gradient-based optimization scheme. Our approximation has a complexity of $O(N)$ (provided the kernel in the convolution is small) and allows for $\nabla H(\mathbf{x})$ to be approximated extremely efficiently. In practice, our model produces approximations of entropy that are usually within $0.01\%$ of the true entropy, which is similar to the accuracy obtained using the fast Gauss transform or the improved fast Gauss transform, and is  $10$ or $100$ times faster than the FGT-based algorithms.

This techniques for computing quadratic entropy for a univariate signal can easily be generalized to higher dimensions. We use a three-dimensional generalization to compute the quadratic entropy of a color (whitened) log-reflectance image. Instead of constructing a 1D histogram with linear interpolation, we construct a 3D histogram using trilinear interpolation, and instead of convolving our 1D kernel with a Gaussian filter, we convolve the 3D histogram with three separable Gaussian filters.

Note that this formulation is extremely similar to the bilateral grid~\cite{Chen2007}, which is a tool for high-dimensional Gaussian filtering (used mostly for bilateral filtering, hence the name). The calculation of our entropy measure is extremely similar to the ``splat, blur, slice'' pipeline in other high-dimensional Gaussian filtering works~\cite{Adams2012}, except that after the ``slice'' operation we take the inner product of the input ``signal'' and the blurred output signal. This means that we need not actually compute the slice operation, but can instead just compute the inner product directly in the histogram space. This connection means that the body of work for efficiently computing this quantity in the context of image filtering can be directly adapted to the problem of computing high-dimensional entropy measures. Recent work \cite{Adams2012} suggests that for dimensionalities of $3$, our bilateral grid formulation is the most efficient among existing techniques, but that this entropy measure could be computed reasonably efficiently in significantly higher-dimensional spaces (up to $8$ or $16$) using more sophisticated techniques.

\section{Mean Curvature}
\label{sec:MeanCurvature}

Here we will detail how to calculate $H(Z)$ (the mean curvature of a depth map $Z$, not entropy) and its analytical derivative efficiently. Mean curvature on a surface is a function of the first and second partial derivatives of that surface.
\begin{equation}
H(Z)  = \frac{\left(1 + Z_x^2 \right) Z_{yy} -2 Z_x Z_y Z_{xy} + \left(1 + Z_y^2 \right) Z_{xx} }{ 2 \left(1 + Z_x^2 + Z_y^2 \right)^{3/2}}
\end{equation}
To calculate this for a discrete depth map, we will first approximate the partial derivatives using filter convolutions.
\begin{eqnarray}
Z_x = Z * h_3^x, \quad Z_y = Z * h_3^y \quad\quad\quad\quad \\
\quad Z_{xx} = Z * h_3^{xx}, \,\, Z_{yy} = Z * h_3^{yy}, \,\, Z_{xy} = Z * h_3^{xy} \nonumber
\end{eqnarray}
\begin{equation}
\resizebox{2.0 in}{!}{$
h_3^x = \frac{1}{8} \left[ \begin{array}{@{\,}c@{\,}c@{\;\;}c@{\,}c@{\;\;}c@{\,}c@{\,}}
&1 & &0 & \text{--}&1 \\
&2 & &0 & \text{--}&2   \\
&1 & &0 & \text{--}&1
\end{array} \right], \quad
h_3^y = \frac{1}{8} \left[ \begin{array}{@{\,}c@{\,}c@{\;\;}c@{\,}c@{\;\;}c@{\,}c@{\,}}
&1 &  &2 &  &1 \\
&0 & &0 & &0    \\
\text{--}&1 & \text{--}&2 & \text{--}&1
\end{array} \right]
$} \nonumber
\end{equation}
\begin{equation}
\resizebox{3.2 in}{!}{$
h_3^{xy} = \frac{1}{4} \left[ \begin{array}{@{\,}c@{\,}c@{\;\;}c@{\,}c@{\;\;}c@{\,}c@{\,}}
&1 & &0 & \text{--}&1 \\
&0 & &0 & & 0   \\
\text{--}&1 & &0 & &1
\end{array} \right], \quad
h_3^{yy} = \frac{1}{4} \left[ \begin{array}{@{\,}c@{\,}c@{\;\;}c@{\,}c@{\;\;}c@{\,}c@{\,}}
&1 &  &2 &  &1 \\
\text{--}&2 & \text{--}&4 & \text{--}&2    \\
&1 & &2 & &1
\end{array} \right], \quad
h_3^{xx} = \frac{1}{4} \left[ \begin{array}{@{\,}c@{\,}c@{\;\;}c@{\,}c@{\;\;}c@{\,}c@{\,}}
&1 & \text{--} &2 &  &1 \\
&2 & \text{--}&4 & &2    \\
&1 & \text{--} &2 & &1
\end{array} \right]
$} \nonumber
\end{equation}
We then compute the following intermediate ``images'', and use them to compute $H(Z)$.
\begin{eqnarray}
M &=& \sqrt{1 + Z_x^2 + Z_y^2 }  \nonumber \\
N &=& (1 + Z_x^2)Z_{yy} -2Z_xZ_yZ_{xy} + (1 + Zy^2)Z_{xx} \nonumber \\
D &=& 2M^3 \nonumber \\
H(Z) &=& N / D
\end{eqnarray}
When computing $H(Z)$, we also compute the following, which are stored until after the loss function with respect to $H(Z)$ has been calculated, at which point they will be used to backpropagate the gradient of the loss function using the chain rule.
\begin{eqnarray}
F_x &=& 2(Z_x Z_{yy} - Z_{xy} Z_y) - \frac{3 Z_x N }{ M^2} \nonumber \\
F_y &=& 2(Z_{xx} Z_y - Z_x Z_{xy}) - \frac{3 Z_y N }{ M^2} \nonumber \\
F_{xx} &=& 1 + Z_y^2 \nonumber \\ 
F_{yy} &=& 1 + Z_x^2 \nonumber \\
F_{xy} &=& -2 Z_x Z_y
\end{eqnarray}
Given $f(H(Z))$ and $\nabla_{H(Z)} f$, a loss function and the gradient of that loss function with respect to $H(Z)$, we can calculate $\nabla_Z f$, the gradient of the loss with respect to $Z$, as follows:
\begin{eqnarray}
B &=& \frac{\nabla_{H(Z)} f }{ D} \\
\nabla_Z f &=& (B F_x) \star h_3^x + (B F_y) \star h_3^y \nonumber \\
&+& (B F_{xx}) \star h_3^{xx} + (B F_{yy}) \star h_3^{yy}  + (B F_{xy}) \star h_3^{xy} \nonumber
\end{eqnarray}
Adjacent variables are component-wise multiplication of two images, $/$ is component-wise division, $*$ is convolution and $\star$ is cross-correlation.

\section{Noisy Shape Observation}
\label{sec_shape}

One of the reasons that using shading cues to recover shape (as we are attempting here) is challenging, is that shading is a fundamentally poor cue for low-frequency (coarse) shape variation. Shading is directly indicative of only the shape of a point relative to its neighbors: fine-scale variations in shape produce sharp, localized changes in an image, while coarse-scale shape variations produce very small, subtle changes across an entire image. Both algorithms \cite{belhumeur1999bas} and humans \cite{Koenderink} therefore make errors in estimating coarse depth when using only shading. Bas relief sculptures take advantage of this by conveying the impression of a rich, deep 3D scene, using only the shading produced by a physically shallow object.

To deal with this issue, we will construct our prior on shape to allow for an external observation of shape to be incorporated into inference. This observation may be produced by a stereo algorithm, or by some depth sensor such as a laser rangefinder or the Kinect. These depth sensors or stereo algorithms often produce depth maps which are noisy or incomplete, or most often blurry --- lacking fine-scale shape detail. Because of the complementary strengths of stereo and shading, combining the two can often yield very accurate results \cite{BlakeZisserman,Barron2011}.

We will construct a loss function to encourage our recovered depth $Z$ to resemble the raw sensor depth $\hat{Z}$:
\begin{eqnarray}
f_o(Z, \hat{Z}) = \sum_{i} \left( ( (Z * b(\sigma_Z))_i - \hat{Z}_i)^2 + \epsilon^2 \right)^{\frac{\gamma_o}{2}}
\end{eqnarray}
This is simply a hyperlaplacian distribution with an exponent of $\gamma_o$ on the difference between $(Z * b(\sigma_Z) )$ and $\hat{Z}$ at every pixel, with $\epsilon$ added in to make the loss differentiable everywhere. $b(\sigma_Z)$ is a 2D Gaussian filter with a standard deviation of $\sigma_Z$, and $*$ is convolution, so $(Z * b(\sigma_Z))_i$ is the value of a blurry version of our shape estimate $Z$ at pixel location $i$.  We tune $\gamma_o$ on the training set, which sets it to $\sim 1$, and we set  $\epsilon = 1/100$. The robust nature of this cost encourages $Z$ to resemble $\hat{Z}$, while allowing it to occasionally differ drastically. In our experiments we use $Z^* * b(30)$ as our $\hat{Z}$, which is a reasonably proxy for a stereo algorithm or low-resolution depth-sensor, and we set $\sigma_Z = 30$ as that value (unsurprisingly) performs best during cross-validation.

For the one variant of our model in which we incorporate a noisy external shape observation, our prior on shapes gains an additional term, and becomes:
\begin{equation}
f(Z) = \lambda_k f_k(Z) + \lambda_i f_i(Z)  + \lambda_c f_c(Z) + \lambda_o f_o(Z, \hat Z)
\end{equation}
Where $\lambda_o$ is cross-validated on the training set.

\section{Efficient Computation}
\label{sec:Efficient}

Our model is fairly computationally expensive. Evaluating our loss function and its gradient takes close to a second, and optimization requires that the loss be evaluated hundreds of times. To make this model more tractable, we use some additional tricks to speed up the computation of the loss function.

First, our smoothness priors for reflectance and shape require repeatedly computing the negative log-likelihood of a Gaussian scale mixture. Computing this naively is very expensive, but it can be made extremely efficient by pre-computing a lookup table of the negative log-likelihood, and indexing into that to compute the gradient and its loss. For the multivariate GSM used in our smoothness prior for color reflectance, we can construct a lookup table of negative log-likelihood with respect to Mahalanobis distance under the covariance matrix $\Sigma$ in our GSM. 

When computing our smoothness priors, it's often fastest to pre-compute the pairs of pixels within all $5 \times 5$ windows, and construct a sparse matrix where for each pair, we have a row in which the column corresponding to one pixel in the pair is set to $1$ and the column corresponding to the other pixel is set to $-1$. With this, a vector of pairwise distances between pixels can be computed efficiently with one sparse matrix-vector product. Also, expressing this pairwise distance computation as a matrix multiplication allows gradients to be easily backpropagated from the vector of differences onto the raw pixels by simply multiplying the gradient vector by the transpose of this matrix.

The prior for absolute reflectance can be computed efficiently using the same bilateral-grid trick used for entropy: splat the signal into a histogram, compute the loss of the histogram, and then backpropagate onto the data. For even more efficiency, we can use the same histogram for both the entropy prior and the absolute prior, which means that for each evaluation of the loss function, we only need to compute one histogram from the reflectance and one backpropagation from the histogram to the reflectance.

\section{Error Metrics}

Choosing good error metrics for this task is challenging. We will use the geometric mean of six error metrics: two for shape, one for illumination, one for shading, one for reflectance, and the MIT intrinsic images error metric introduced in \cite{grosse09intrinsic}, which we will refer to as $\mitMSE$ (though which the original authors call ``LMSE''). We use the geometric mean of these metrics as it is insensitive to the different dynamic ranges of the constituent error metrics, and is difficult to trivially minimize in practice.

Our first shape error metric is:
\begin{equation}
\ZMAE(\hat Z, Z^*) = \frac{1}{n} \min{\beta} \sum_{x,y}    \left| \hat Z_{x,y} - Z^*_{x,y} + b \right|
\end{equation}
This is the shift-invariant absolute error between the estimated shape $\hat Z$ and the ground-truth shape $Z^*$.  This error metric is sensitive to all errors in shape estimation, except for the absolute distance of the shape from the viewer (which is unknowable under orthographic projection). It can be computed efficiently by setting $b$ to the median of $\hat Z - Z^*$.

Our second shape error metric is:
\begin{equation}
\NMAE(\hat N, N^*) = \frac{1}{n} \sum_{x,y}    \mathrm{\arccos}\left(\hat N_{x,y} \cdot N^*_{x,y} \right)
\end{equation}
This is the mean error between the normal field $\hat N$ of our estimated shape $\hat Z$ and the normal field $N^*$ of the ground-truth shape $Z^*$, in radians. This metric is most sensitive to very fine-scale errors in $\hat Z$, which is what determines surface orientation.

For illumination, our error metric is:
\begin{equation}
\lightMSE(\hat L, L^*) = \frac{1}{n} \min{\alpha} \sum_{x,y}  || \alpha V(\hat L)_{x,y} - V(L^*)_{x,y}  ||_2^2
\end{equation}
Which is the scale-invariant MSE of a rendering of our recovered illumination $\hat L$ and the ground-truth illumination $L^*$. $V(L)$ is a function that renders the spherical harmonic illumination $L$ on a sphere and returns the log-shading. $V(L)_{x,y}$ is a 3-vector of log-RGB at position $(x,y)$ in the renderings. The $\alpha$ multiplier makes this error metric invariant to absolute scaling, meaning that estimating illumination to be twice as bright or half as bright doesn't change the error. But because there is only one multiplier rather than individual scalings for each RGB channel, this error metric is sensitive to the overall color of the illuminant. This choice seems consistent with what we would like: estimating absolute intensity of an illuminant from a single image is both incredibly difficult and not very useful, but estimating the color of the illuminant is a reasonable thing to expect from an algorithm, and would be useful for many applications (color constancy, relighting, reflectance estimation, etc). We impose our error metric in the space of visualizations of the illumination rather than in the space of the actual spherical harmonic coefficients that generated that visualization, both because it makes our error metric invariant to the choice of illumination model, and because we found that often the recovered illumination could look quite similar to the ground-truth, while having a very different spherical harmonic representation.

For shading and reflectance, we use:
\begin{eqnarray}
\SMSE(\hat s, s^*) = \frac{1}{n} \min{\alpha} \sum_{x,y} \norm{  \alpha \hat s_{x,y} - s^*_{x,y}   }_2^2 & \\
\AMSE(\hat r, r^*) = \frac{1}{n} \min{\alpha} \sum_{x,y} \norm{  \alpha \hat r_{x,y} - r^*_{x,y}   }_2^2 &
\end{eqnarray}
These are the scale-invariant MSEs of our recovered shading $\hat s = \exp(S(\hat Z, \hat L))$ and reflectance $\hat r = \exp(\hat R)$. Just like in $\lightMSE$, we are invariant to absolute scaling of all RGB channels at once, but not invariant to scaling each channel individually. This makes these error metrics sensitive to errors in estimating the overall color of the shading and reflectance images, but invariant to illumination. Note that these error metrics are of shading and reflectance, not of log-shading and log-reflectance, even though the rest of this paper is written almost entirely in terms of log-intensity. We could have used shift-invariant error metrics in log-intensity space, but we found these to be too sensitive to errors in dark regions of the image --- places in which we'd expect any algorithm to do worse, simply because there is less signal.

Our final error metric is the metric introduced in conjunction with the MIT intrinsic images dataset \cite{grosse09intrinsic}, which the authors refer to as LMSE, but which we will call $\mitMSE$ to minimize confusion with $\lightMSE$. This metric measures error for both reflectance and shading, and is locally scale-invariant. The intent of the local scale-invariance is to make the metric insensitive to low-frequency errors in either shading or reflectance. In keeping with this spirit, we apply this error metric individually to each RGB channel and take the mean of those three errors as $\mitMSE$, making this error metric not just robust to low-frequency error, but robust to most errors in estimating the color of the illumination. This error metric therefore serves to be somewhat complementary to $\SMSE$ and $\AMSE$, which are sensitive to everything except absolute intensity.

$\mitMSE$ is the mean of the local scale-invariant MSE of shading and reflectance, normalized so that an estimate of all zeros has the maximum score of 1:
\begin{equation}
\mitMSE(\hat s, \hat r, s^*, r^*) = \frac{1}{2} \left( \frac{\IMSE(\hat s, s^*) }{ \IMSE(\hat s, 0)} + \frac{\IMSE(\hat r, r^*) }{ \IMSE(\hat r, 0)} \right)
\end{equation}
Where $\IMSE(\cdot)$ is the sum of the scale-invariant MSE at all local windows $w$ of size $20 \times 20$, spaced in steps of $10$:
\begin{equation}
\IMSE(\hat x, x^*) = \sum_{w \in W} \min{\alpha} \norm{  \alpha \hat x_w - x^*_w   }_2^2
\end{equation}

As an aside, in our error metrics we repeatedly use scale-invariant MSE, of the form:
\begin{equation}
\min{\alpha} \norm{ \alpha \hat x - x^* }_2^2
\end{equation}
The closed-form solution to this problem is:
\begin{equation}
\norm{ \left( \frac{\hat{x}^\mathrm{T} x^* }{ \hat{x}^\mathrm{T} \hat{x}} \right) \hat x - x^* }_2^2
\end{equation}

\section{Dataset}
\label{sec:dataset}

Here we will detail how we recover ``ground-truth'' shape and spherical harmonic illumination for each image of each object in our dataset. This is a simple photometric stereo algorithm, in which we optimize over shapes and illuminations to minimize the absolute error between renderings of our dataset and the actual images in our dataset. Absolute error is used to give us robustness to errors due to shadows and specularities, which our rendering engine (and therefore, our dataset) do not consider or address properly. Recovered shapes and illuminations were then cleaned up by hand to address bas-relief ambiguity issues\cite{belhumeur1999bas}. We treat each RGB channel of each image as a separate image.

To account for varying reflectance, we compute a ``shading'' image for each image on our dataset.
\begin{eqnarray}
s^*_{i,j} = \exp(I_{i,j} - R_i)
\end{eqnarray}

We will now detail each step in the inner loop of our iterative photometric stereo algorithm.  We first take each current shape estimate $Z$, and linearize it to get a set of fixed surface normals. For each image $j$, we solve for the SH illumination that minimizes absolute error between the rendering and the shading image:
\begin{eqnarray}
L_j \leftarrow \argmin{L} \sum_i | \exp(S(\mathbf{n}_i, L)) -  s^*_{i,j} | 
\end{eqnarray}
This optimization problem is solved using Iteratively Reweighted Least-Squares. We then fix each image's illumination $L_j$, and optimize over each object's normals $\mathbf{n}_i$.
\begin{eqnarray}
\mathbf{n}_i \leftarrow \argmin{\mathbf{n}} \sum_j | \exp(S(\mathbf{n}, L_j)) - s^*_{i,j} | 
\end{eqnarray}
This optimization is done with L-BFGS. In this step, the normals are decoupled, and so surface integrability is not enforced. Given this estimate of surface normals, we can compute a integrable surface $Z$ which approximates this normal field using least-squares: 
\begin{equation}
Z \leftarrow \argmin{Z} \sum_i \left( Z * h^x - \frac{\mathbf{n}^x_i}{ \mathbf{n}^z_i} \right)^2 + \left( Z * h^y - \frac{\mathbf{n}^y_i }{ \mathbf{n}^z_i } \right)^2 \nonumber
\end{equation}

These three optimization steps are repeated until convergence ($30$ iterations). For the first $10$ iterations, we constrain all of the illuminations belonging to the same object to be scaled and shifted versions of each other, but for the next $20$ iterations we allow each illumination for every image to vary freely. The result of this algorithm is an estimate of $Z$ for each object and an estimate of $L$ for each RGB channel of every image.

This photometric stereo algorithm still suffers from Bas-Relief ambiguity\cite{belhumeur1999bas} issues, despite the abundance of data. We therefore manually adjust each recovered $Z$ over the three parameters of the Bas-Relief ambiguity by hand. Also, some regions of $Z$ are clearly incorrect due to shadows. These regions are manually removed (and are not included in the evaluation of our error metrics which concern $Z$).  After these manual tweaks to each shape, we update the set of illuminations  to minimize absolute error once again.  The two ``cup'' and ``teabag'' images did not have discriminative enough shape features for photometric stereo to recover reasonable second-order spherical harmonic illuminations, so for those objects we instead recover only first-order spherical harmonic illumination parameters (equivalent to point-light + ambient illumination), and set the other coefficients to $0$.

The MIT Intrinsic Images dataset was not acquired with the goal of having the product of the ``shading'' and ``reflectance'' images be exactly equal to the diffuse image, which our model (and our baseline models) assume. That is, a lambertian rendering of our recovered shape and illumination resembles a scaled version of the original ``shading'' image. We correct for this by adjusting the brightness of the ``shading'' image such that it matches our rendering in a least-squares sense, and we use this ``corrected'' shading image in all of our experiments.

Note that the optimization tools we use for our photometric stereo algorithm are completely disjoint from the optimization techniques used by algorithm in our paper, despite the fact that those techniques could have been adapted to do photometric stereo. This was done intentionally to dispel any concerns that our results might be good simply because they were obtained using similar techniques as our photometric stereo algorithm.

\begin{figure}[!]
	\centering
	\begin{tabular}{@{\,}c@{\,}c@{\,}}
	\begin{tabular}{@{\,}c@{\,}c@{\,}}
	\subfigure[\scriptsize Depth map]{
	\resizebox{1.12in}{!}{\includegraphics{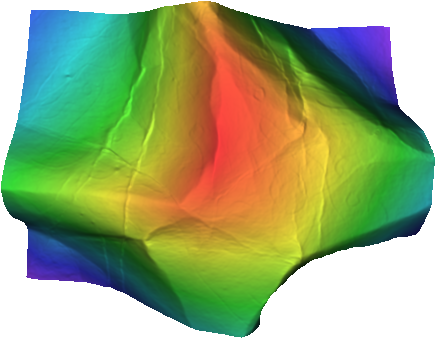}}
	\label{fig:Dataset_Z}
	}
	&
	\subfigure[\scriptsize Shading]{
	\resizebox{1.12in}{!}{\includegraphics{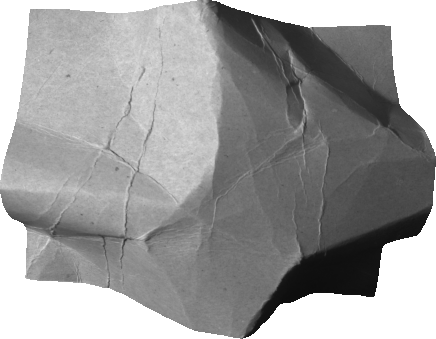}}
	\label{fig:Dataset_S}
	}
	\\
	\subfigure[\scriptsize Surface normals]{
	\resizebox{1.12in}{!}{\includegraphics{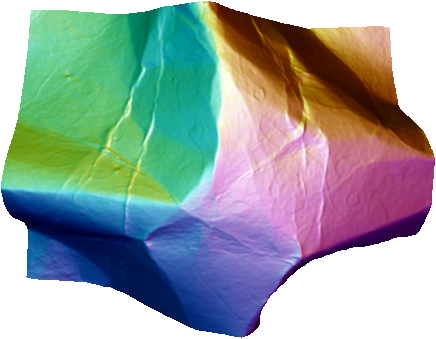}}
	\label{fig:Dataset_N}
	}
	&
	\subfigure[\scriptsize Reflectance]{
	\resizebox{1.12in}{!}{\includegraphics{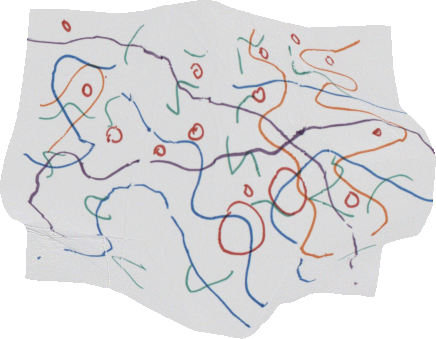}}
	\label{fig:Dataset_R}
	}
	\end{tabular}
	&
	\raisebox{-.9in}{
	\subfigure[\scriptsize images]{
	\resizebox{.6in}{!}{\includegraphics{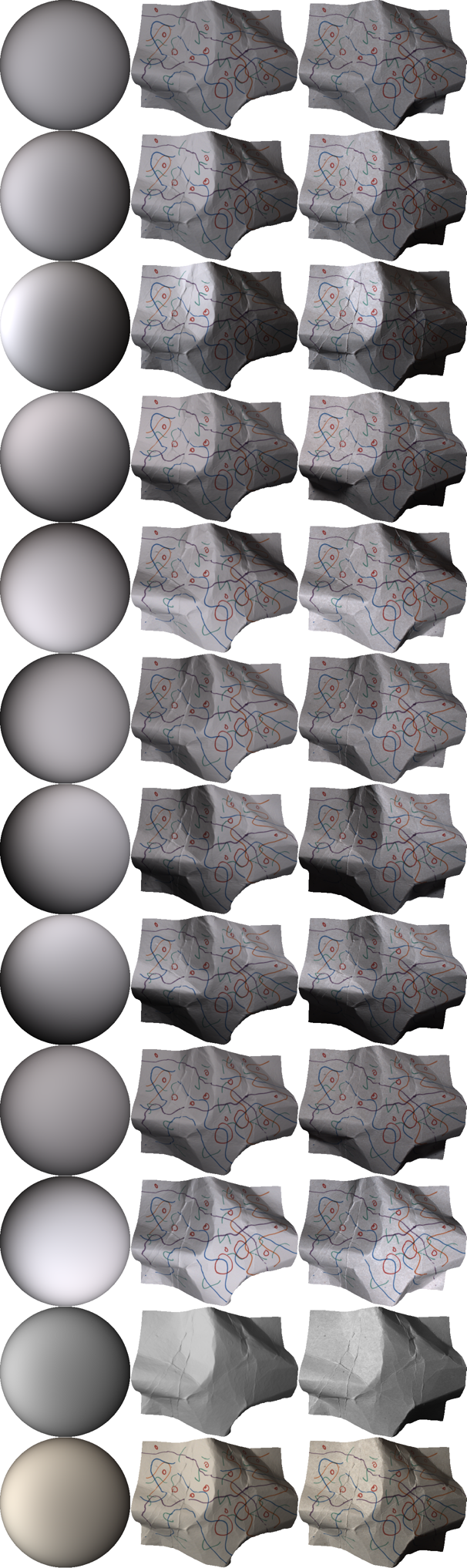}}
	\label{fig:Dataset_input}
	}
	}
	\end{tabular}
	\caption{An object from our dataset.  In \ref{fig:Dataset_Z}, \ref{fig:Dataset_S}, \ref{fig:Dataset_N}, and \ref{fig:Dataset_R} we have our ``ground-truth'' shape, shading, surface normals, and reflectance, respectively. The shading and reflectance images come from the MIT Intrinsic Images dataset \cite{grosse09intrinsic}, and the shape and surface normals were produced by our photometric stereo algorithm. In \ref{fig:Dataset_input} we have three columns, where the first contains the images from the MIT Intrinsic Images dataset \cite{grosse09intrinsic}, the third contains the illuminations recovered by our photometric stereo algorithm for each image, and the second column contains renderings of our ground-truth for each illumination, which demonstrate that our recovered models are reasonable. The second to last row of Figure~\ref{fig:Dataset_input} is the ``shading'' image from the MIT dataset, and the last row is the ``diffuse'' image, which is used as input to our model. The illumination on the last row is therefore what is referred to as the ``ground-truth'' illumination for this scene, in the rest of the paper.
 \label{fig:dataset1}}
\end{figure}

\begin{figure}[!]
	\centering
	\begin{tabular}{@{\,}c@{\,}c@{\,}}
	\begin{tabular}{@{\,}c@{\,}c@{\,}}
	\subfigure[\scriptsize Depth map]{
	\resizebox{1.12in}{!}{\includegraphics{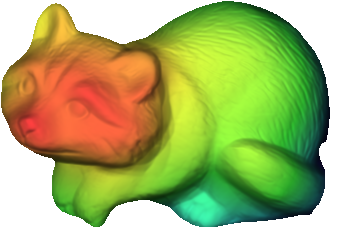}}
	}
	&
	\subfigure[\scriptsize Shading]{
	\resizebox{1.12in}{!}{\includegraphics{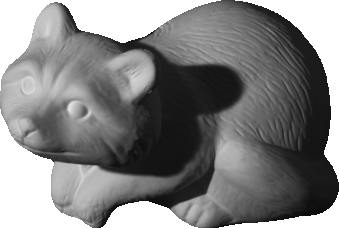}}
	}
	\\
	\subfigure[\scriptsize Surface normals]{
	\resizebox{1.12in}{!}{\includegraphics{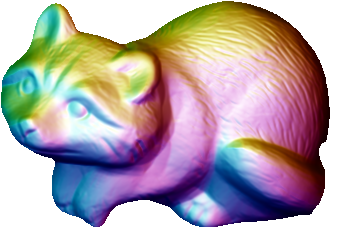}}
	}
	&
	\subfigure[\scriptsize Reflectance]{
	\resizebox{1.12in}{!}{\includegraphics{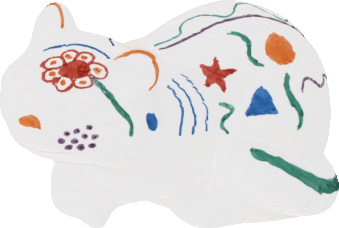}}
	}
	\end{tabular}
	&
	\raisebox{-.79in}{
	\subfigure[\scriptsize images]{
	\resizebox{.6in}{!}{\includegraphics{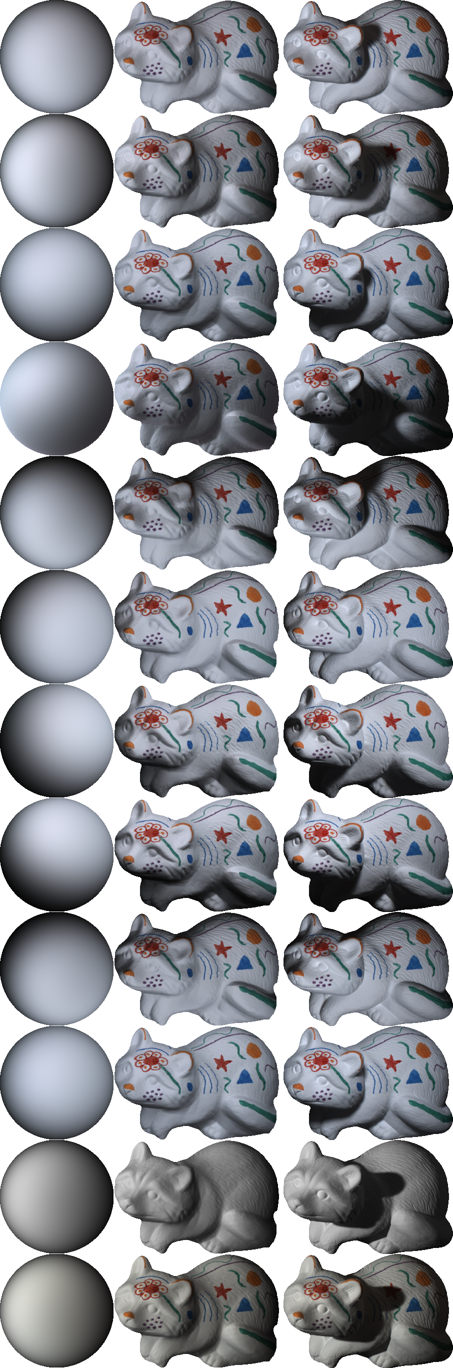}}
	}
	}
	\end{tabular}
	\caption{Another object from our dataset, shown in the same format as Figure~\ref{fig:dataset1}. Note that our illumination model cannot capture the cast shadows in the input images, which is why our renderings are shadowless.
 \label{fig:dataset2}}
\end{figure}

Examples of our recovered shapes and illuminations, as well as the shading and reflectance images already contained in the MIT Intrinsic Images dataset, can be seen in Figures~\ref{fig:dataset1} and \ref{fig:dataset2}.


\section{Shape From Shading}

Our model for recovering shape and albedo given a single image and illumination can easily be reduced to a model for doing classic shape-from-shading (recovering shape given a single image and illumination). Our optimization problem becomes:
\begin{eqnarray}
\underset{Z}{\operatorname{minimize}} && \lambda | I -  S(Z, L) | + f(Z)
\end{eqnarray}
Where $I$ is the input log-image, and $\lambda$ is a multiplier that trades off the importance of the reconstruction terms against the regularizer on $Z$. $f(Z)$ and $S(Z,L)$ are the same as defined in the paper. Optimization is done using our multiscale optimization algorithm. This SFS algorithm is similar to past algorithms which optimize over a linearized representation of a depth map, with the primary difference being our choice of $f(Z)$. 

This SFS algorithm is run on the shading images produced by the ``intrinsic image'' algorithms we benchmark against. This is a very generous comparison on our part, as we are effectively giving these other algorithms one-half of the model we present here, and we are assuming that illumination is known. We used our own shape-from-shading algorithm for fairness's sake, as it appears to outperform previous SFS algorithms. This means, however, that our improvement over these algorithms is not as much a reflection of the effectiveness of $f(Z)$ \emph{in isolation}, but is instead a demonstration of the effectiveness of optimizing over $f(Z)$ and $g(A)$ to jointly recover shape and albedo, as opposed to recovering a shading image and then recovering shape from that shading image.

\begin{figure}[t!]
	\centering
    \resizebox{3.25in}{!}{\includegraphics{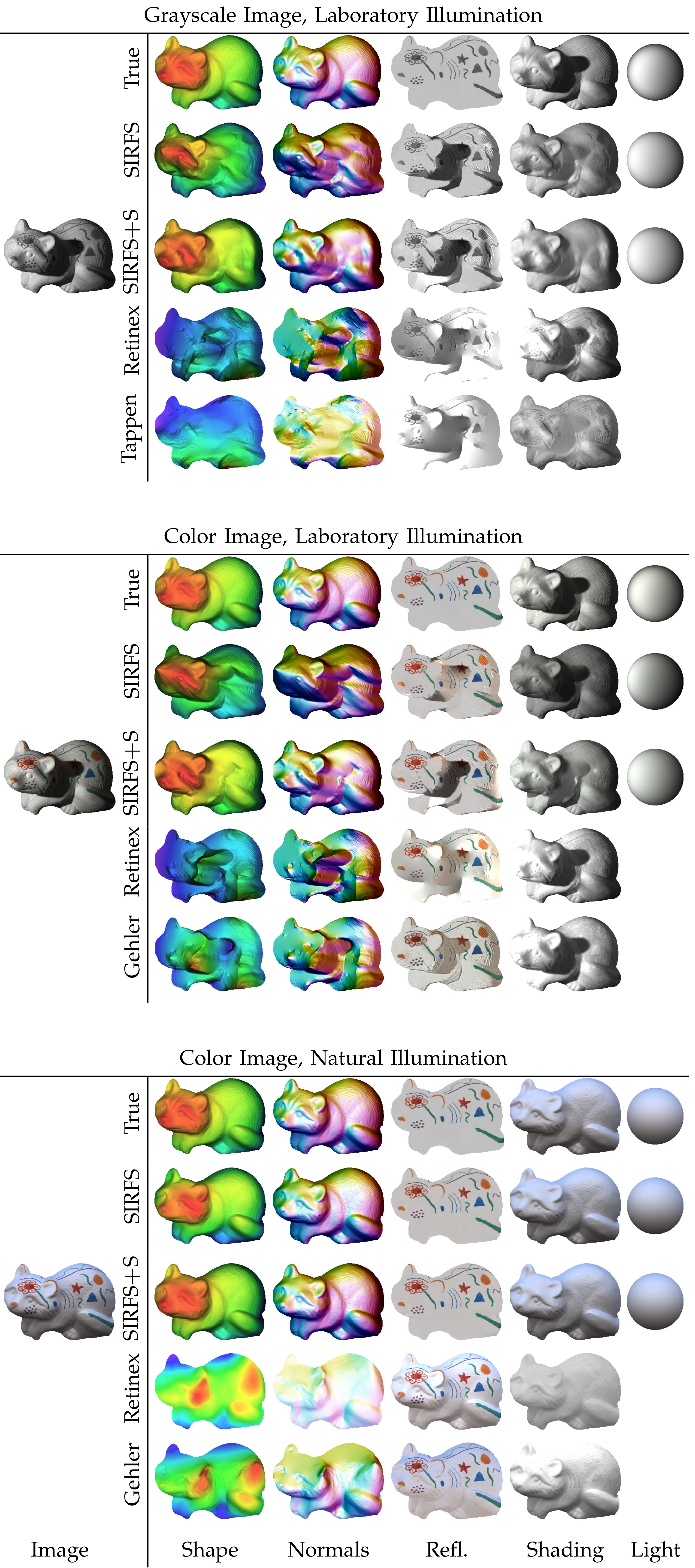}}
	\caption{Here we have a single image from the MIT-Berkeley Intrinsic Images dataset, under three color/illumination conditions. For each condition, we present the ground-truth, the output of SIRFS, the output of SIRFS$+$S (which uses external shape information), and the two best-performing intrinsic image techniques (for which we do SFS on the recovered shading to recover shape).}
\end{figure}

\begin{figure}[t!]
	\centering
    \resizebox{3.25in}{!}{\includegraphics{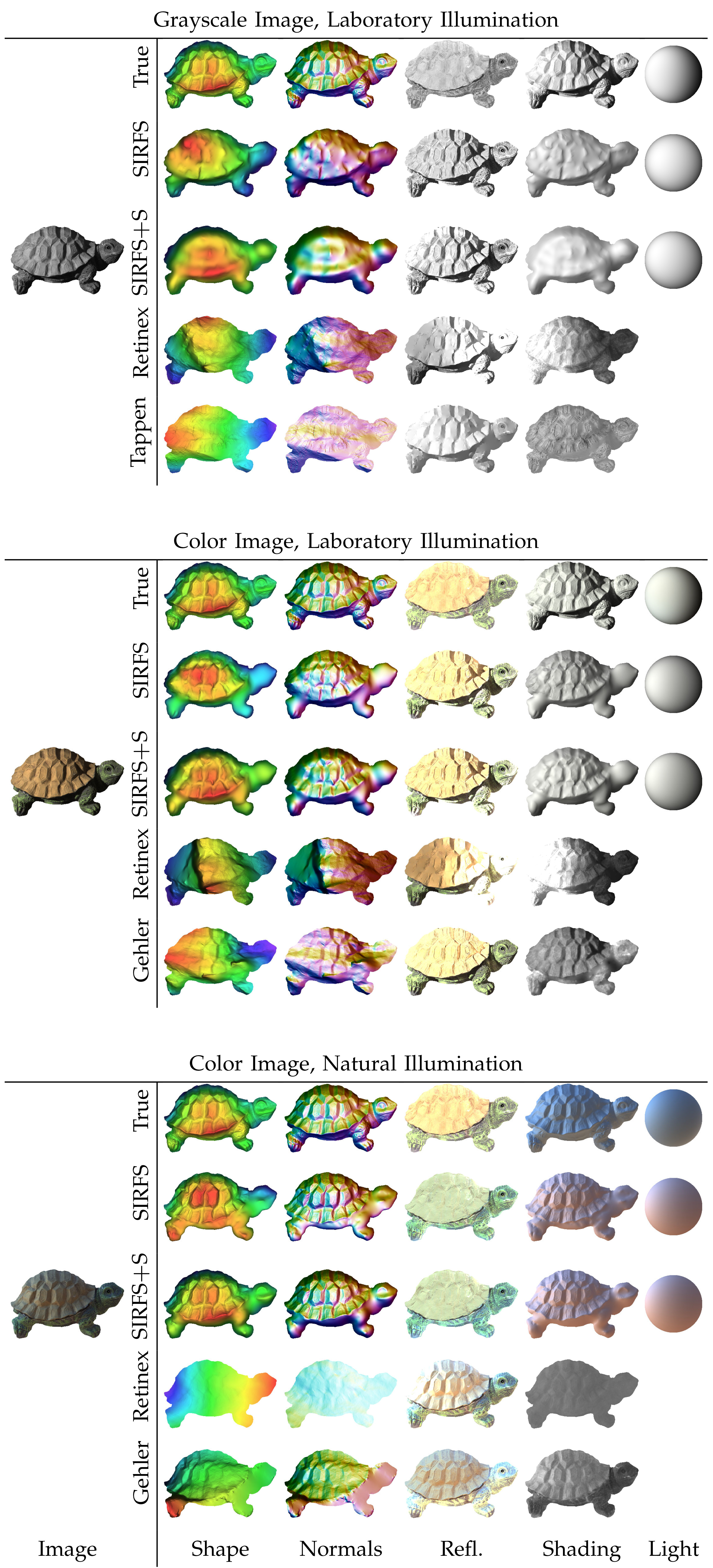}}
	\caption{Here we have another image from the MIT-Berkeley Intrinsic Images dataset.}
\end{figure}

\begin{figure}[t!]
	\centering
    \resizebox{3.25in}{!}{\includegraphics{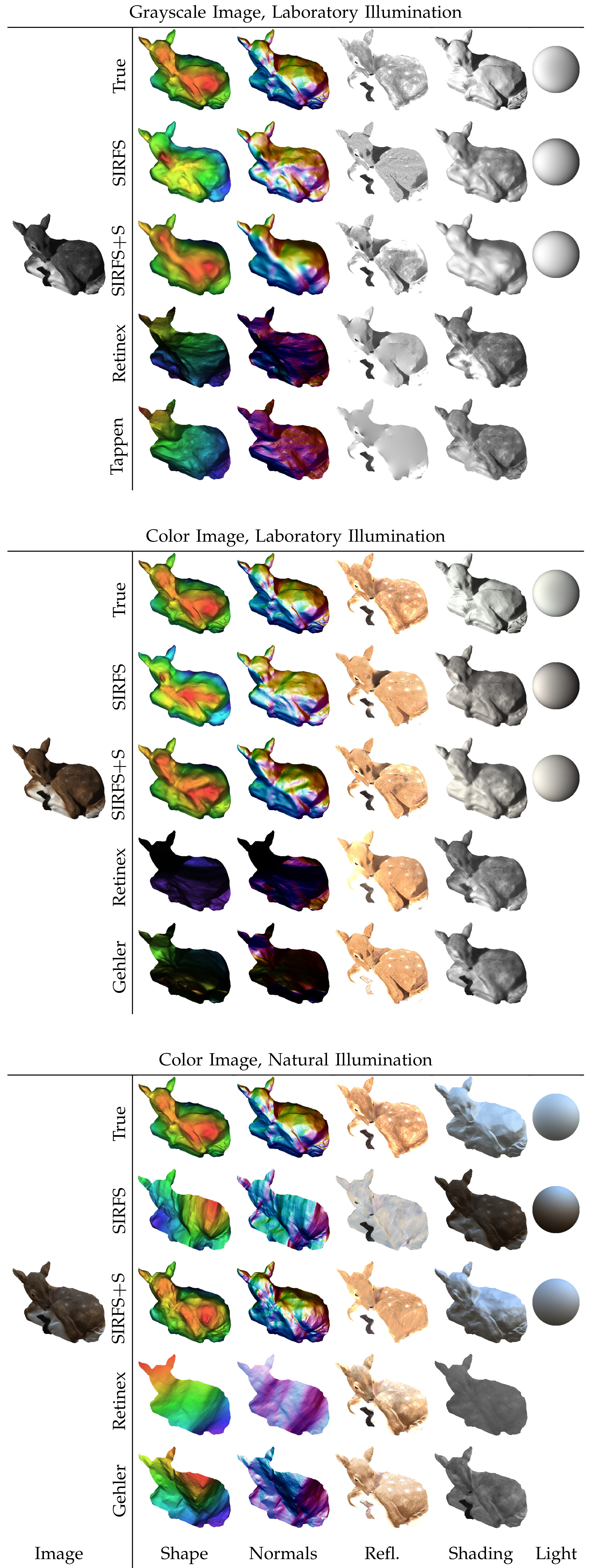}}
	\caption{Here we have another image from the MIT-Berkeley Intrinsic Images dataset.}
\end{figure}

\end{appendices}

\end{document}